\documentclass[conference]{IEEEtran}
\IEEEoverridecommandlockouts
\usepackage{cite}

\usepackage{algorithmic}
\usepackage{url}
\usepackage{graphicx}
\usepackage{textcomp}
\usepackage{xcolor}
\usepackage{import}
\usepackage{caption}
\usepackage{booktabs}
\usepackage{subcaption}
\usepackage{adjustbox}
\usepackage{booktabs}
\usepackage{comment}
\usepackage{acronym}
\usepackage{xspace}
\usepackage{amsfonts} 
\usepackage[mathscr]{eucal}
\usepackage{multirow}
\usepackage{subcaption}  
\usepackage{amsmath,amssymb,amsfonts}
\usepackage{multicol}

\DeclareMathOperator*{\argmin}{arg\,min}

\newcommand*{\eg}		{e.g.,\ }
\newcommand*{\ie}		{i.e.,\ }

\acrodef{ML} 		[\textsc{ML\xspace}]				{Machine Learning}
\acrodef{DL} 		[\textsc{DL\xspace}]				{Deep Learning}
\acrodef{ANN} 		[\textsc{ANN\xspace}]				{Artificial Neural Network}
\acrodef{DNN} 		[\textsc{DNN\xspace}]				{Deep Neural Network}
\acrodef{CNNs} 		[\textsc{CNNs\xspace}]				{Convolutional Neural Networks}
\acrodef{CNN} 		[\textsc{CNN\xspace}]				{Convolutional Neural Network}
\acrodef{MTL} 		[\textsc{MTL\xspace}]				{Multi-Task Learning}
\acrodef{RL} 		[\textsc{RL\xspace}]				{Reinforcement Learning}
\acrodef{LL} 		[\textsc{LL\xspace}]				{Lifelong Learning}
\acrodef{NLP} 		[\textsc{NLP\xspace}]				{Natural Language Processing}
\acrodef{LSTM} 		[\textsc{LSTM\xspace}]				{Long Short-Term Memory}
\acrodef{MAML} 		[\textsc{MAML\xspace}]				{Model Agnostic Meta Learning}
\acrodef{MMMTL} 	[\textsc{3MTL\xspace}]	            {Multi-modal Multi-task Meta Transfer Learning}
\acrodef{ADME}  	[\textsc{ADME\xspace}]	            {Absorption, Distribution, Metabolism, and Excretion}
\acrodef{GAN}   	[\textsc{GANs\xspace}]	               {Generative Adversarial Nets}
\acrodef{VAE}   	[\textsc{VAEs\xspace}]	               {Variational Auto-Encoders}
\acrodef{MTML}   	[\textsc{MTML\xspace}]	               {Multi-task Meta Learning}

\begin{document}

\title{Multi-Task Meta Learning: learn how to adapt to unseen tasks\\
}

          

\author{\IEEEauthorblockN{1\textsuperscript{st} Richa Upadhyay}
\IEEEauthorblockA{Lule\aa~University of Technology\\
richa.upadhyay@ltu.se}
\and
\IEEEauthorblockN{2\textsuperscript{nd} Prakash Chandra Chhipa}
\IEEEauthorblockA{Lule\aa~University of Technology\\
prakash.chandra.chhipa@ltu.se}
\and
\IEEEauthorblockN{3\textsuperscript{rd} Ronald Phlypo }
\IEEEauthorblockA{\textit{Universit\'{e} Grenoble Alpes,} \\
ronald.phlypo@grenoble-inp.fr}
\hspace{5cm}
\and
\hspace{5cm}
\IEEEauthorblockN{4\textsuperscript{th} Rajkumar Saini}
\IEEEauthorblockA{\hspace{5cm}Lule\aa~University of Technology\\
\hspace{5cm}
rajkumar.saini@ltu.se}
\and
\IEEEauthorblockN{5\textsuperscript{th} Marcus Liwicki}
\IEEEauthorblockA{Lule\aa~University of Technology\\
marcus.liwicki@ltu.se}
}
\maketitle
\begin{abstract}
This work proposes \ac{MTML}, integrating two learning paradigms \ac{MTL} and meta learning, to bring together the best of both worlds.
In particular, it focuses simultaneous learning of multiple tasks, an element of \ac{MTL} and promptly adapting to new tasks, a quality of meta learning. 
It is important to highlight that we focus on heterogeneous tasks, which are of distinct kind, in contrast to typically considered homogeneous tasks (e.g., if all tasks are classification or if all tasks are regression tasks).
The fundamental idea is to train a multi-task model, such that when an unseen task is introduced, it can learn in fewer steps whilst offering a performance at least as good as conventional single task learning on the new task or inclusion within the \ac{MTL}. 
By conducting various experiments, we demonstrate this paradigm on two datasets and four tasks: NYU-v2 and the taskonomy dataset for which we perform semantic segmentation, depth estimation, surface normal estimation, and edge detection.
\ac{MTML} achieves state-of-the-art results for three out of four tasks for the NYU-v2 dataset and two out of four for the taskonomy dataset.
In the taskonomy dataset, it was discovered that many pseudo-labeled segmentation masks lacked classes that were expected to be present in the ground truth; however, our \ac{MTML} approach was found to be effective in detecting these missing classes, delivering good qualitative results. 
While, quantitatively its performance was affected due to the presence of incorrect ground truth labels.
The the source code for reproducibility can be found at https://github.com/ricupa/MTML-learn-how-to-adapt-to-unseen-tasks.


\end{abstract}

\begin{IEEEkeywords}
Multi-task learning, meta learning, semantic segmentation, depth estimation, surface normal estimation
\end{IEEEkeywords}

\section{Introduction}
\label{sec:intro}
Multi-task learning (\ac{MTL}) involves learning many tasks in a single, combined network architecture \cite{MTL_rich}. This is in contrast to single task learning, which trains dedicated networks, one for each task.
The prime argument backing \ac{MTL} is that the knowledge absorbed by the network while learning one task may help to improve the performance on another task when they are trained together. 
However, in a multi-task setting, when there is a need to add a new task to the existing architecture, the new network has to be re-trained from scratch for the new set of tasks so that there is efficient knowledge transfer between the tasks, but, it leads to the loss of previously gained knowledge.
Fine-tuning the new task is another option, but there is a risk of overfitting \cite{overfitting}.
Meta learning \cite{hospedales2020meta, pmlr-v70-finn17a}, on the other hand, involves reusing information obtained during the learning of a task to quickly adapt to a new one.
The paradigm of meta-learning---also referred to as \emph{learning to learn} \cite{Thrun1998}---gathers experience by learning several homogeneous tasks (learning episodes) and utilizes the overall meta knowledge to enhance its future performances on yet unseen tasks. 
Meta-learning learns the distribution over the tasks rather than the specific tasks themselves. 
The latter are considered samples from this distribution, which implies that all tasks must be of similar nature or uni-modal \cite{hospedales2020meta}, \eg classification or regression. 
Similar tasks are referred to as homogeneous in this work, while those of a distinct kind are called heterogeneous.
In this study, we focus on the optimization based meta learning algorithms \cite{pmlr-v70-finn17a, Ravi2017OptimizationAA, reptile, meta_sgd,lee2018gradient}.
\begin{figure*}[t]
    \centering
    \includegraphics[width = \linewidth]{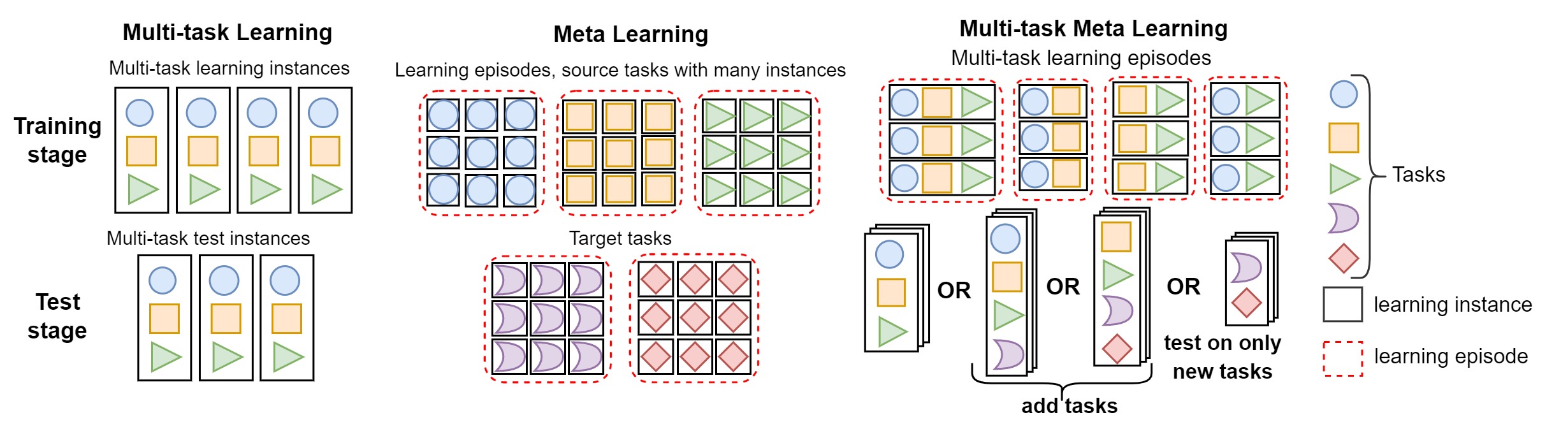}
    \captionof{figure}{An intuitive explanation of how `tasks' are introduced in MTL, meta learning, and MTML. In MTL, one learning instance has all the tasks, in meta learning one learning episode has multiple instances for one task and many episodes of similar but non-identical tasks are used for training. While in MTML, the multi-task learning episodes consist of many instances of all the combinations of the tasks.} 
    \vspace{-2em}
    \label{fig:simple_fig}    
\end{figure*}

This work introduces a \ac{MTML} paradigm, taking advantage from both multi-task and meta learning and constructed as follows:
the episodes used for meta training are assorted multi-task combinations, as illustrated in Fig. \ref{fig:simple_fig}, trained by employing the two-level meta optimization scheme introduced by MAML \cite{pmlr-v70-finn17a}.
The \ac{MTL} allows the joint learning of homogeneous as well as heterogeneous tasks. The latter is currently considered a limitation of meta learning \cite{hospedales2020meta}. 
In contrast, meta learning aids in quicker learning of an unseen task with fewer data samples, which is challenging for \ac{MTL}. 
As a result, \ac{MTML} is focused on learning how to learn unseen tasks in a multi-task setting.
The contributions of this work are:
\begin{enumerate}

\setlength\itemsep{0em}
    \item a new approach for creating multi-task episodes with heterogeneous tasks essential for the meta training phase;
    \item a learning mechanism called \ac{MTML} enabling faster training of new tasks and better performance on all of the tasks, when meta learning is introduced in a \ac{MTL} framework ;
    \item extensive comparative performance analysis of single-task, \ac{MTL}, and \ac{MTML} learning paradigms on two publicly available datasets.
\end{enumerate}

This article is organized as follows; Section 2, discusses the related works in multi-task learning and also multi-task meta learning.  Section 3, introduces \ac{MTL}, and meta learning, while Section 4 details the formulation of \ac{MTML}. The experimental setup is given in Section 5. Section 6 contains a comprehensive analysis of the results. The reasons behind the unsatisfactory performance of semantic segmentation task are discussed in Section 7. At last, Section 8 draws the conclusion and opens up to future work.

\section{Related work}


In this section, we highlight significant research studies that claim to combine \ac{MTL} and meta learning.
\cite{liuMTL2018, tarunesh2021meta, lee-etal-2021-generating} apply meta learning optimization in a multi-task scenario to achieve good generalization for unseen data sets on the same tasks.
In \cite{liuMTL2018} for efficient communication between two tasks in \ac{MTL}, a gradient passing mechanism is proposed which has similar traits to gradient based meta learning.   
A meta learning approach is followed for sharing parameters across multiple tasks and languages in \cite{tarunesh2021meta}, the model is trained on various task-language pairs rather than training all the tasks simultaneously.
In \cite{lee-etal-2021-generating}, the tasks of dialogue generation give a context and persona information is learned for multiple personas following meta optimization.
Here the multiple learning episodes (persona information) are considered as multiple tasks.
Additionally, \cite{tian2019hierarchical, ghadirzadeh2021bayesian, bronskill2020tasknorm} use multiple tasks for the multiple training episodes in meta learning and hence tagged their paradigm \emph{multi-task meta learning}.
Similarly, in \cite{multi_meta} it is theoretically and empirically  proved that \ac{MTL} is a computationally efficient alternative to gradient based meta learning algorithm, as for an adequately deep network, the learned predictive functions of \ac{MTL} and meta learning are very similar.

Several studies on \ac{MTL} \cite{arch1, misra2016cross, cascade, gao2019nddr, xu2018pad, pram_mtl} place a strong emphasis on network architecture design to improve task performance.
Apart form them, the authors in \cite{zamir2018taskonomy} propose to determine the task transfer relationships between 26 tasks in order to learn to group tasks for \ac{MTL}.
Other \cite{james2018task, lan2019meta, achille2019task2vec, L2MT} discuss task embedding primarily for meta learning as a means to learn task relationships.
\cite{sun2020adashare, guo2020learning} employ neural architecture search in a multi-task setting to reduce the number of parameters. 

Most of the works in the literature related to \ac{MTL} concentrate on making it more efficient in terms of its performance, number of parameters, generalization to unseen data, etc., by adopting new architectures, learning better task grouping, soft parameter sharing, neural architecture search, and integration with other algorithms. 
Adding to the list, this work puts together meta learning concept to enable the addition of an unseen task to an \ac{MTL} architecture, resulting in faster training and better generalization of the newly added task compared to single-task learning.

\section{Background} \label{formulation}
\begin{figure}[ht]
  \centering
  \subfloat[a][Multi-task learning]{\includegraphics[width=0.47\textwidth]{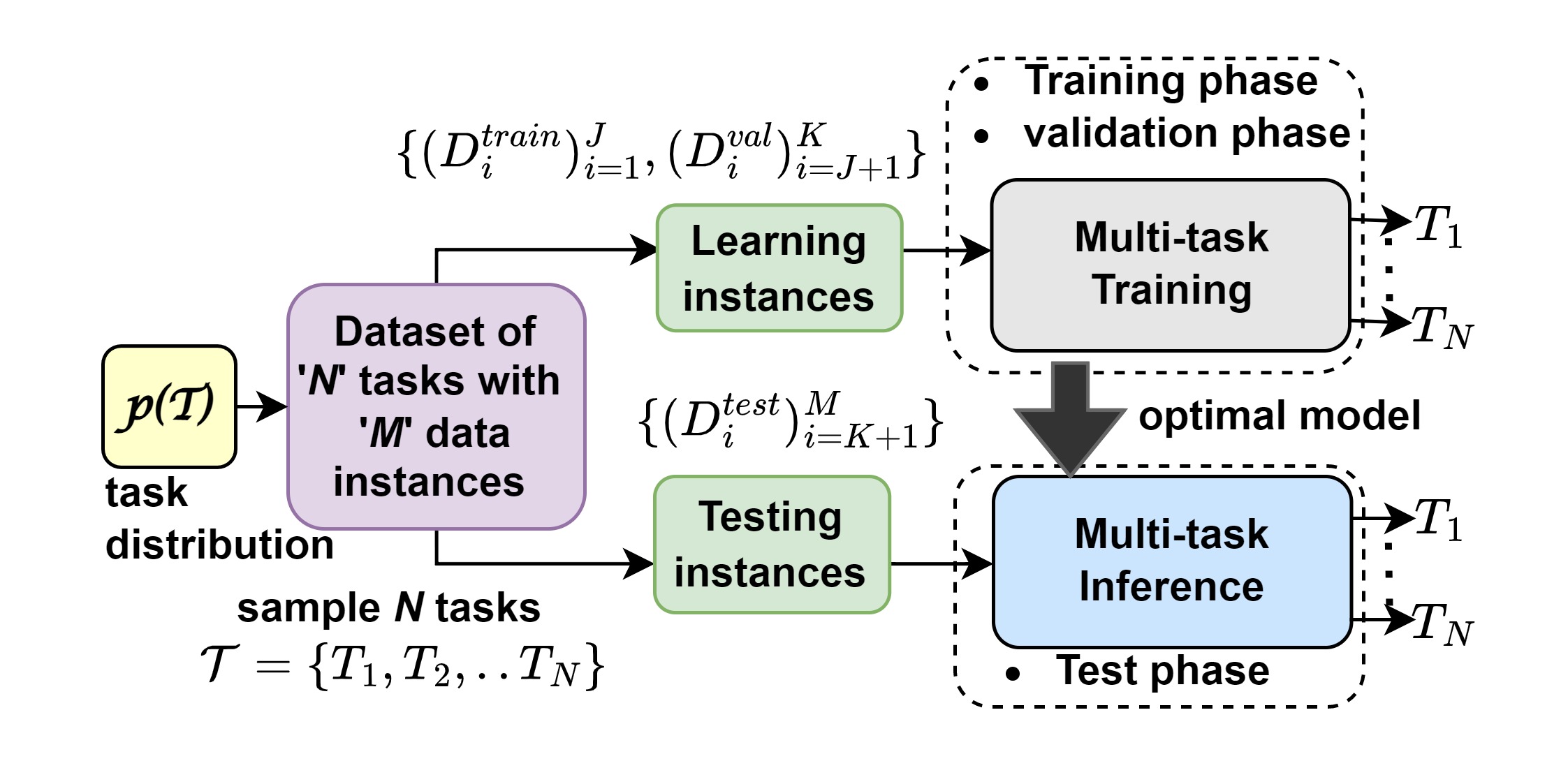} \label{fig:MTL}} \\
  \subfloat[b][Meta learning (MAML \cite{pmlr-v70-finn17a})]{\includegraphics[width=0.47\textwidth]{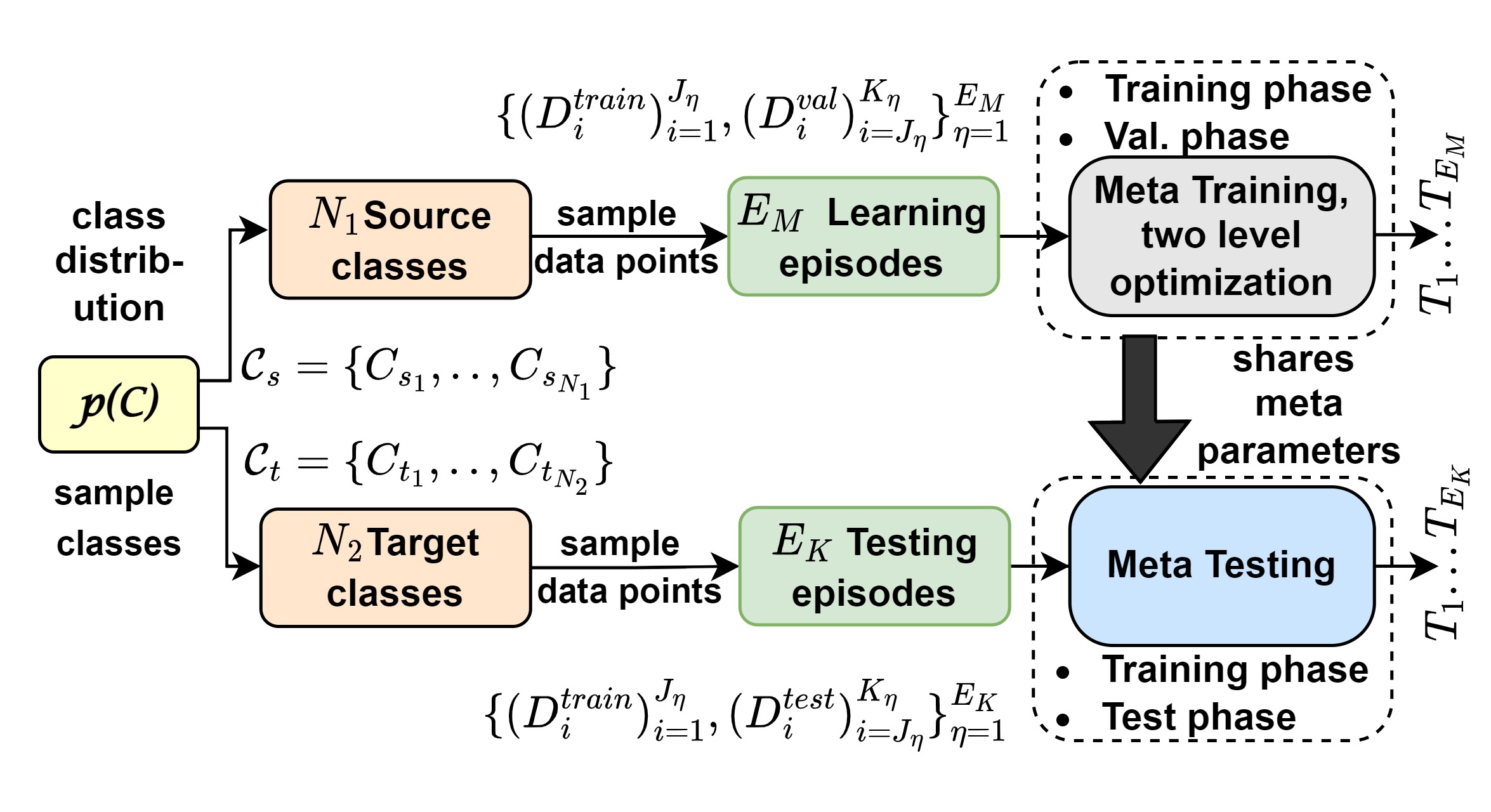} \label{fig:meta}}
  \\
  \subfloat[c][Multi-task meta learning]{\includegraphics[width=0.47\textwidth]{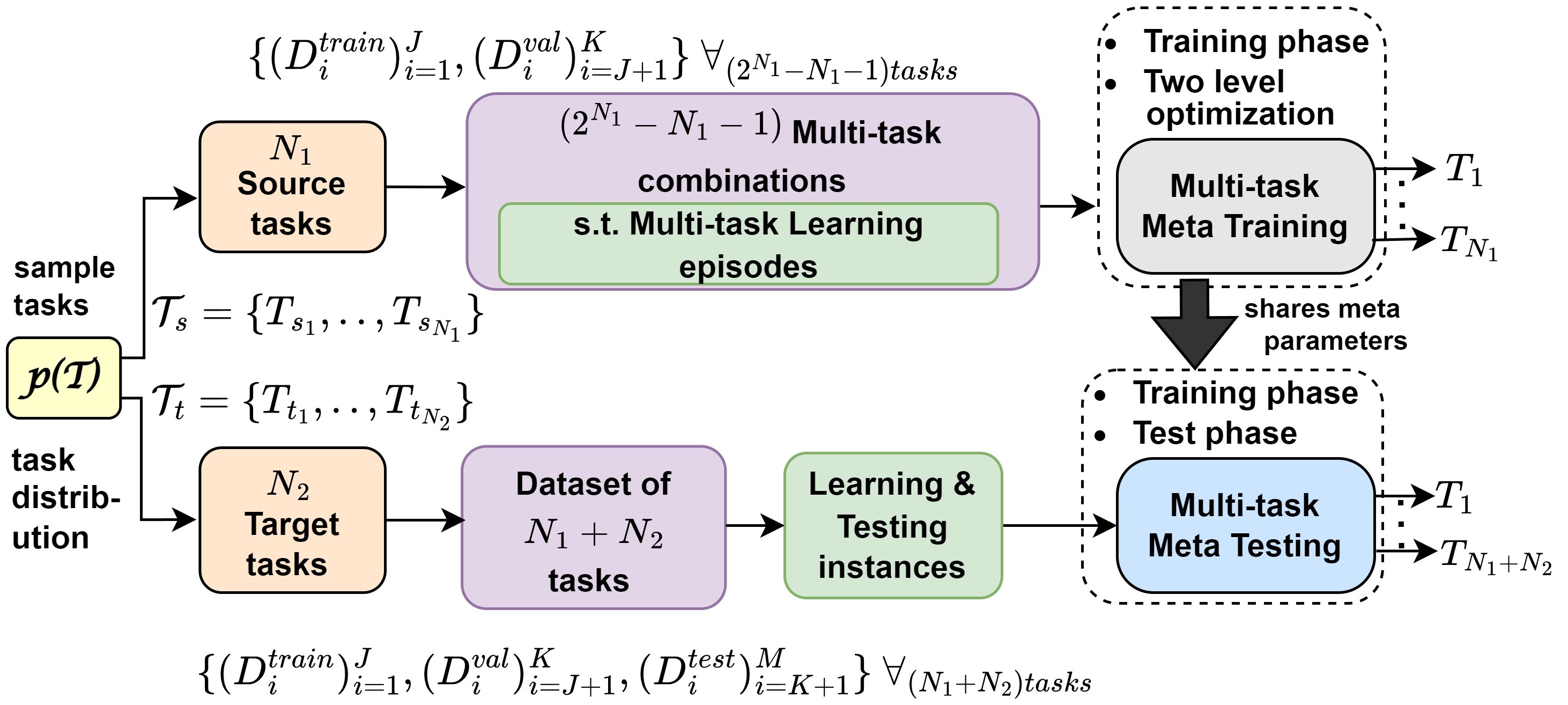} \label{fig:MTL_meta}}
  \caption{Block diagram illustrating the formulation of the learning paradigms} \label{fig:block_diag}
 \vspace{-0.7cm}
\end{figure}

\textbf{Multi-task learning}: a learning paradigm aiming to train multiple tasks together.
The shared representations help to enhance the learning capabilities of all the tasks compared to training them individually \cite{maurer2016benefit}.
As illustrated in Fig. \ref{fig:MTL}, $N$ non-identical---but related---tasks are sampled from a task distribution, $\mathcal{T} = \{T_1, T_2,...,T_N\}$, where $T_i$ is the $i^{th}$ task.
The data set for all the tasks represented by $\mathcal{D}$ is split into $J$ train, $K-J$ validation, and $M-K$ test instances, such that  $\mathcal{D} = \{(D_i^{train})_{i=1}^J, (D_i^{val})_{i = J+1}^K,(D_i^{test})_{i = K+1}^M\}$, here $D_i$ represents the dataset for $i^{th}$ task.
A multi-task network architecture is trained using the training data $(D_i^{train})_{i=1}^J$.
The objective is to minimize the combined loss $\mathcal{L}$, by optimizing the network parameters $\omega=\{(\omega_i)_{i=1}^N\}$, such that
\begin{equation}
\omega^* = \min_{\omega} \sum_{i=1}^N \mathcal{L}_i(\omega_i, (D^{train})_i)\enspace.
\end{equation}
$\{D^{val}_i\}_{i=J+1}^K$ is used to evaluate the performance of the multi-task model during the training process, thereby assures generalization of the model to data instances not used during training.
The optimal parameters $\omega^*$ are used in the inference on the unseen test data $\{D^{test}_i\}_{i=K+1}^M$ to report model performance. 
     
   

\textbf{Meta learning}:
In a few-shot learning framework (a species of meta learning), every $n$-way, $k$-shot learning episode (\ie $n$ classes and $k$ data samples of each class) is sampled from a base data set using two-step episodic sampling, as shown in Fig. \ref{fig:meta}.
Here classes are discrete output variables also known as labels.
First, one samples the episode classes from the class distribution, organized into source classes $\mathcal{C}_s=\{C_{s_1}, C_{s_2},...,C_{s_{N_1}}\}$ and target classes $\mathcal{C}_t=\{C_{t_1}, C_{t_2},.., C_{t_{N_2}}\}$; where $N_1$ and $N_2$ represent the number of source and target classes, respectively, and $\mathcal{C}_s \cap \mathcal{C}_t=\emptyset$.
Second, one samples an episode ($k$ data points) from the data set based on the classes sampled in the previous step. 
Consider an example of 2-way $k$-shot learning: each learning episode ( \ie task) will contain the training instances of two classes, say $ \mathcal{T}_1=\{C_{s_1}, C_{s_2}\} ..... \mathcal{T}_{E_M}=\{C_{s_{N_1-1}}, C_{s_{N_1}}\}$, thereby creating $E_M$ learning episodes (\emph{tasks}) for meta training.
The meta learners iterate over the learning episodes intending to carry out the process of \emph{learning to learn} \cite{hospedales2020meta}. 
It employs a two-step optimization \cite{pmlr-v70-finn17a} for all the learning episodes. 
First, it follows task-specific learning by optimizing task-specific parameters $\{\omega_\eta\}_{\eta=1}^{E_M}$ given the meta parameters $\theta^{(p)}$ of the $p$th iteration:
\begin{align}\label{step1}
    \omega^{(p+1)}_{\eta}(\theta^{(p)})  = \argmin_{\omega} \mathcal{L}_{\eta}(\omega, \theta^{(p)}, (D^{train})_{\eta})
\end{align}
Here, $\mathcal{L}_{\eta}$ represents the loss for the $\eta$th learning episode, $(D^{train})_{\eta}$ are the training data set. The meta parameters \ie $\theta$ are often referred to as \emph{meta knowledge} or \emph{knowledge across tasks} \cite{hospedales2020meta}.
In Fig. \ref{fig:meta}, a data instance is indexed by $i$, and an episode is indexed by $\eta$. 
The second step corresponds to multiple task learning: at this meta stage, the aim is to reduce the meta loss $\mathcal{L}^{meta}$---using the unseen validation instances $(D^{val})_{\eta}$---by optimizing the $\theta$ given the task parameters of the $(p+1)$th iteration:
\begin{equation}\label{step2}
    \theta^{(p+1)}(\omega^{(p+1)}) = \argmin_\theta \sum_{\eta = 1}^{E_M} \mathcal{L}^{meta}(\omega^{(p+1)}_{\eta}, \theta, (D^{val})_{\eta})
\end{equation}
Iterating between \eqref{step1} and \eqref{step2} would result in an optimal meta learner, \ie, $\theta^{(p)}\to\theta^\ast$.
During meta testing (adaptation  stage) the meta knowledge $\theta^*$ is used as initial parameters for learning new, unseen tasks (episodes), say, $\mathcal{T}_1=\{C_{t_1}, C_{t_2}\}......\mathcal{T}_{E_K}=\{C_{t_{N_1-1}}, C_{t_{N_1}}\}$. 
Therefore, the test tasks are fine-tuned on the model using meta parameters, which help achieve the best performance for the new tasks in few gradient steps.  

\section{Formulation of Multi-task meta learning (\ac{MTML})}
In this work, the multi-task architecture is used along with the bi-level meta optimization to establish \ac{MTML}. 
Particularly, an optimization-based meta learning approach \cite{huisman2021survey}, recognized as \ac{MAML} \cite{pmlr-v70-finn17a} is adopted, which performs a two-level gradient descent optimization compatible with any model. 
The Fig. \ref{fig:MTL_meta}, illustrates the formulation of \ac{MTML}. 
The source and target tasks are sampled from a distribution of tasks $p(\mathcal{T})$, given by $\mathcal{T}_s = \{T_{s_1},T_{s_2},..,T_{s_{N_1}}\}$ and $\mathcal{T}_t = \{T_{t_1},T_{t_2},..,T_{t_{N_2}}\}$, respectively. 
\textit{Multi-task learning episodes} are created from the source tasks analogous to meta learning, but since the nature of the tasks can be heterogeneous, the classical meta learning approach of merely sampling the source classes (or labels) is insufficient. 
Therefore, using the power set of the source task set $2^{{\mathcal T}_s}$, after excluding the singletons and the empty set, one has $2^{N_1} - N_1 - 1$ multi-task combinations of the source classes that can be used as multi-task learning episodes.
These multi-task episodes are used to train the network using the two-level meta optimization, discussed in equations \eqref{step1} and \eqref{step2}, \ie the multi-task meta training stage.
New unseen tasks can now be introduced as target tasks in the meta testing stage.
Either all source and target tasks or only the target tasks (as required) are then fine-tuned in the meta testing stage, which is similar to training in \ac{MTL}, except it utilizes the meta parameters from the  multi-task meta training stage.
The task heads (introduced in Section \ref{sec:exp} under network architecture) which are the task-specific (un-shared) layers, are fine-tuned if the purpose is to solely improve the target tasks.

Similar to $n$-way $k$-shot meta learning, \ac{MTML} can be considered as $N$-task $many$-shot learning. 
Here $N$ is the sum of the source and target tasks.
It should be emphasized that, the number of multi-task learning episodes exponentially rises with $N$, as a result the number of multi- task training episodes also increase.
The investigation of the effect of the number of tasks on MTML's performance is beyond the scope of this work.
Although, it is required that $N \ge 3$ to generate enough training episodes for the meta-training stage.
If $N = 2$ there will be only one learning episode, and which is insufficient for meta training.


\vspace{-0.2cm}
\section{Experimental setup} \label{sec:exp}
\setlength{\tabcolsep}{2pt}
\renewcommand{\arraystretch}{1.5}
\begin{figure*}[ht]
\begin{center}

\begin{tabular}{c c c c c c| c c c c c}
\hline
\multirow[b]{ 2}{*}{\rotatebox[origin=t]{90}{Models}} & Input &  \multicolumn{4}{c}{\textbf{Taskonomy}} & Input & \multicolumn{4}{c}{\textbf{NYU}}\\ \cmidrule(lr){3-6} \cmidrule(lr){8-11}
& Image & Seg. & Depth & SN & Edge & Image & Seg. & Depth & SN & Edge\\
\hline

\footnotesize{\rotatebox[origin=c]{90}{GT}} & \multirow[b]{ 4}{*}{\adjustimage{height=1.5cm,valign=m}{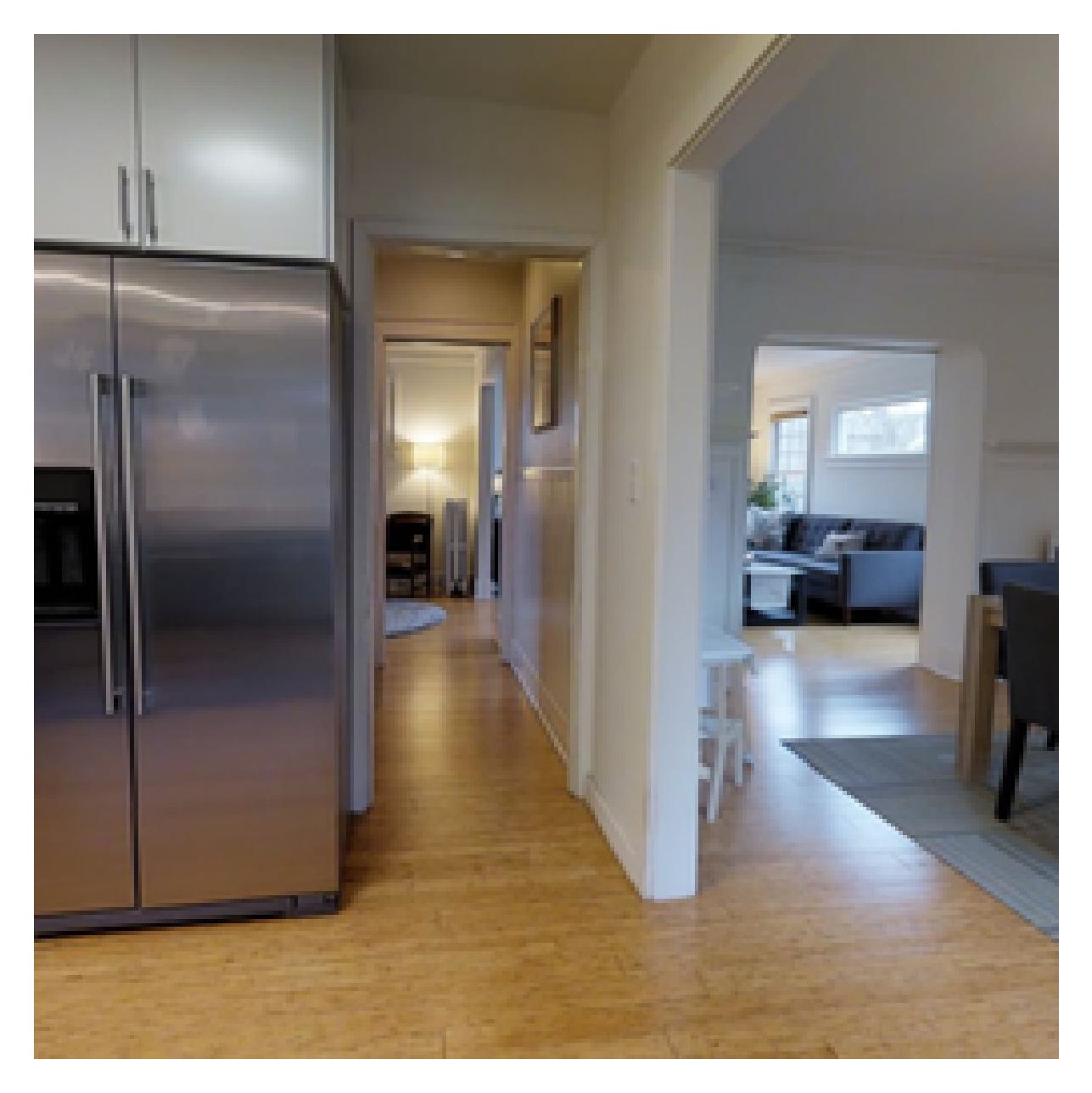}} &
\adjustimage{height=1.46cm,valign=m}{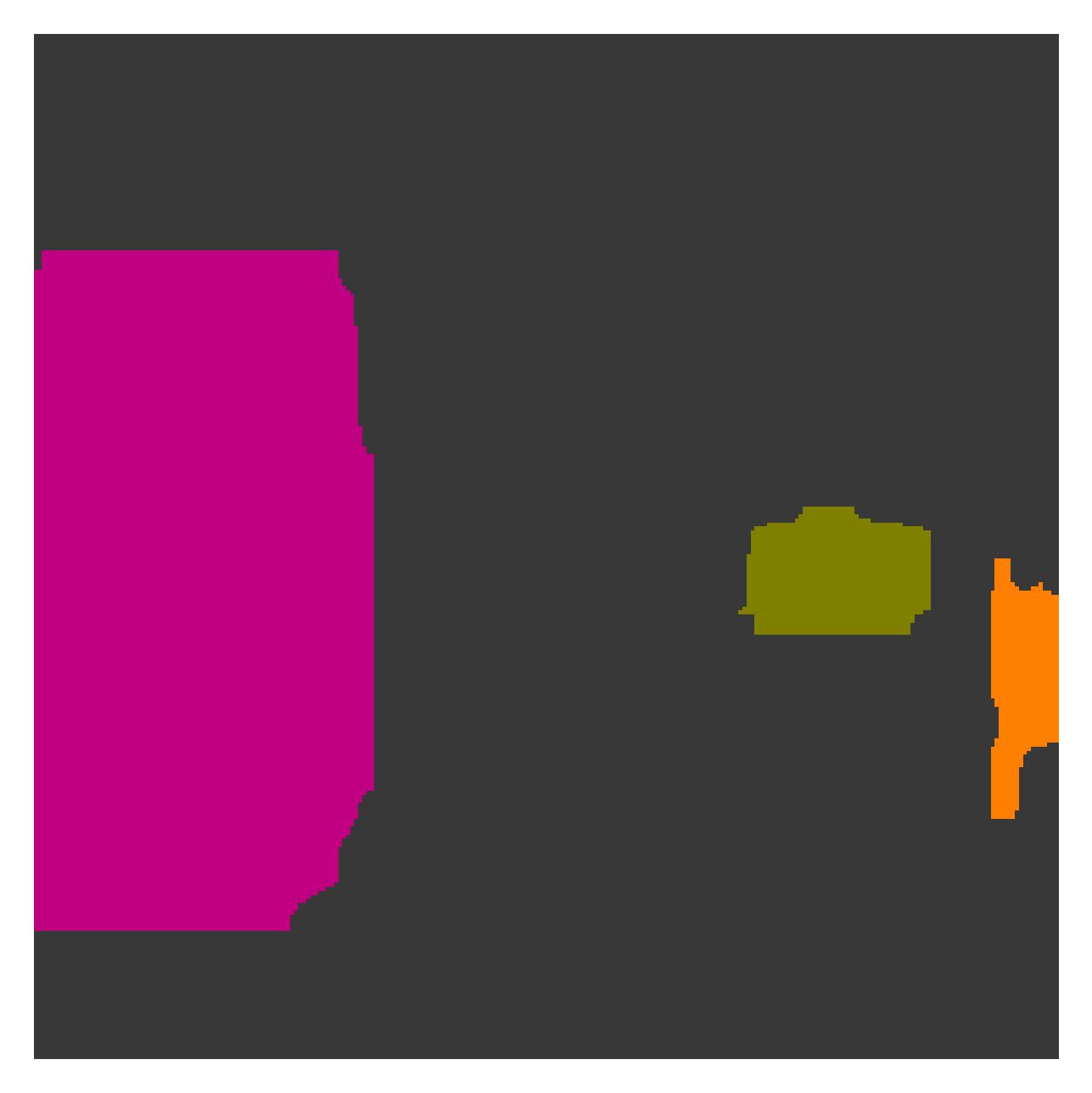} & \adjustimage{height=1.46cm,valign=m}{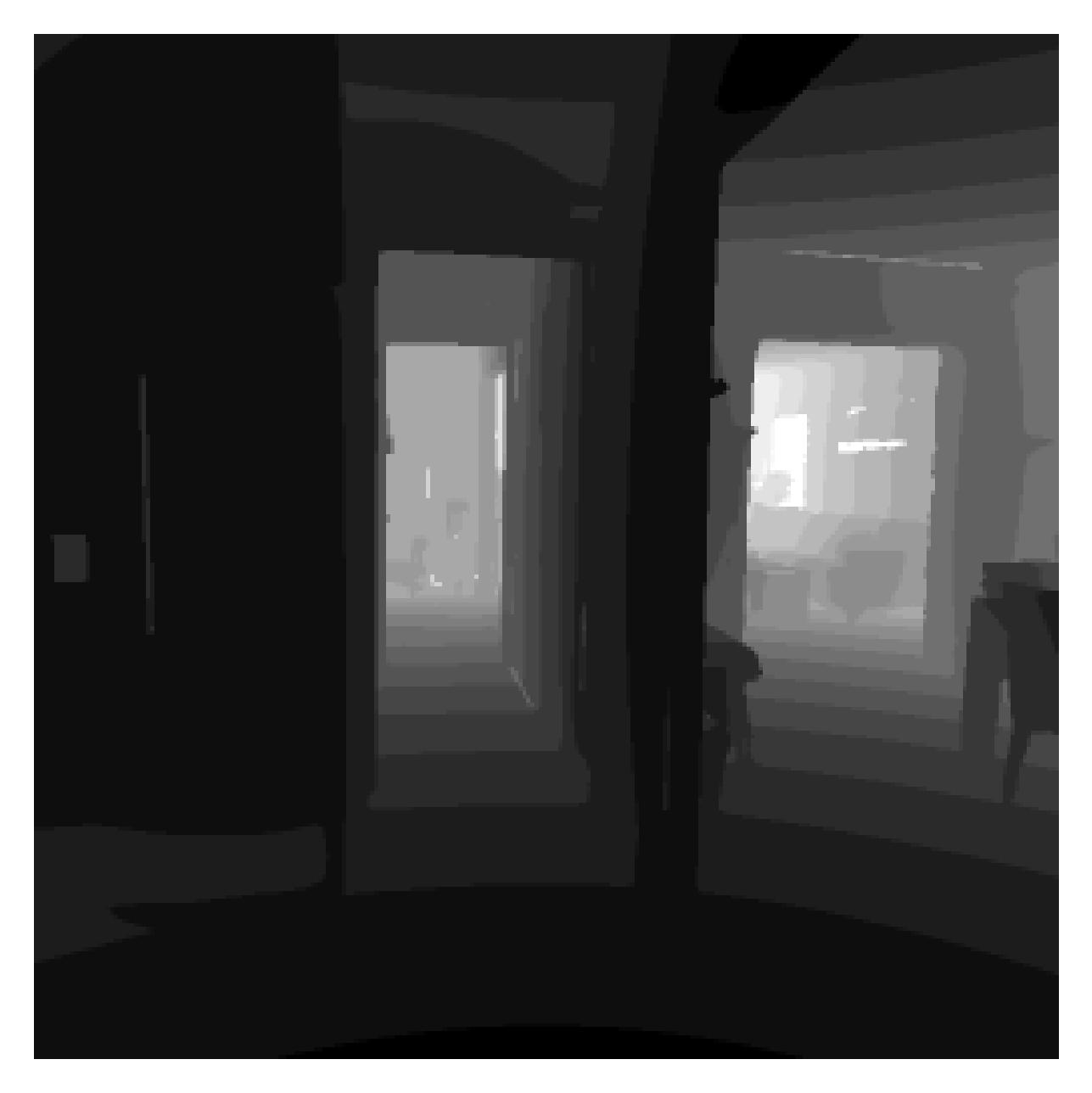} & \adjustimage{height=1.46cm,valign=m}{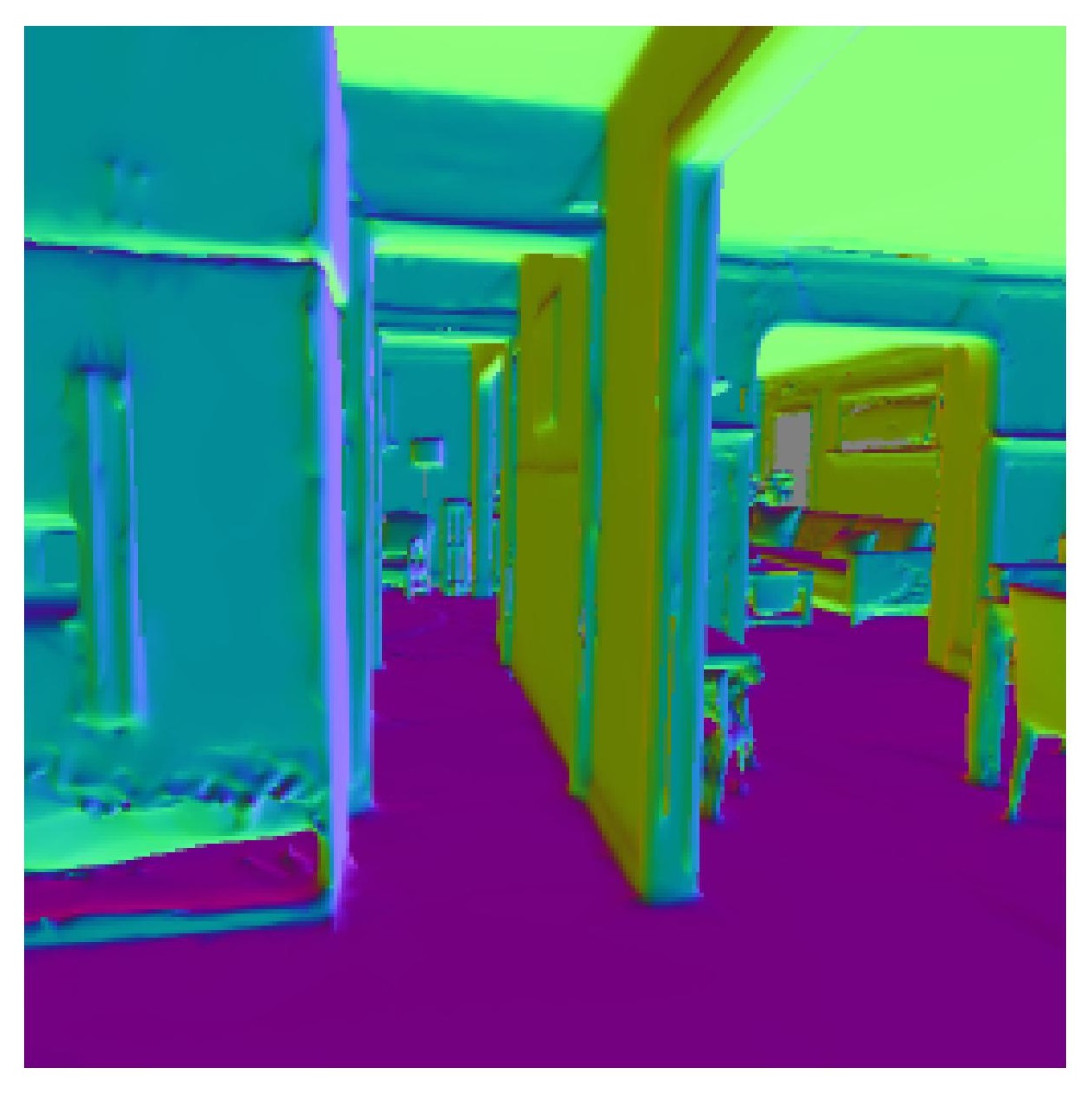} & \adjustimage{height=1.46cm,valign=m}{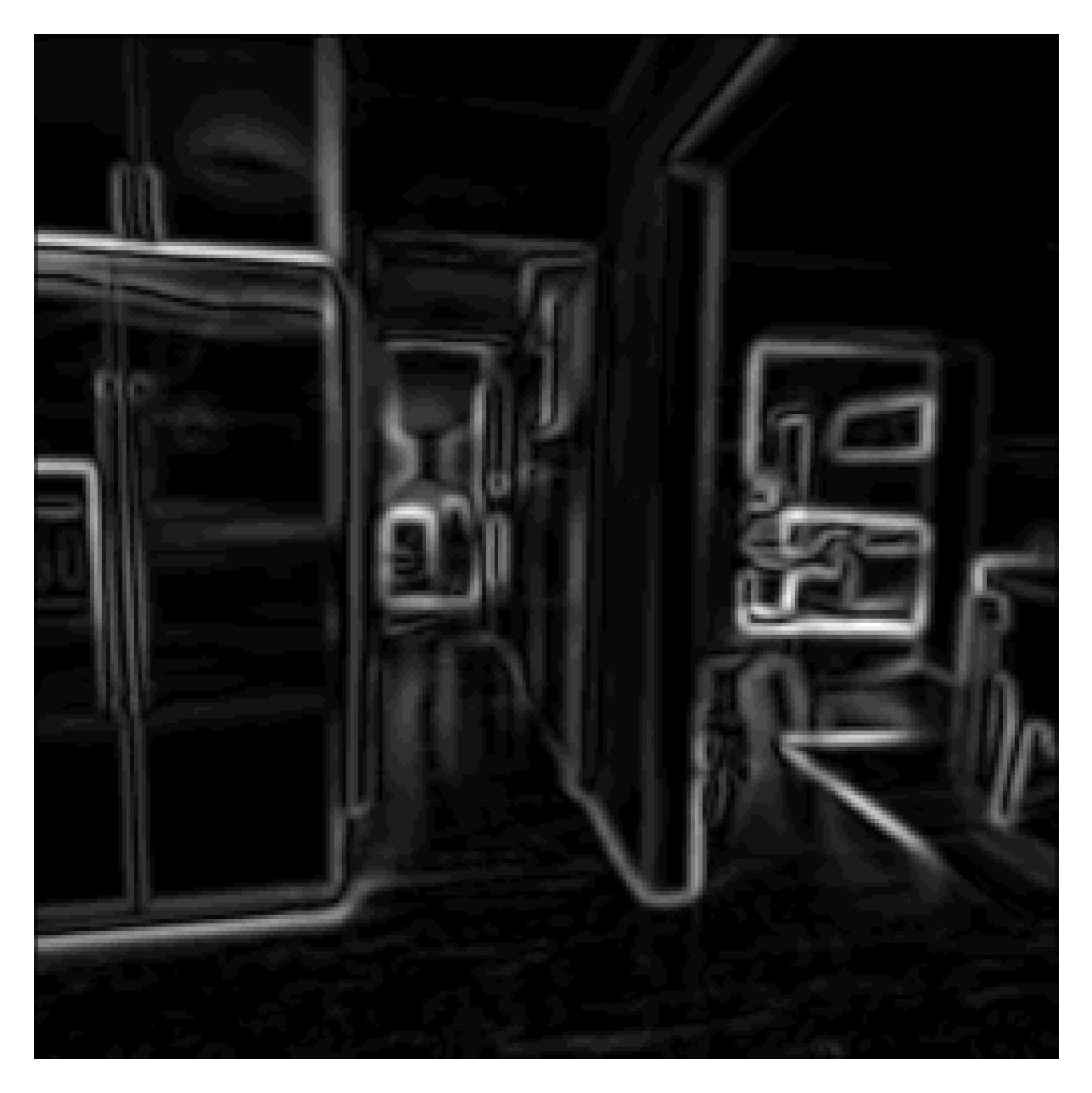} &
\multirow[b]{ 4}{*}{\adjustimage{height=1.5cm,valign=m}{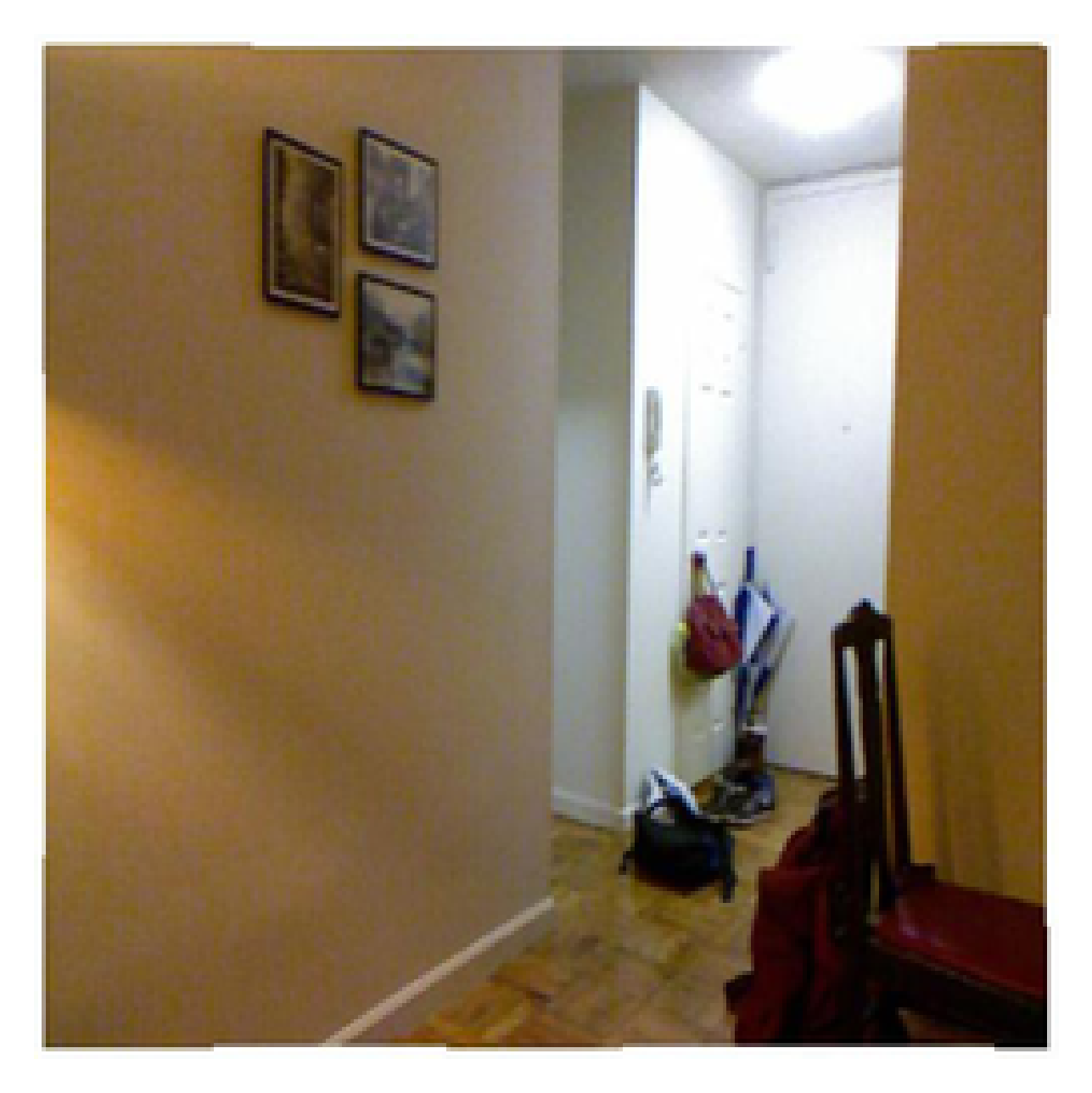}}&
\adjustimage{height=1.5cm,valign=m}{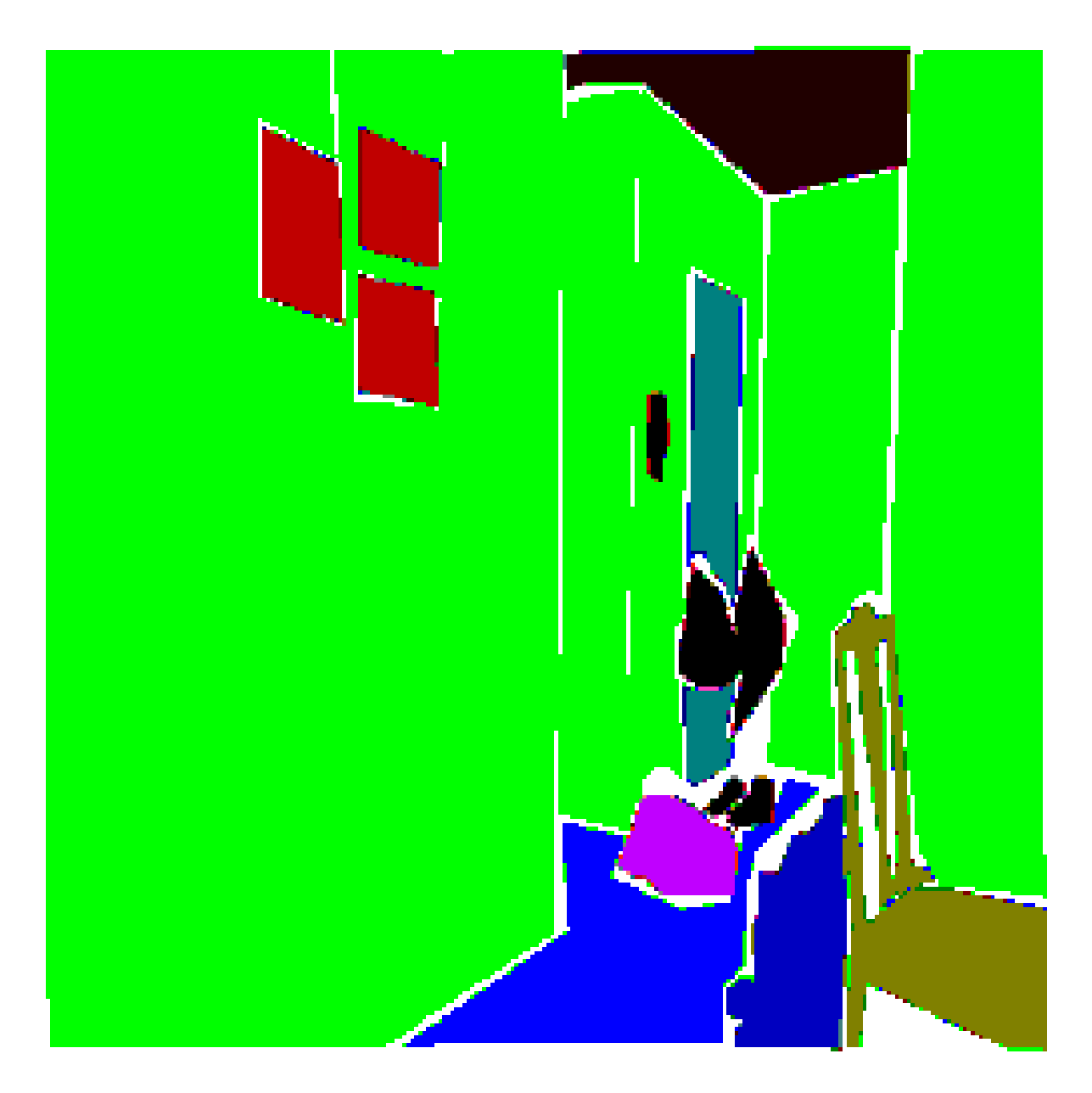} & 
\adjustimage{height=1.5cm,valign=m}{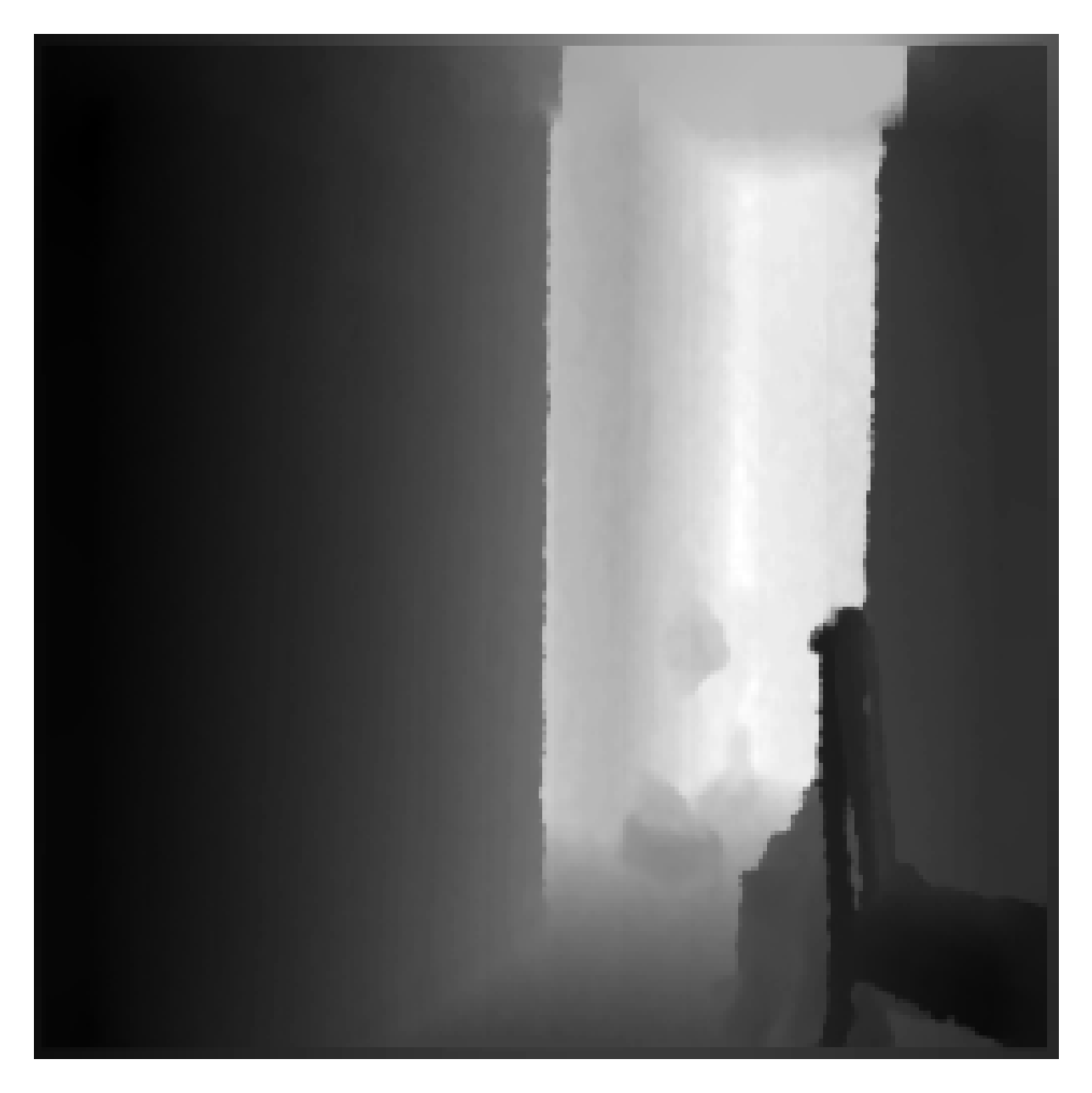} & \adjustimage{height=1.5cm,valign=m}{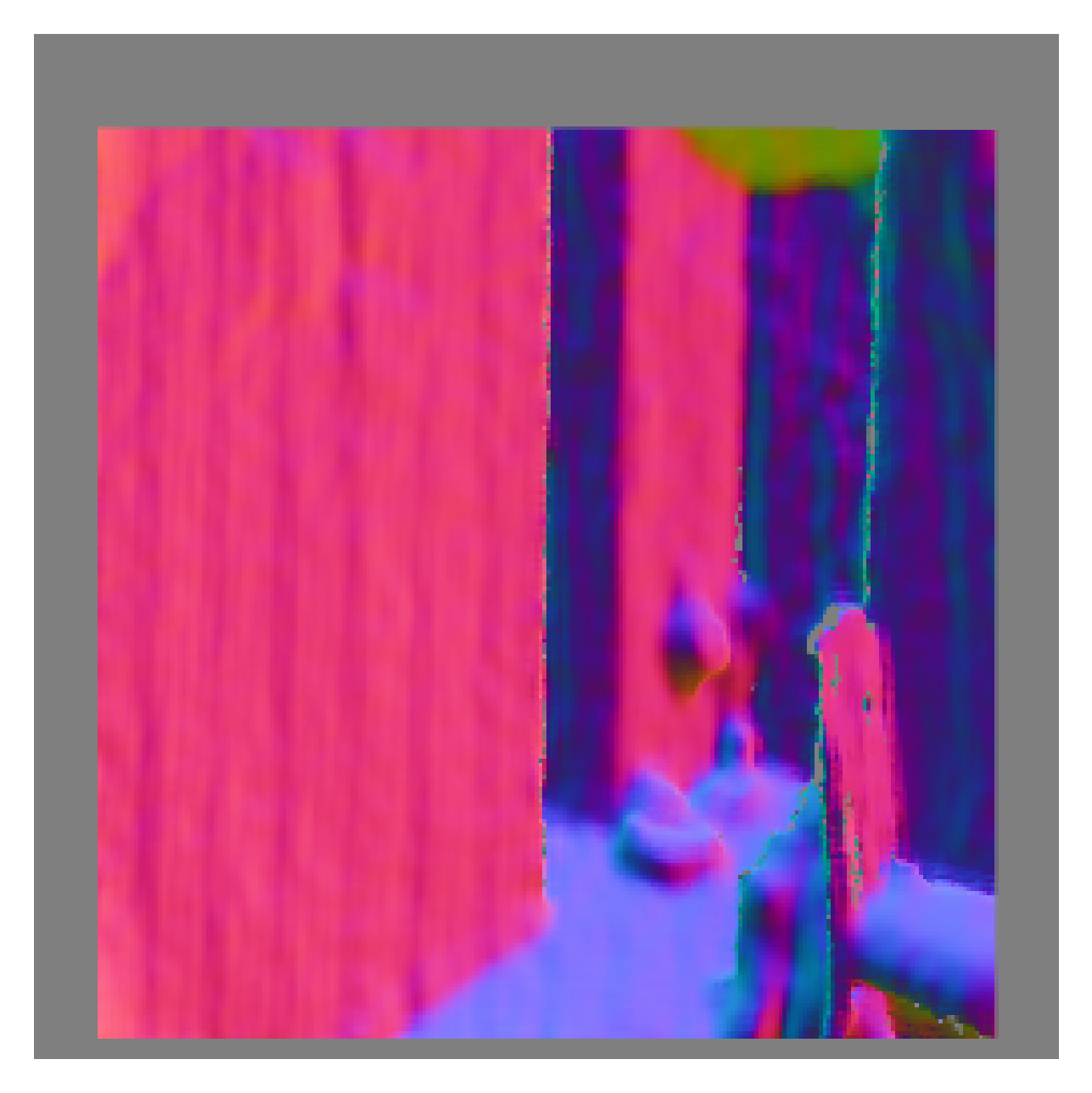}& \adjustimage{height=1.5cm,valign=m}{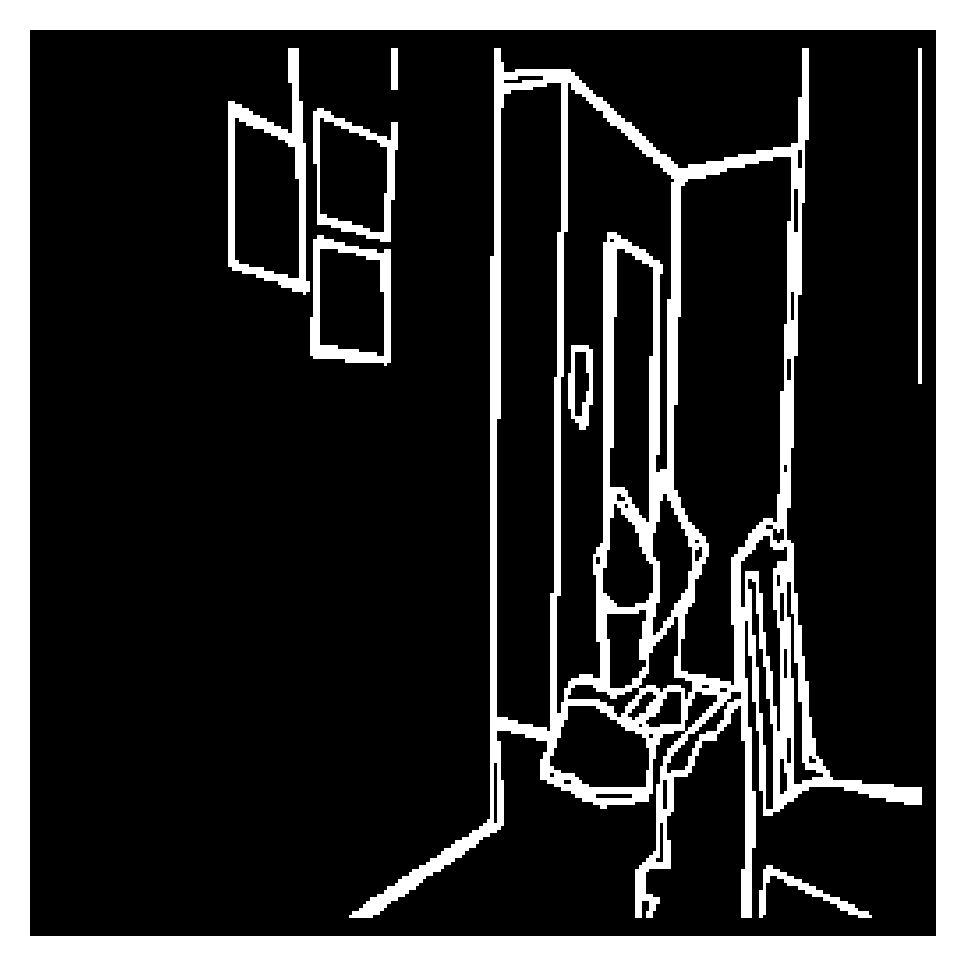}\\
\footnotesize{\rotatebox[origin=c]{90}{Single}} & &
\adjustimage{height=1.5cm,valign=m}{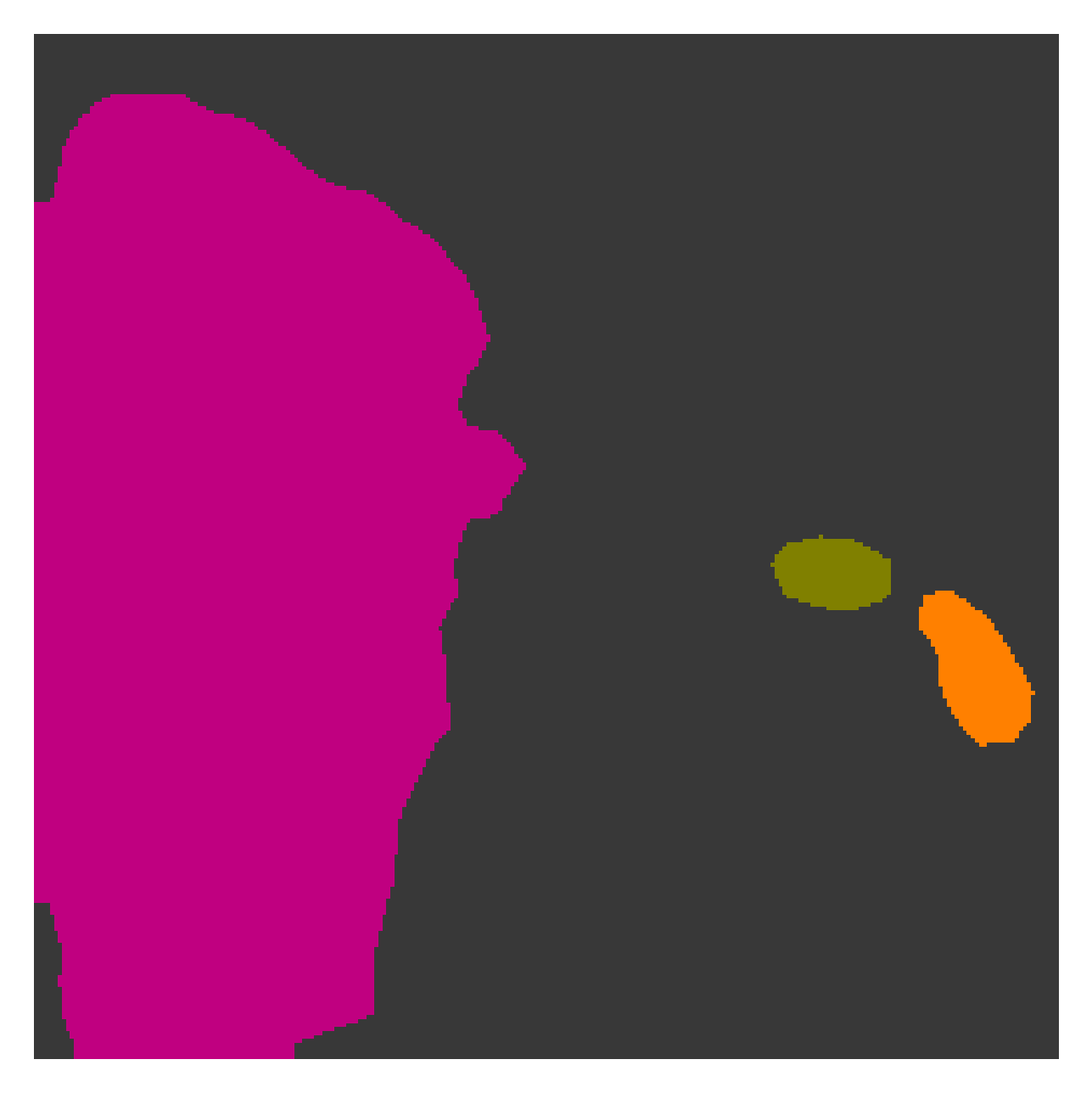} & \adjustimage{height=1.5cm,valign=m}{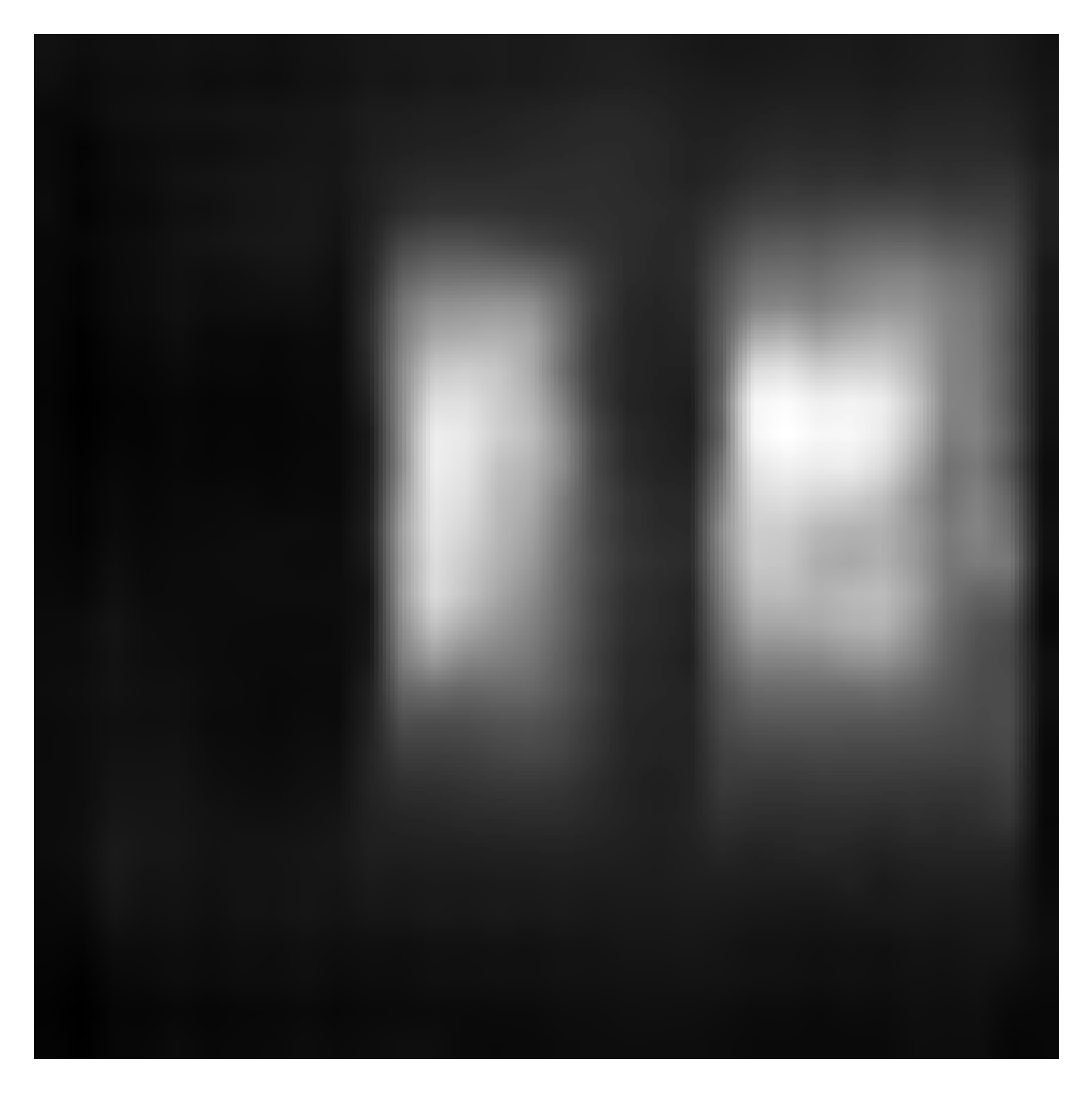} & \adjustimage{height=1.5cm,valign=m}{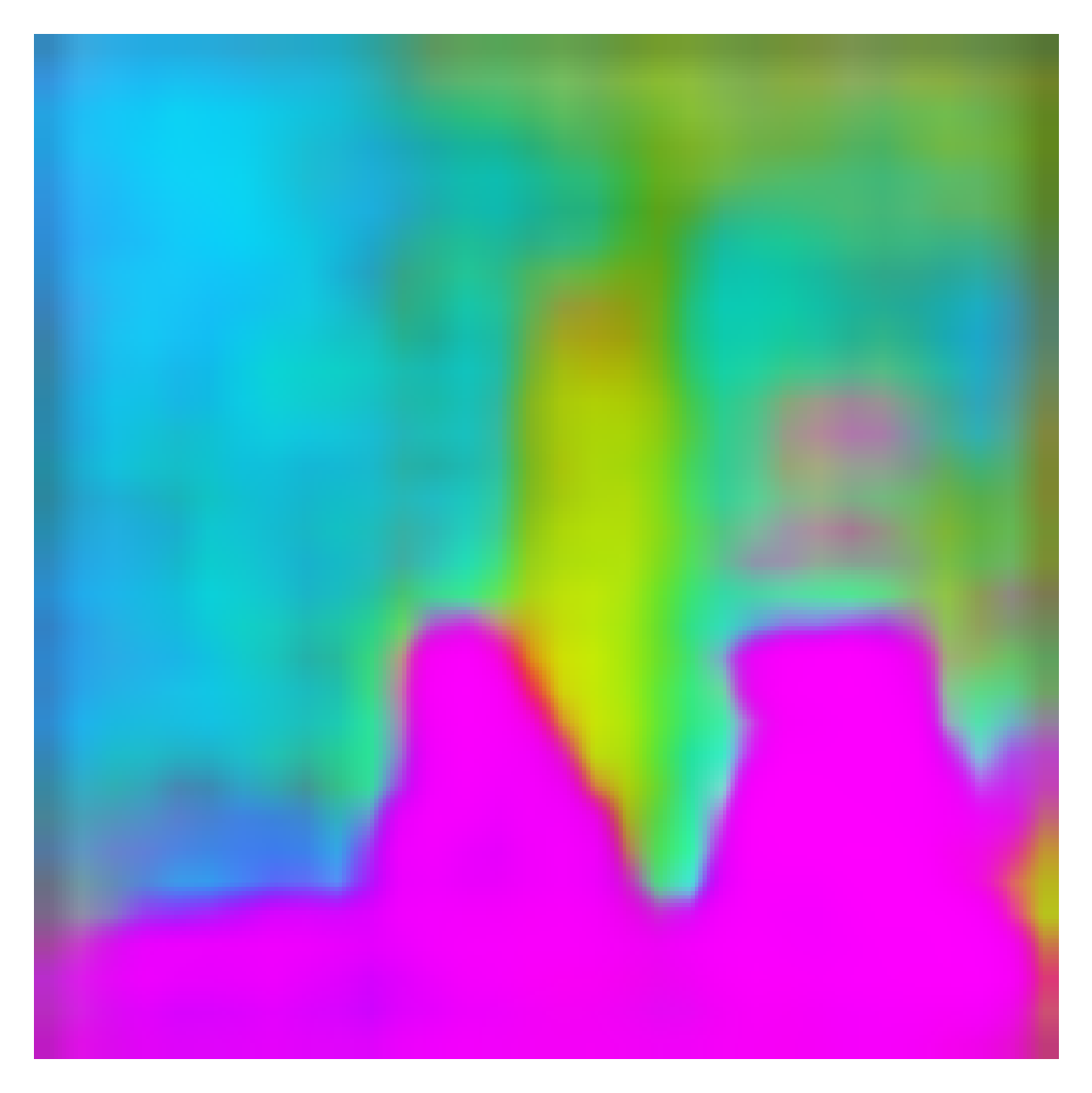} & \adjustimage{height=1.5cm,valign=m}{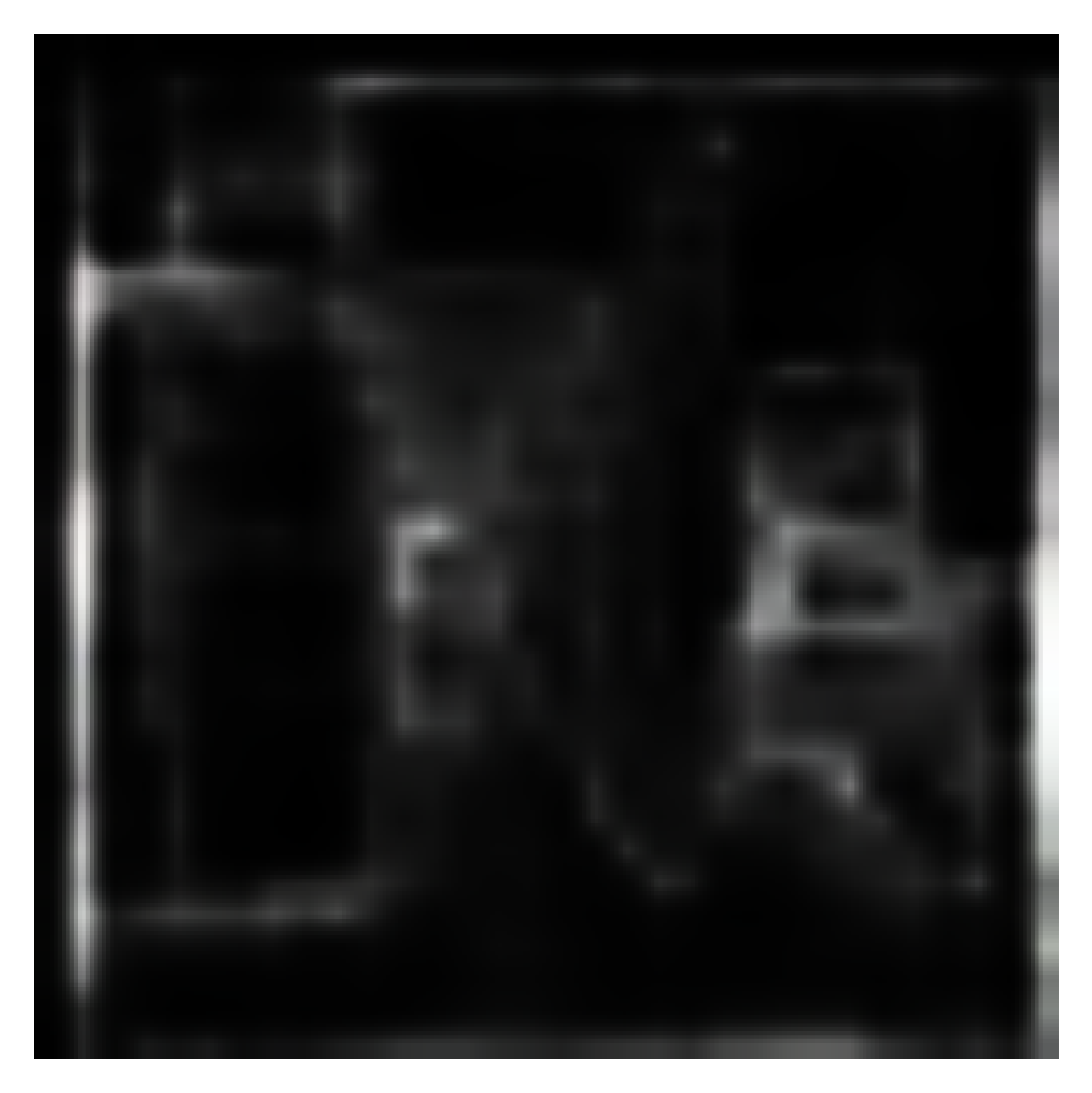} &
&
\adjustimage{height=1.5cm,valign=m}{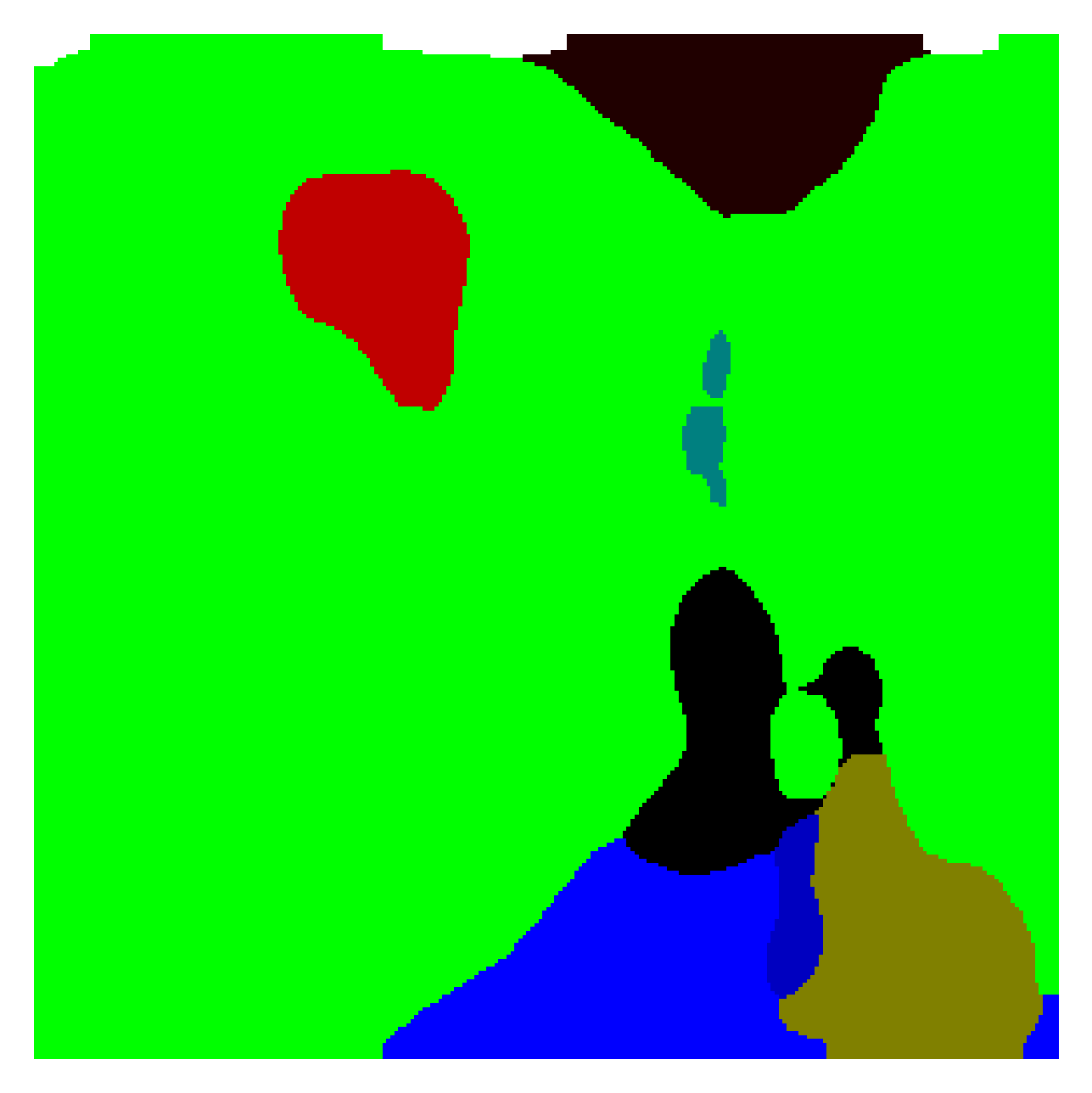} & 
\adjustimage{height=1.5cm,valign=m}{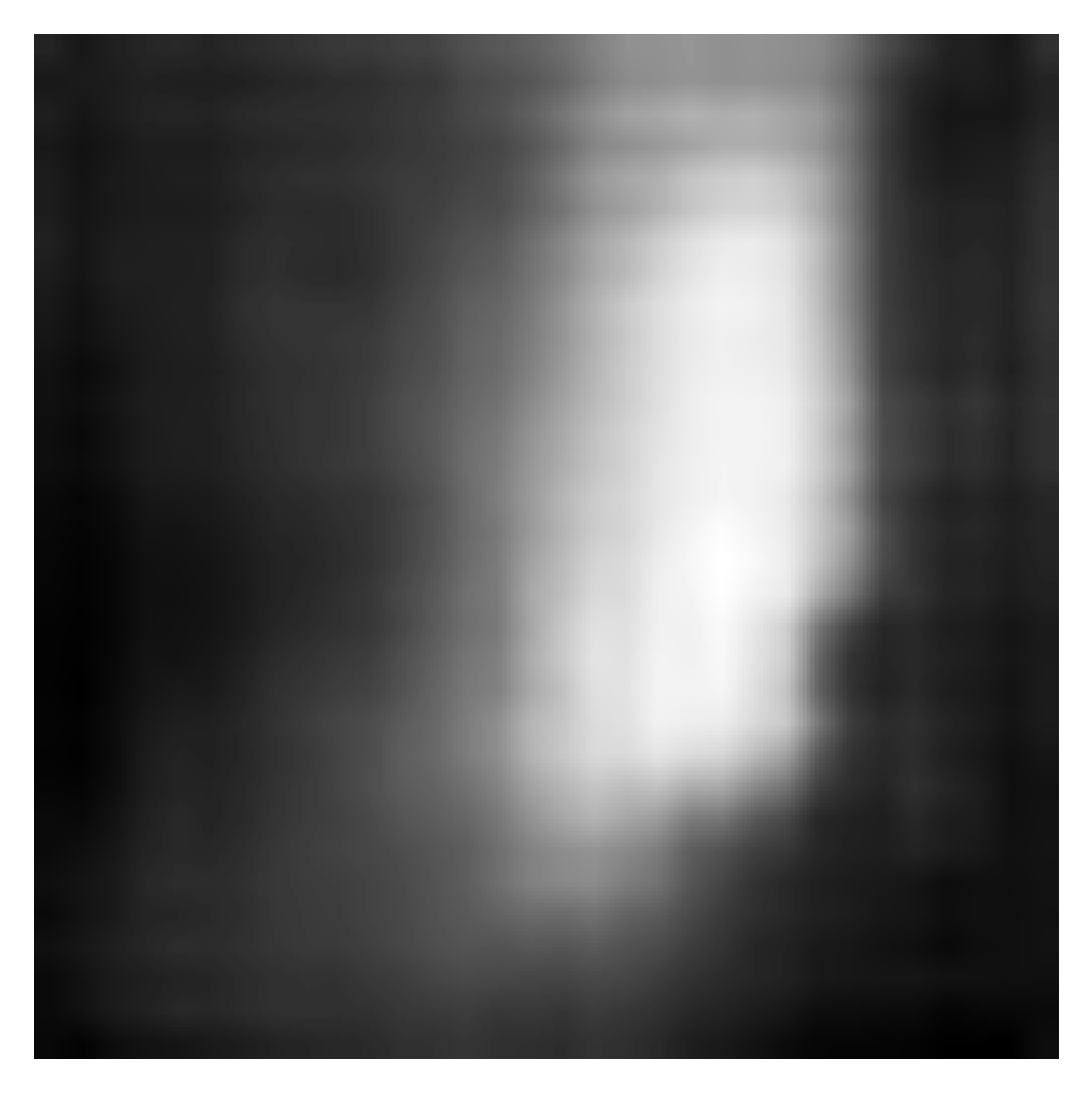} & \adjustimage{height=1.5cm,valign=m}{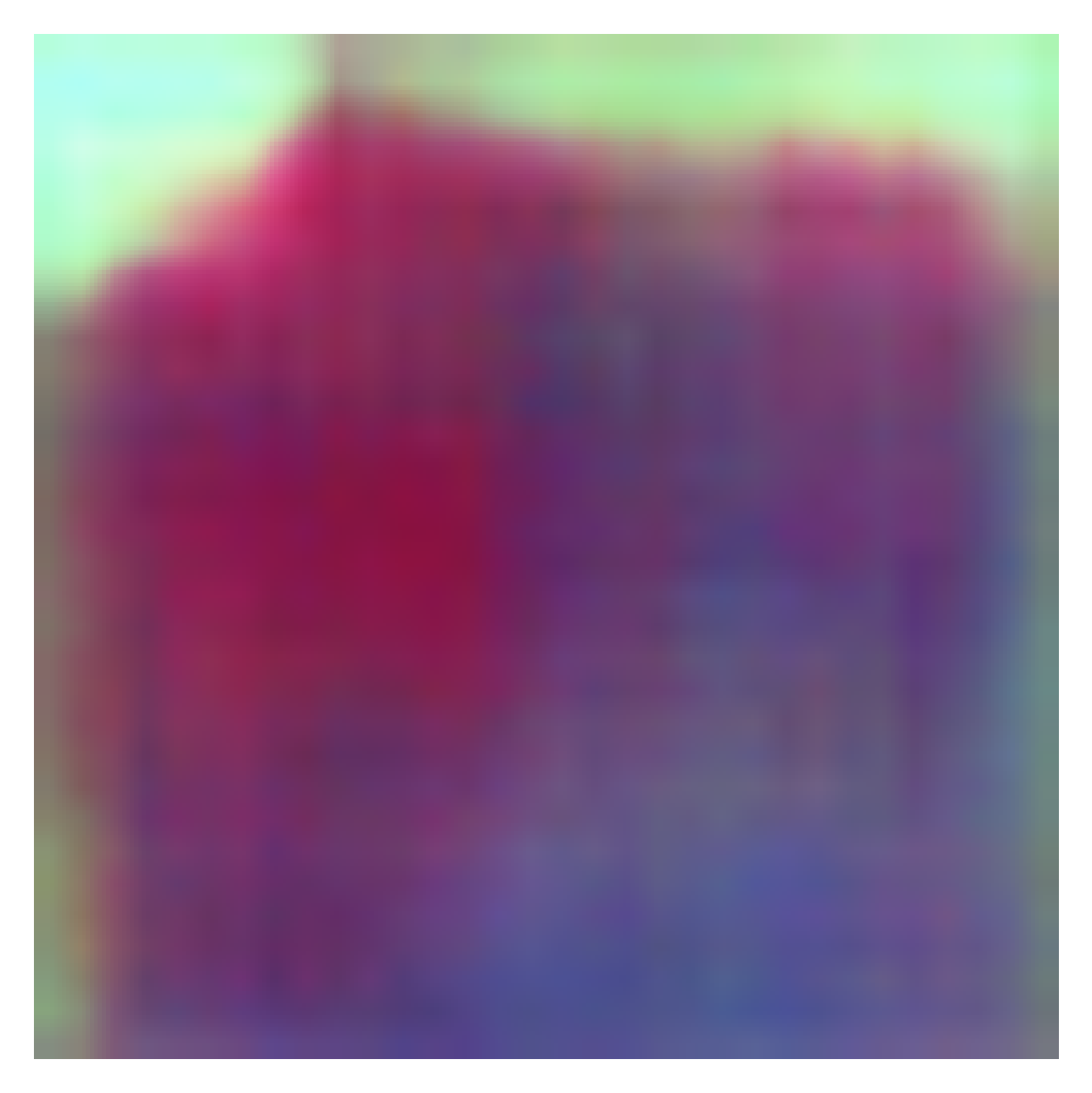}& \adjustimage{height=1.5cm,valign=m}{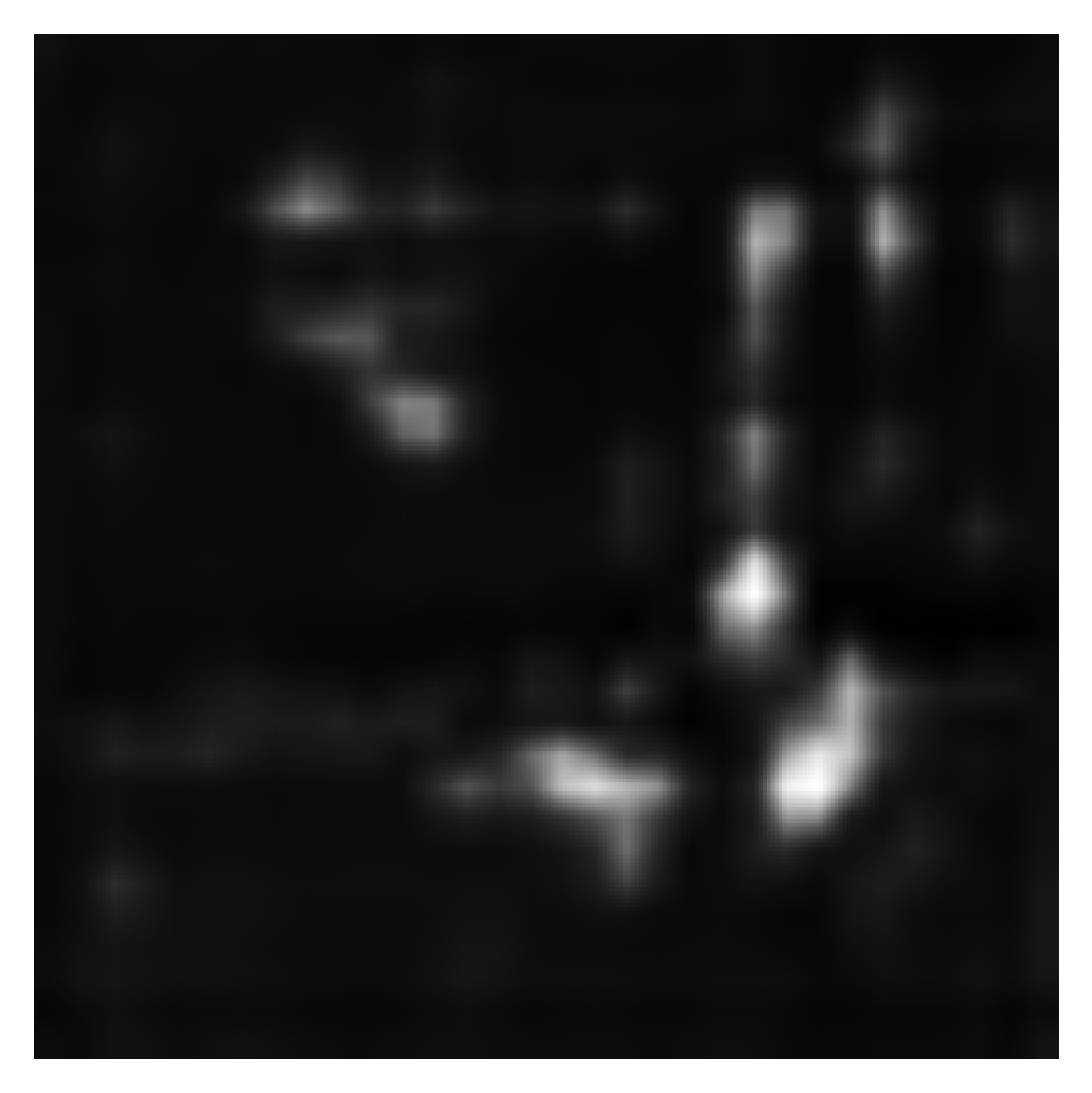}\\
\footnotesize{\rotatebox[origin=c]{90}{\ac{MTL}}} & &
\adjustimage{height=1.5cm,valign=m}{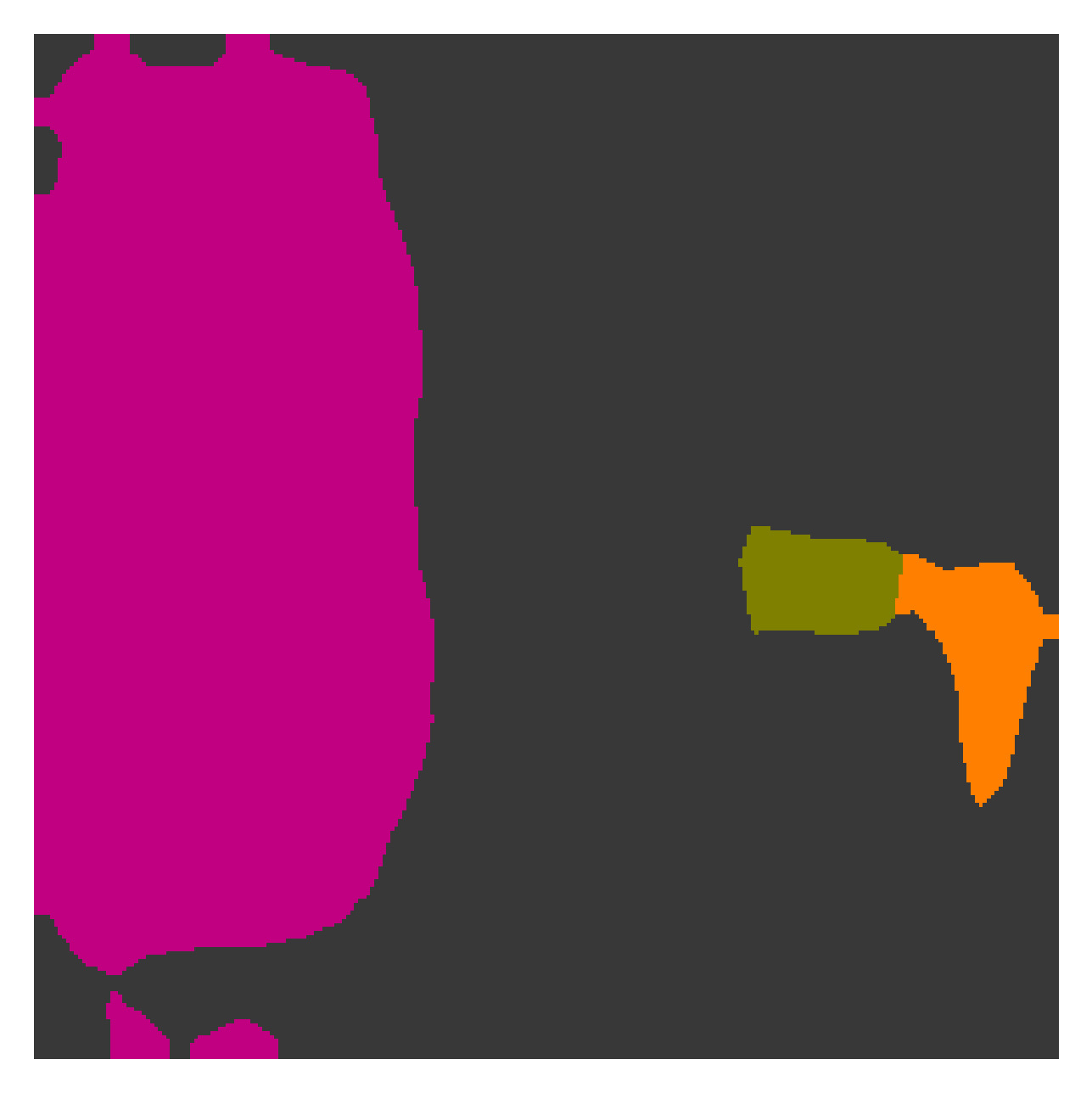} & \adjustimage{height=1.5cm,valign=m}{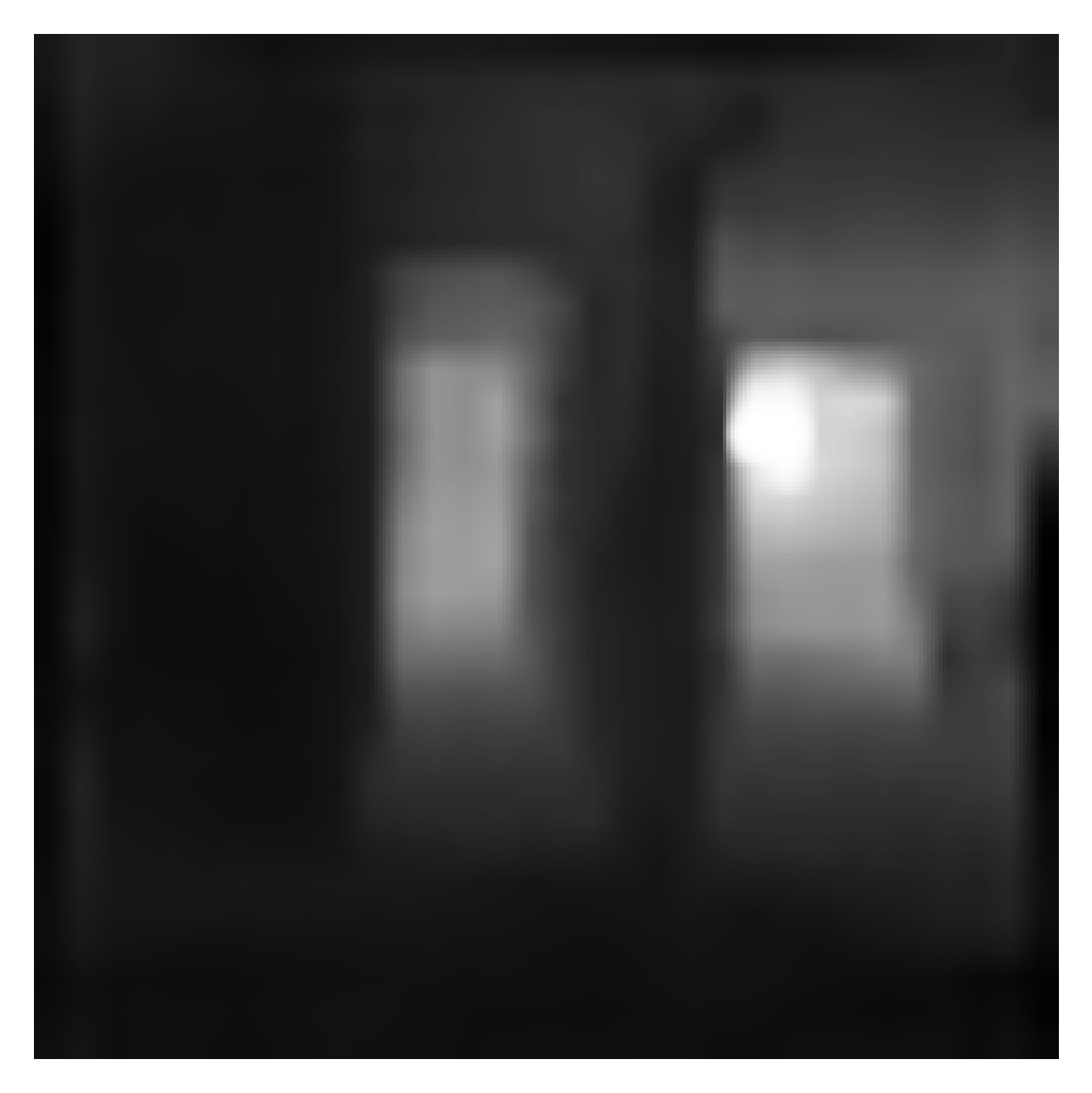} & \adjustimage{height=1.5cm,valign=m}{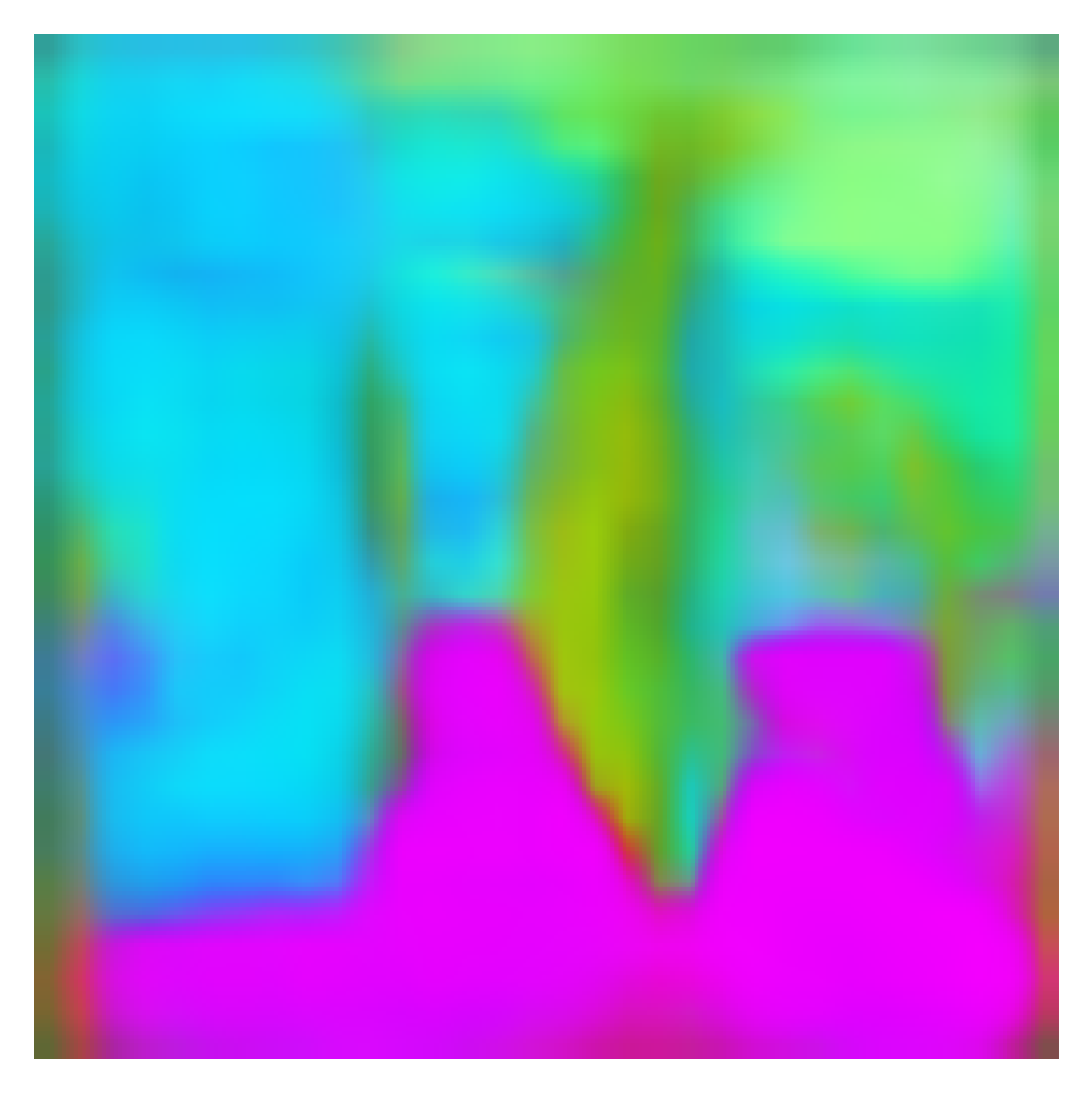} & \adjustimage{height=1.5cm,valign=m}{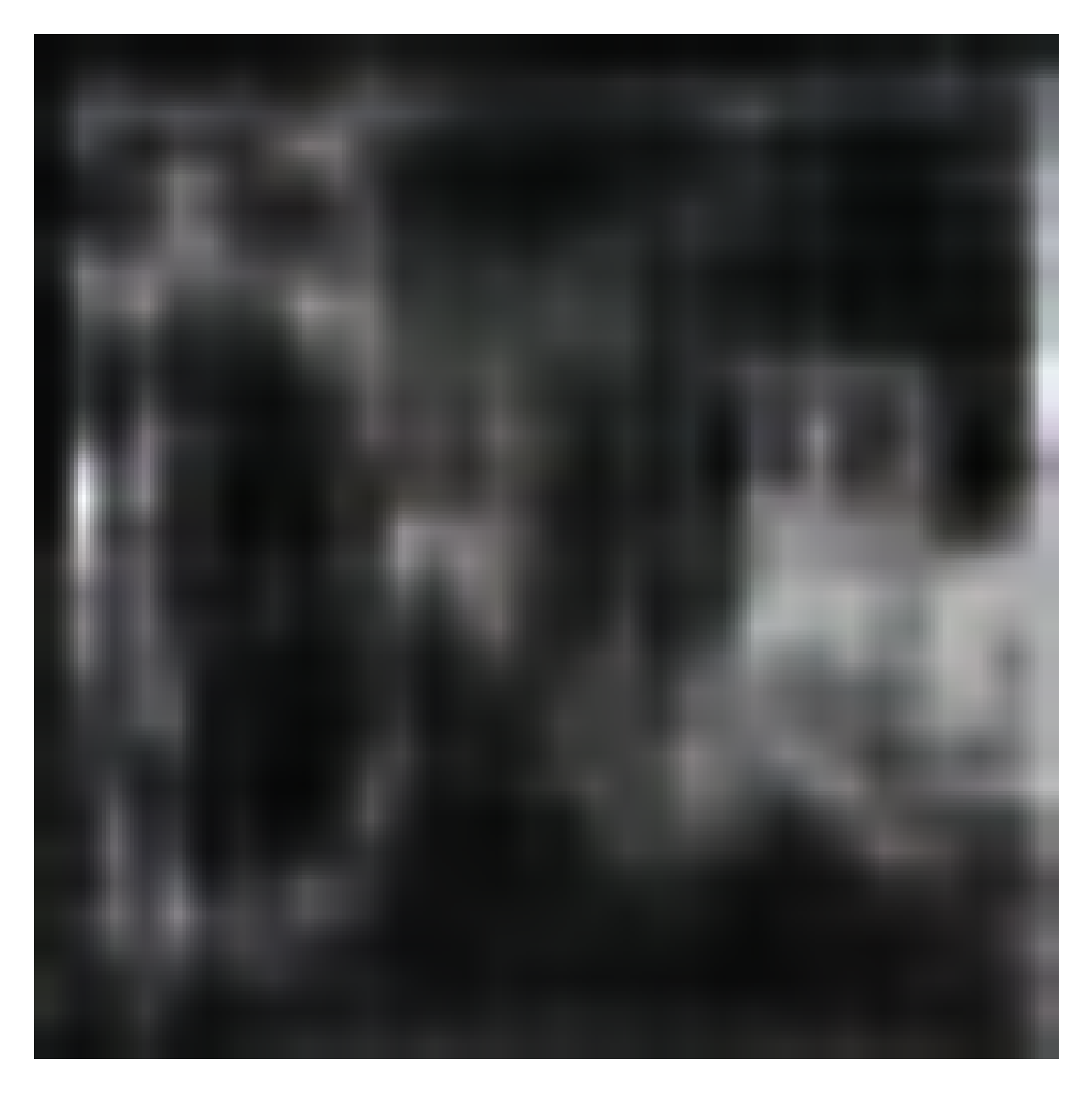} &
&
\adjustimage{height=1.5cm,valign=m}{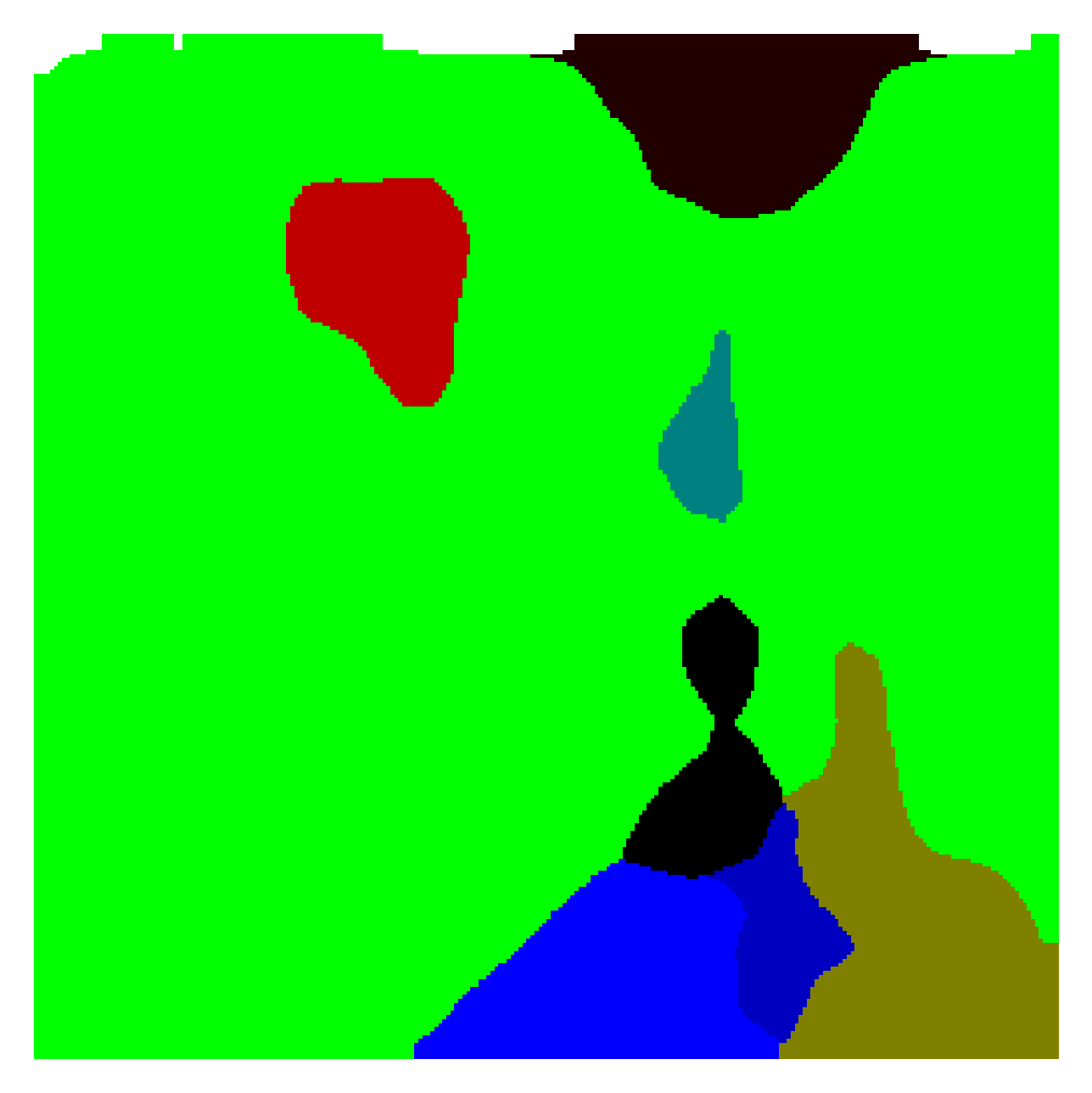} & 
\adjustimage{height=1.5cm,valign=m}{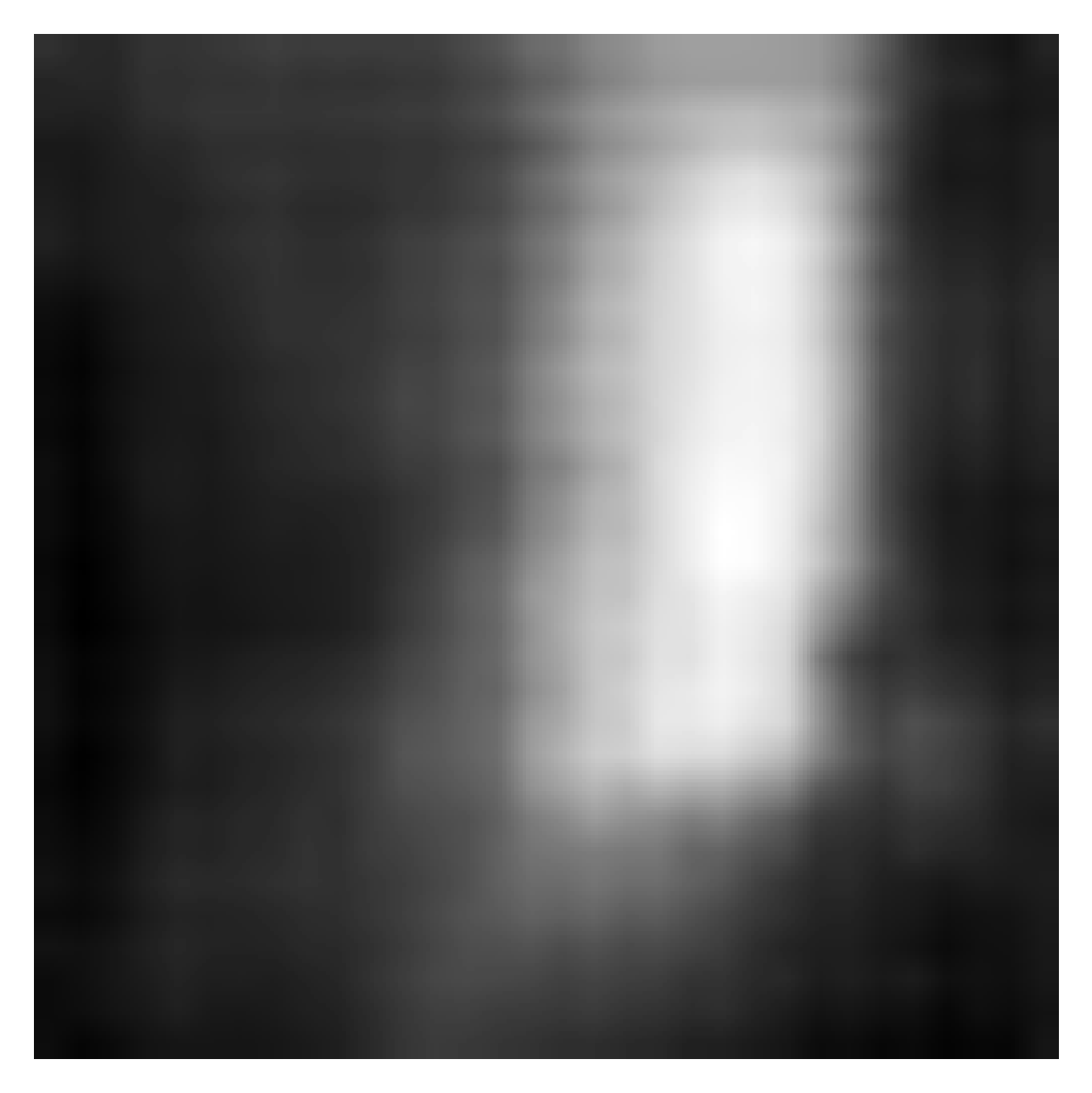} & \adjustimage{height=1.5cm,valign=m}{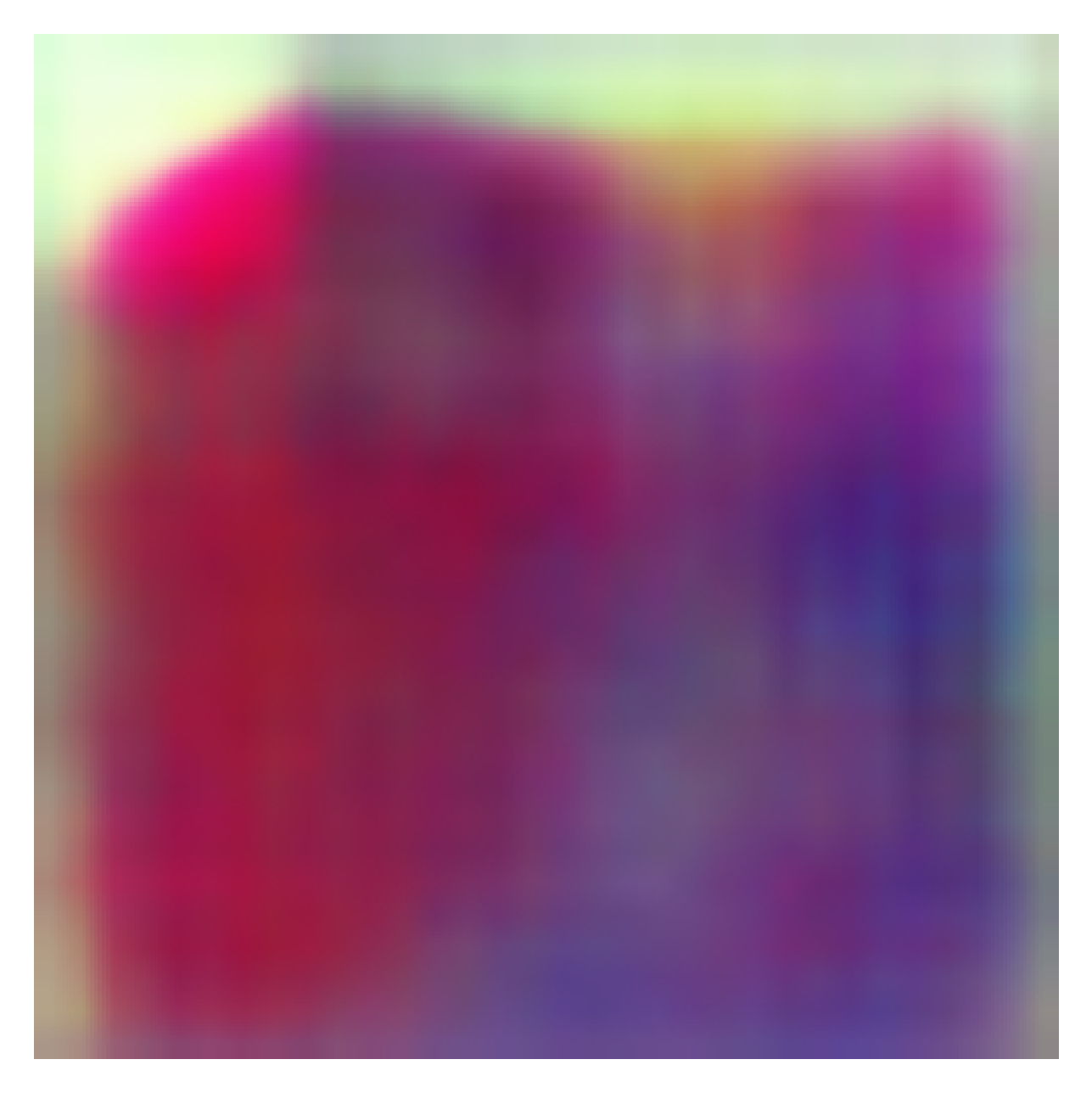}& \adjustimage{height=1.5cm,valign=m}{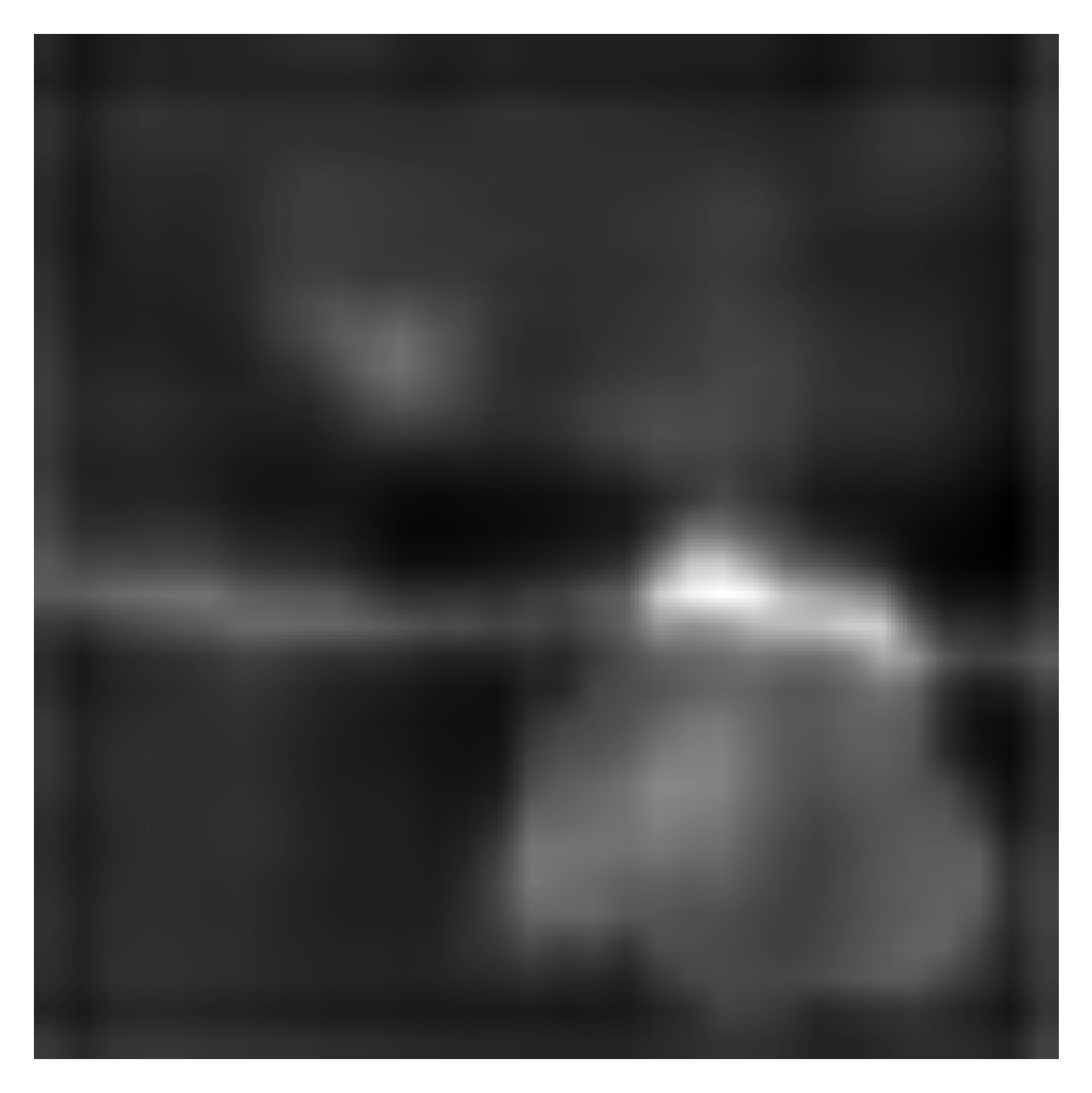}\\
\footnotesize{\rotatebox[origin=c]{90}{\ac{MTML}}} & &
\adjustimage{height=1.5cm,valign=m}{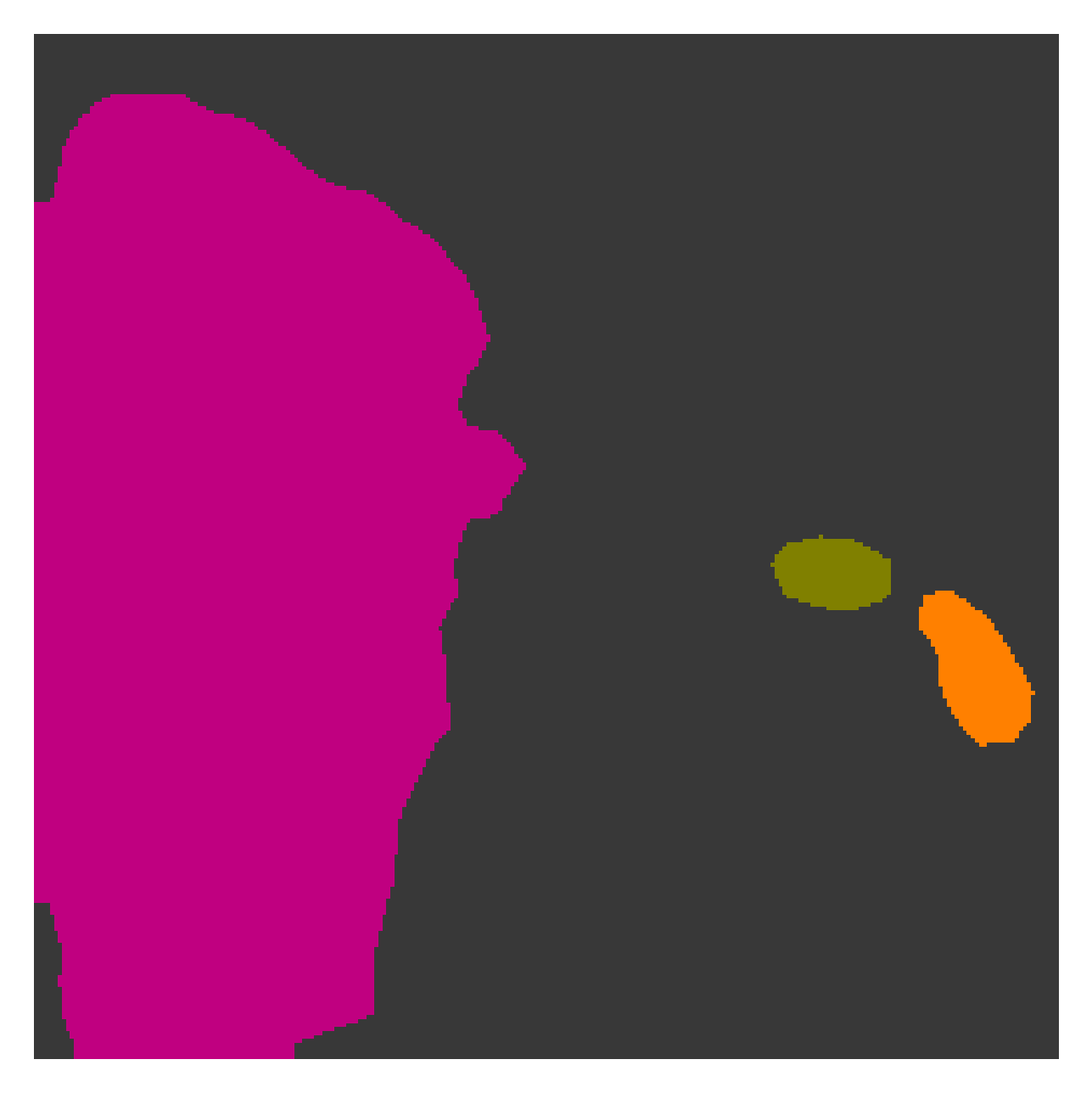} & \adjustimage{height=1.5cm,valign=m}{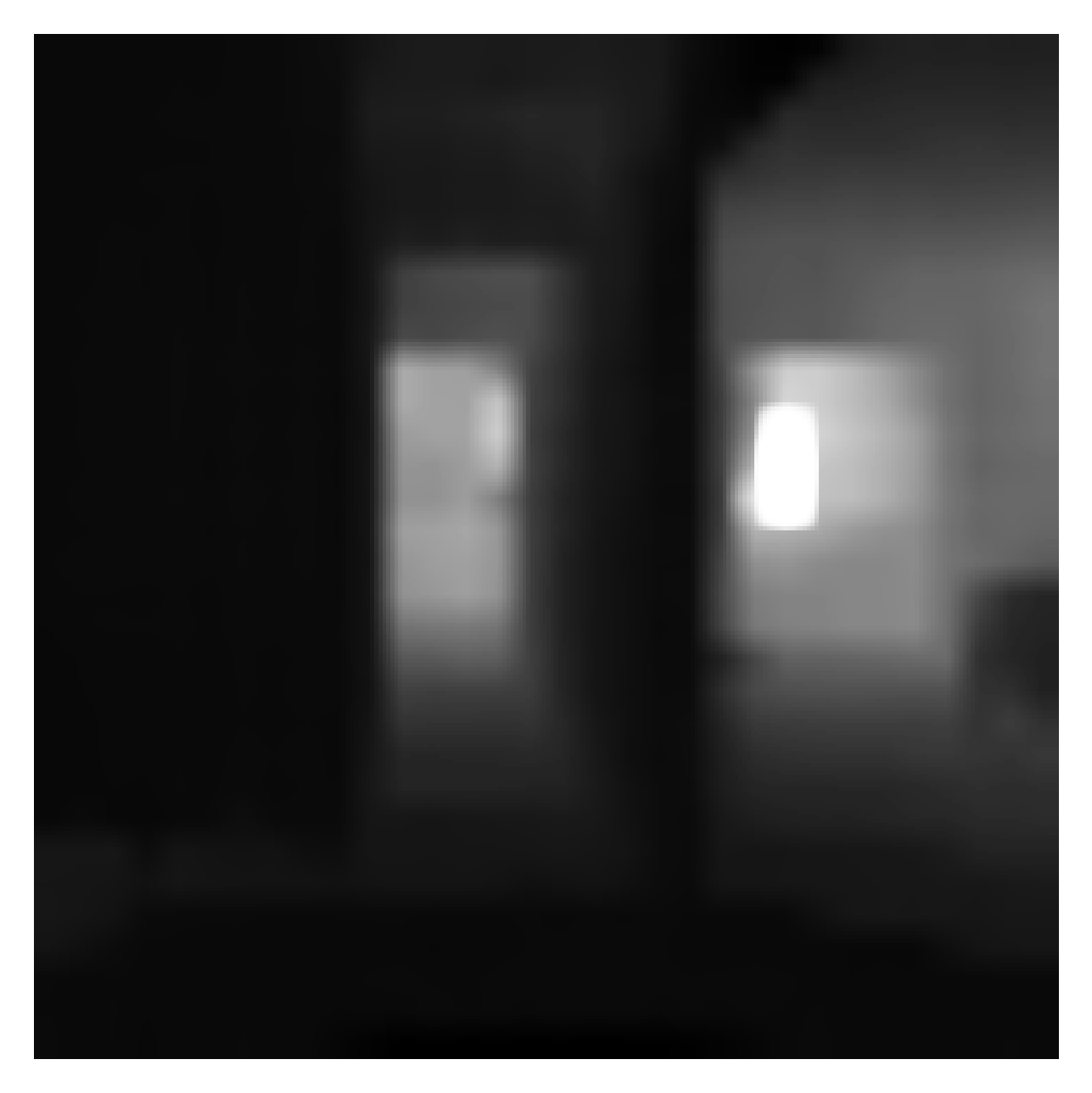} & \adjustimage{height=1.5cm,valign=m}{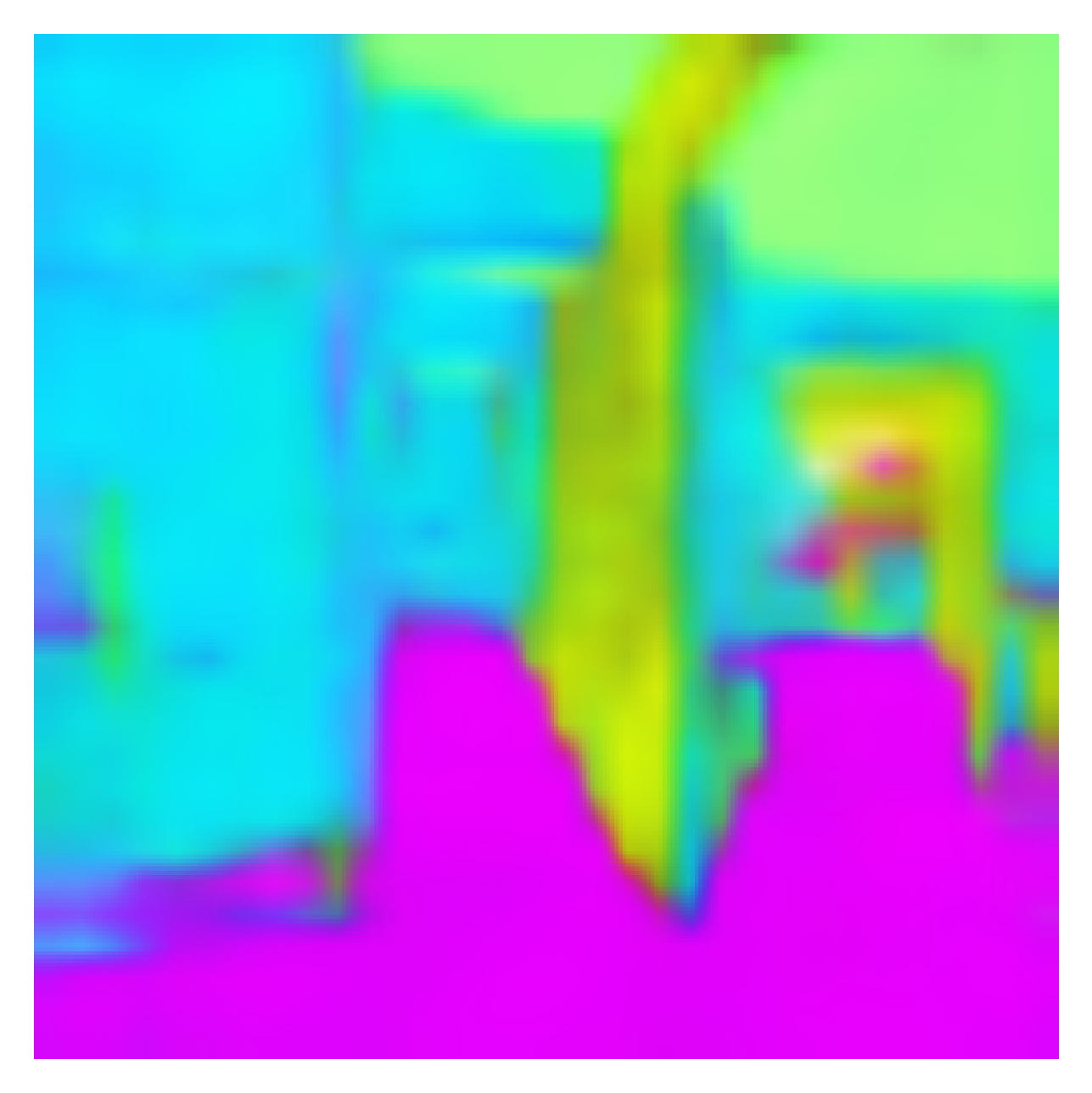} & \adjustimage{height=1.5cm,valign=m}{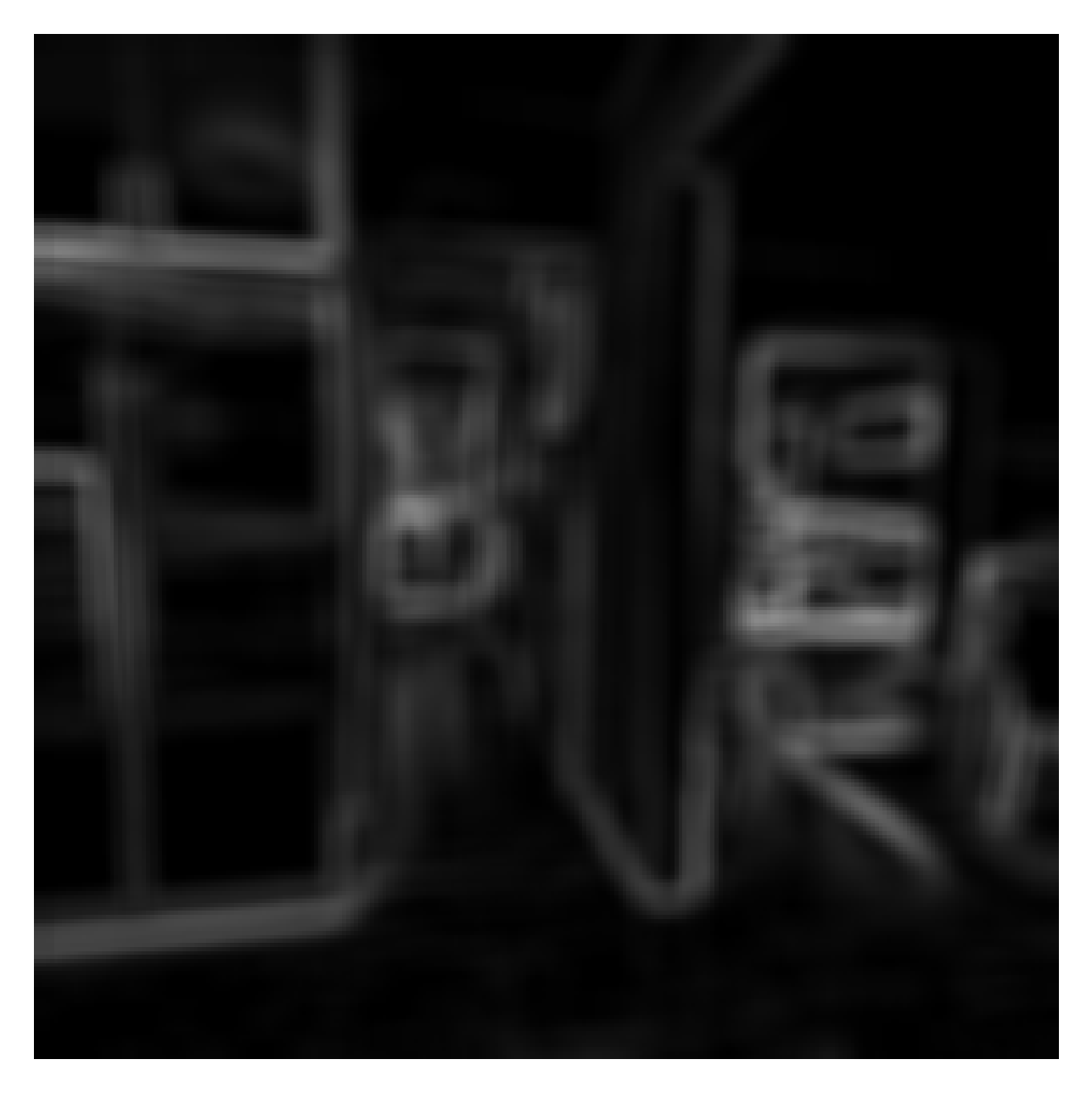} &
&
\adjustimage{height=1.5cm,valign=m}{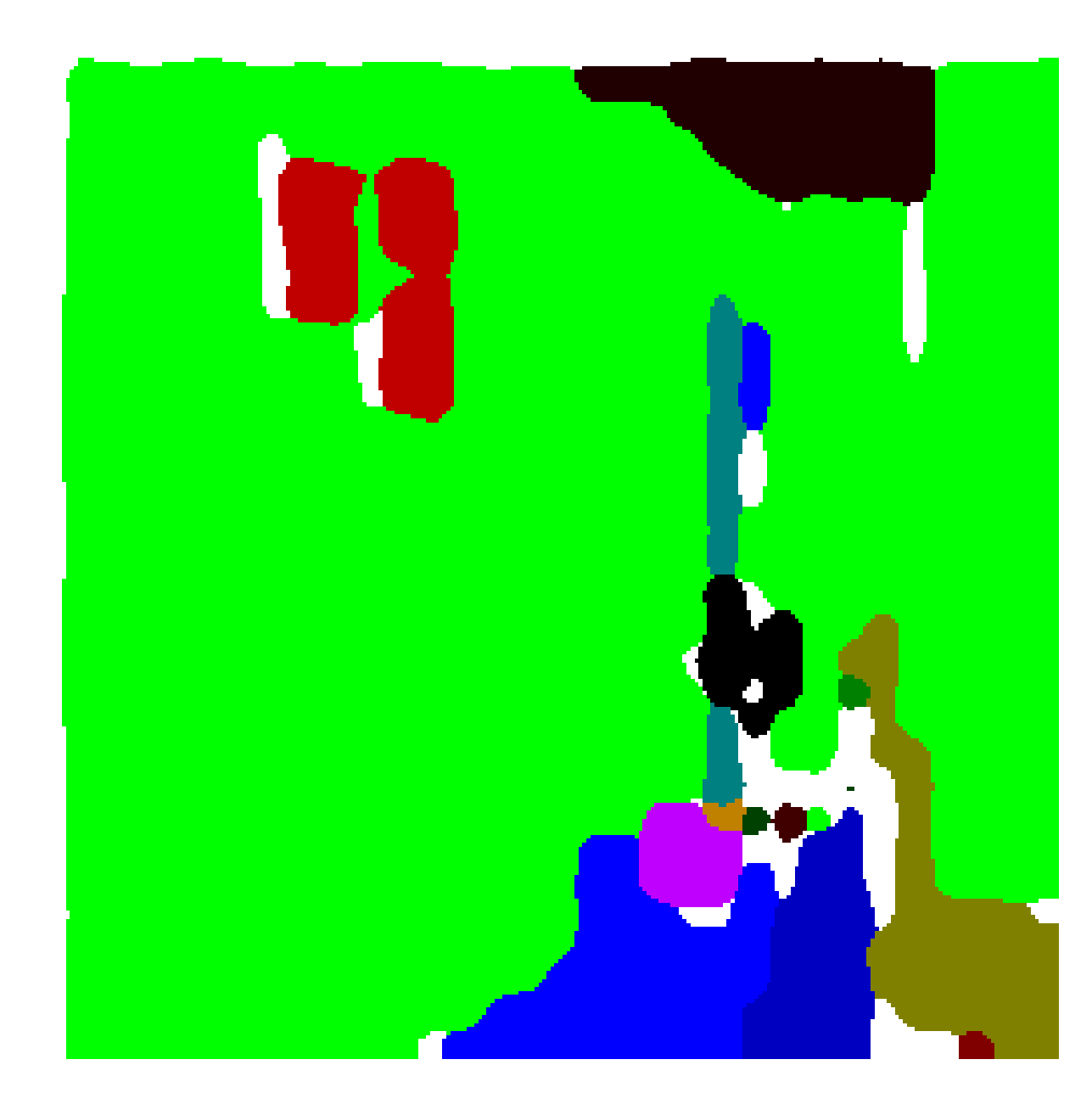} & 
\adjustimage{height=1.5cm,valign=m}{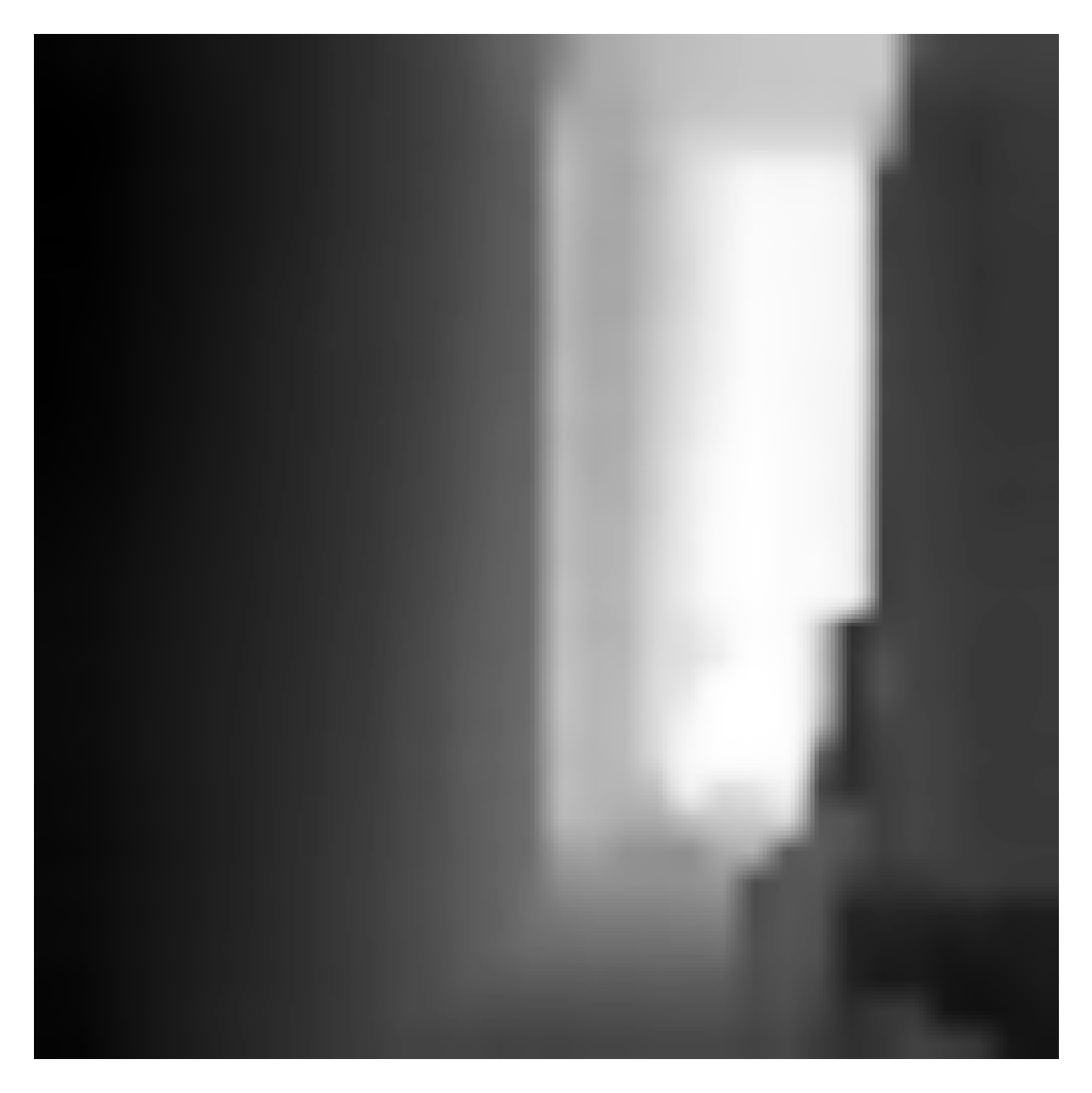} & \adjustimage{height=1.5cm,valign=m}{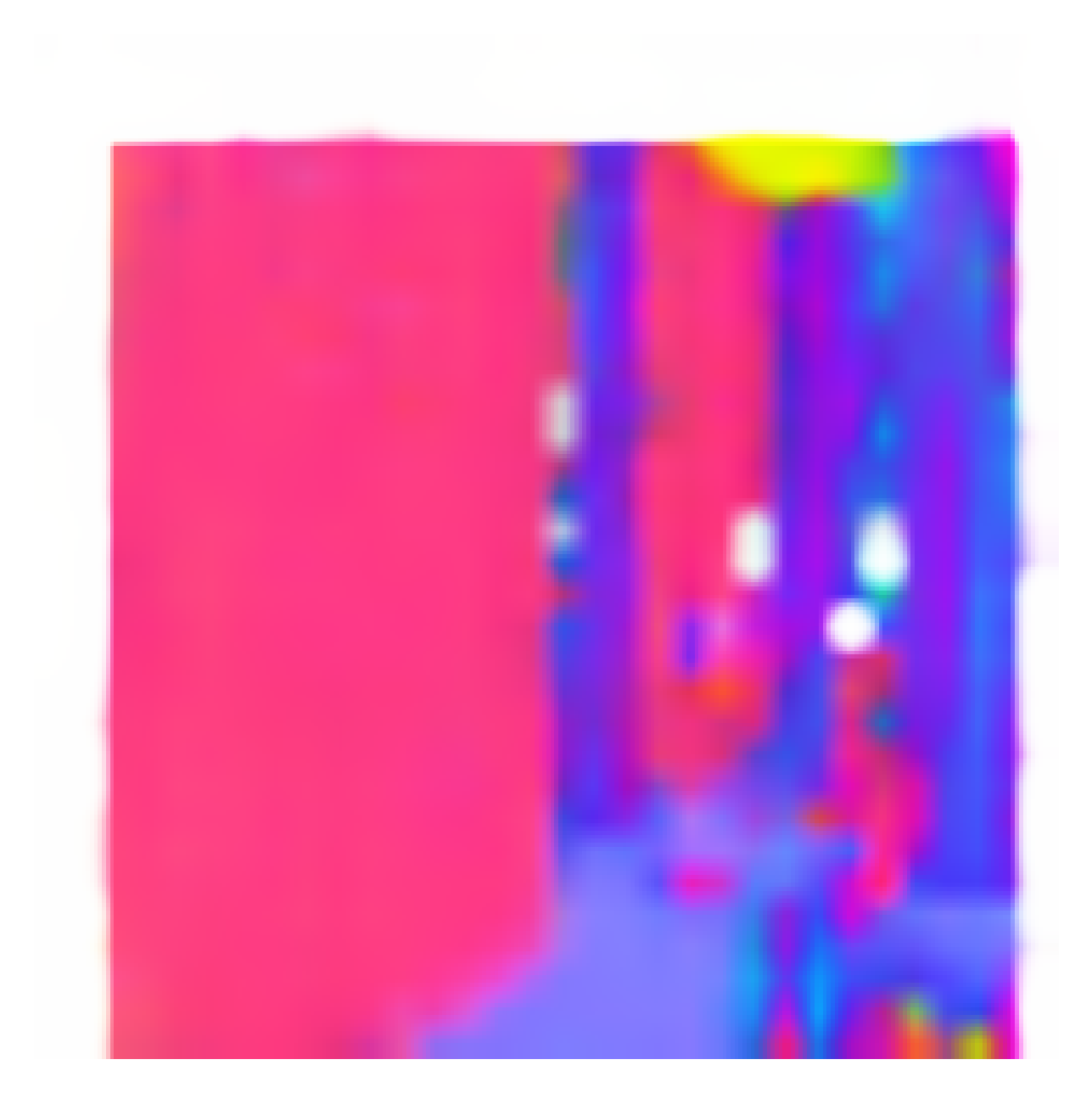}& \adjustimage{height=1.5cm,valign=m}{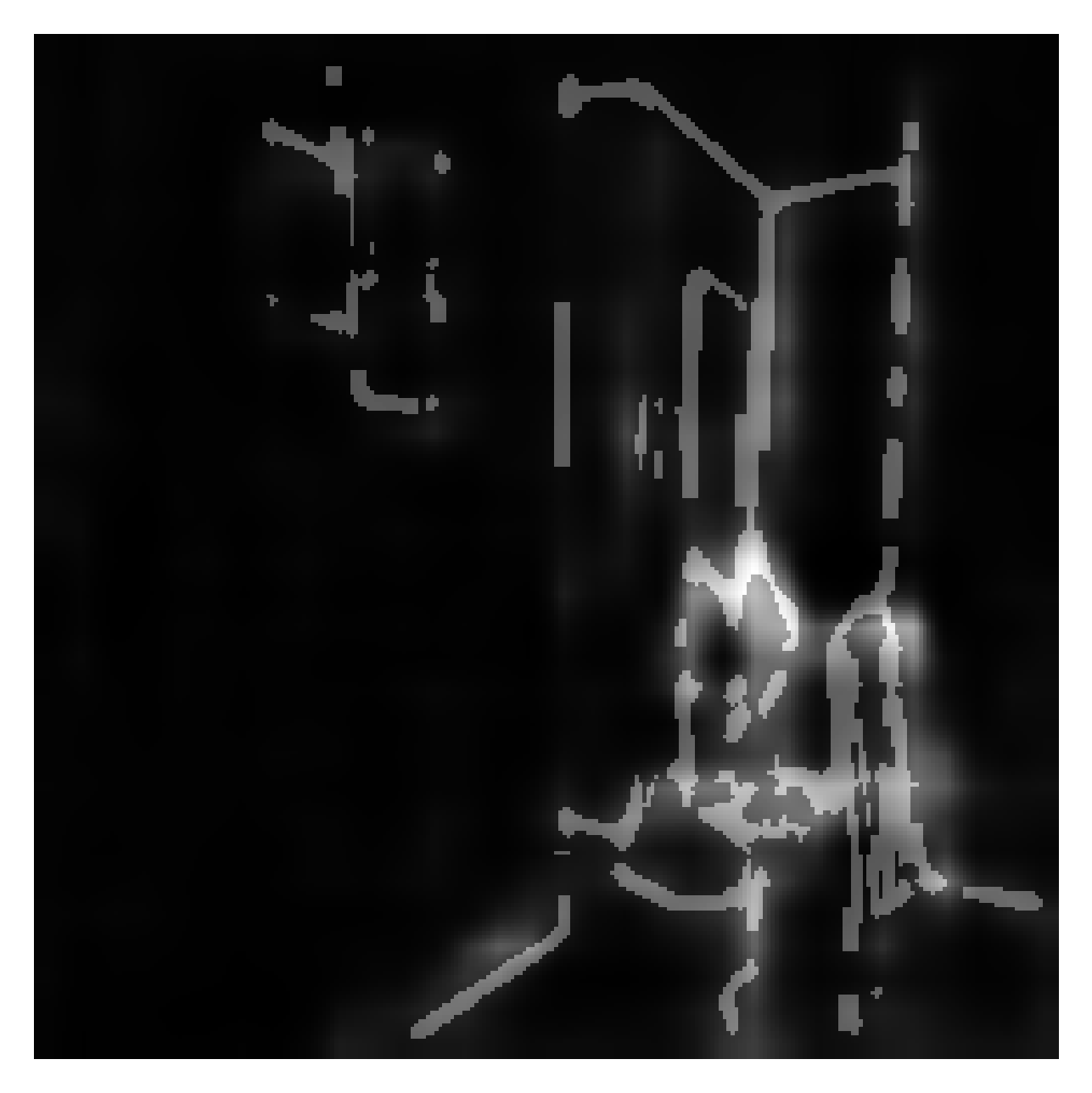}\\

\hline

\footnotesize{\rotatebox[origin=c]{90}{GT}} & \multirow[b]{ 4}{*}{\adjustimage{height=1.5cm,valign=m}{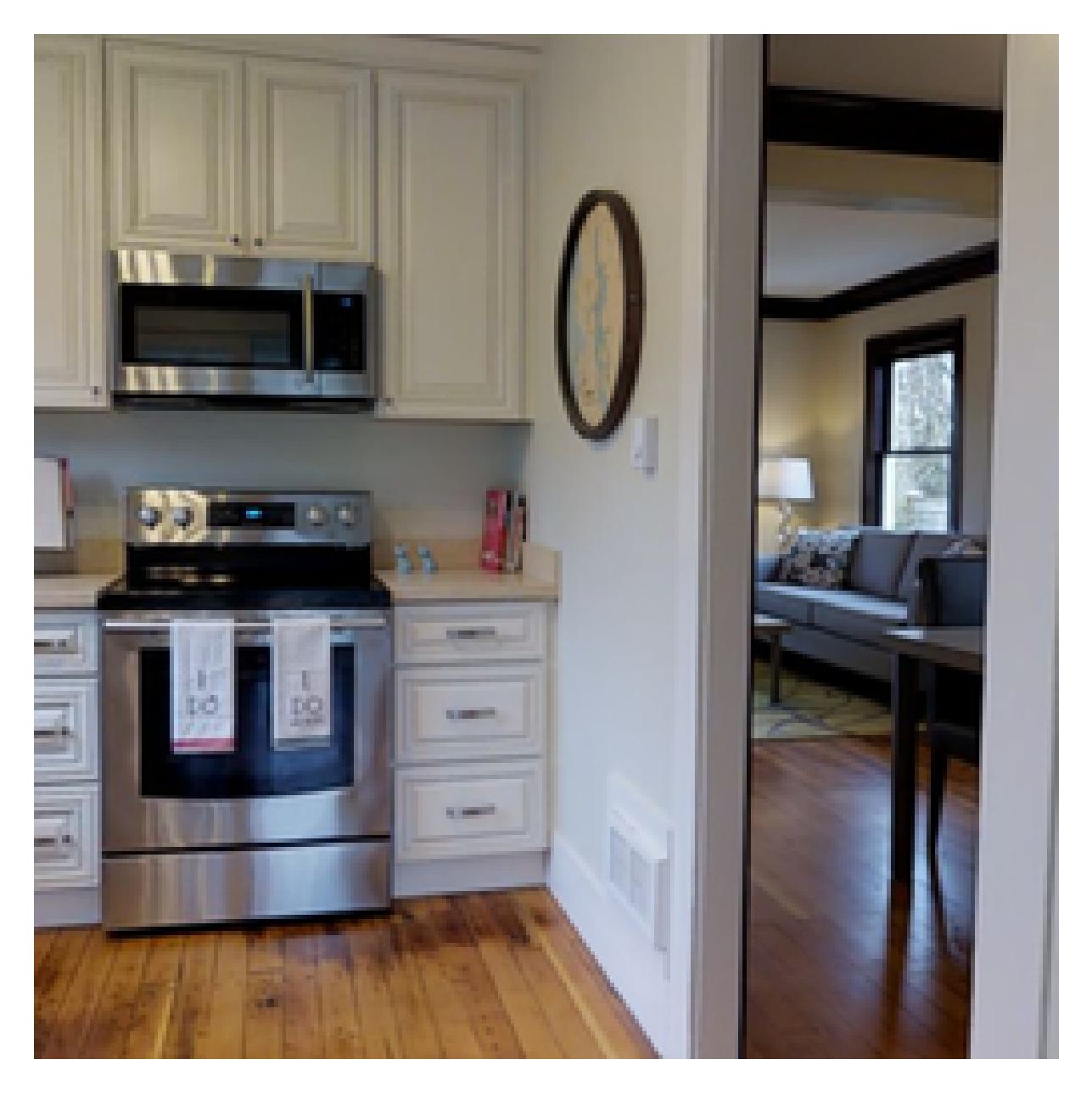}} &
\adjustimage{height=1.46cm,valign=m}{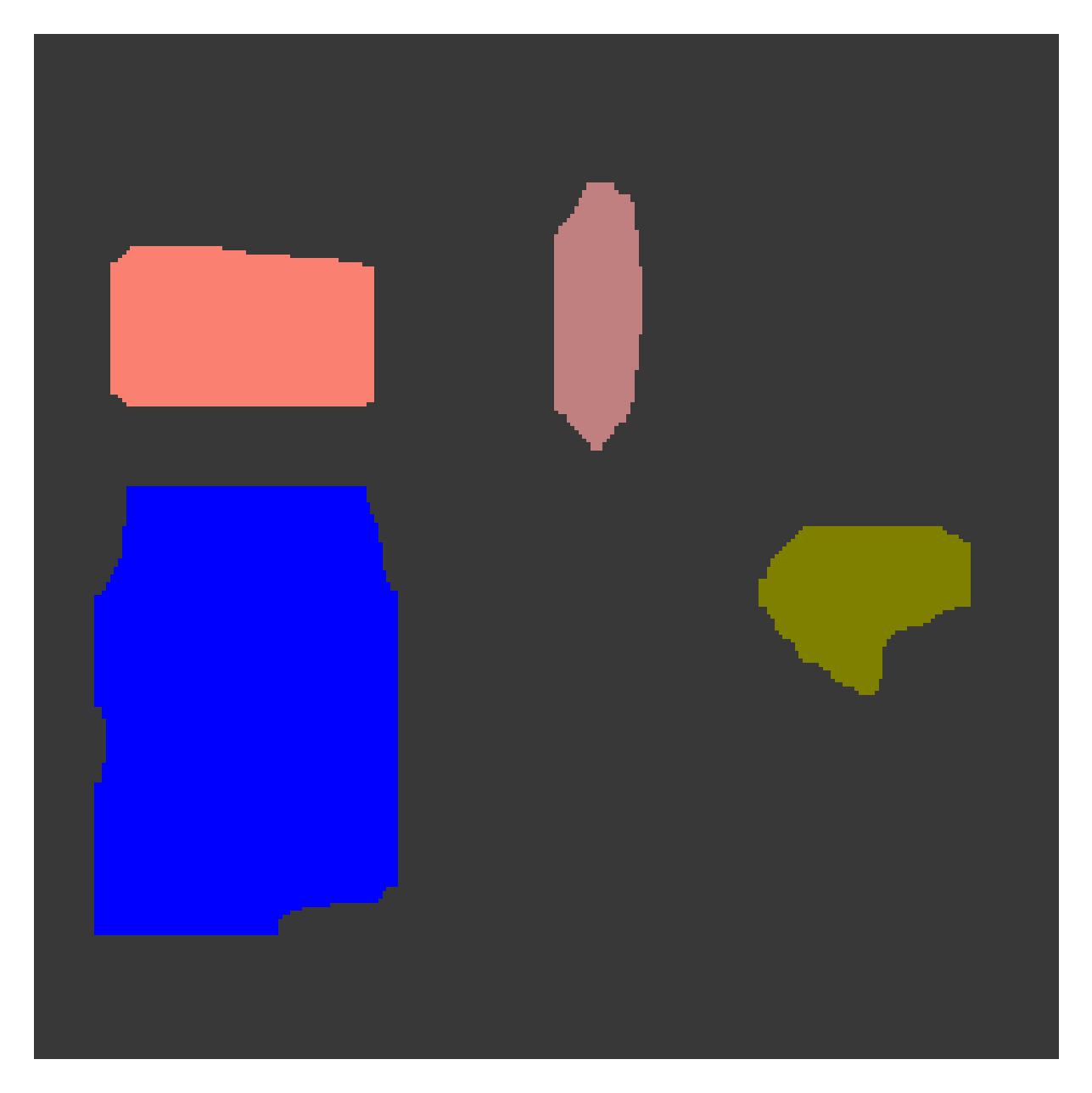} & \adjustimage{height=1.46cm,valign=m}{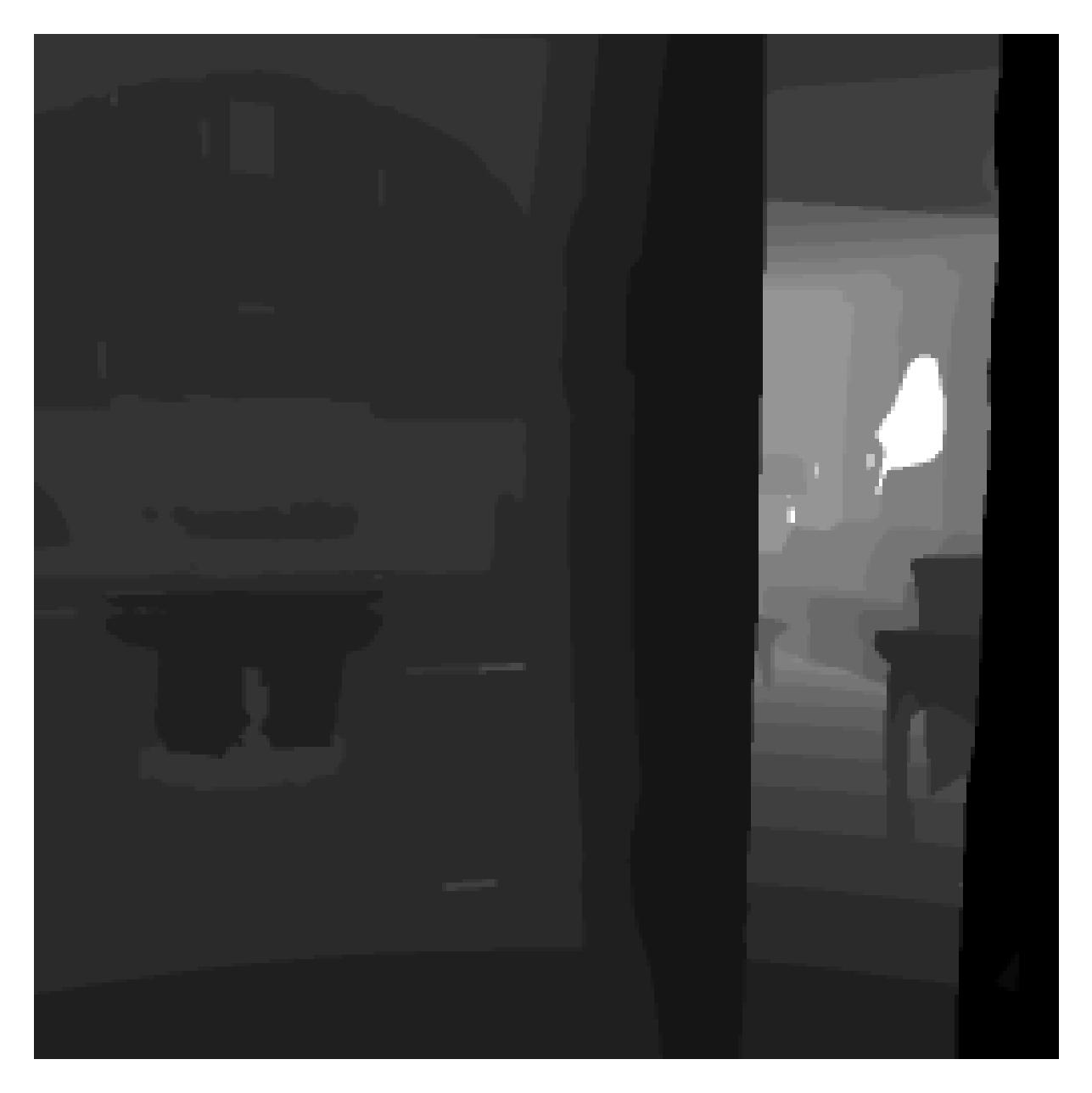} & \adjustimage{height=1.46cm,valign=m}{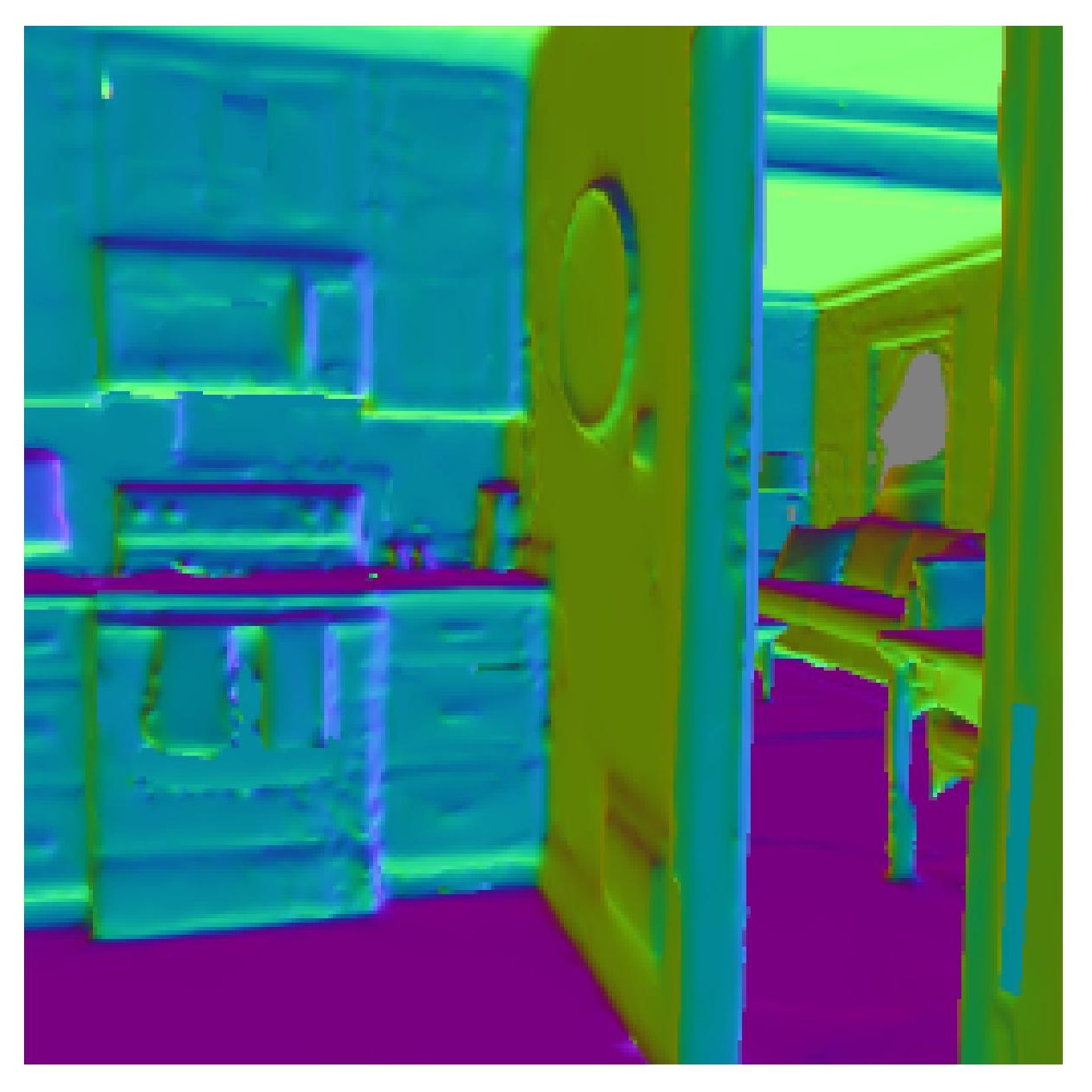} & \adjustimage{height=1.46cm,valign=m}{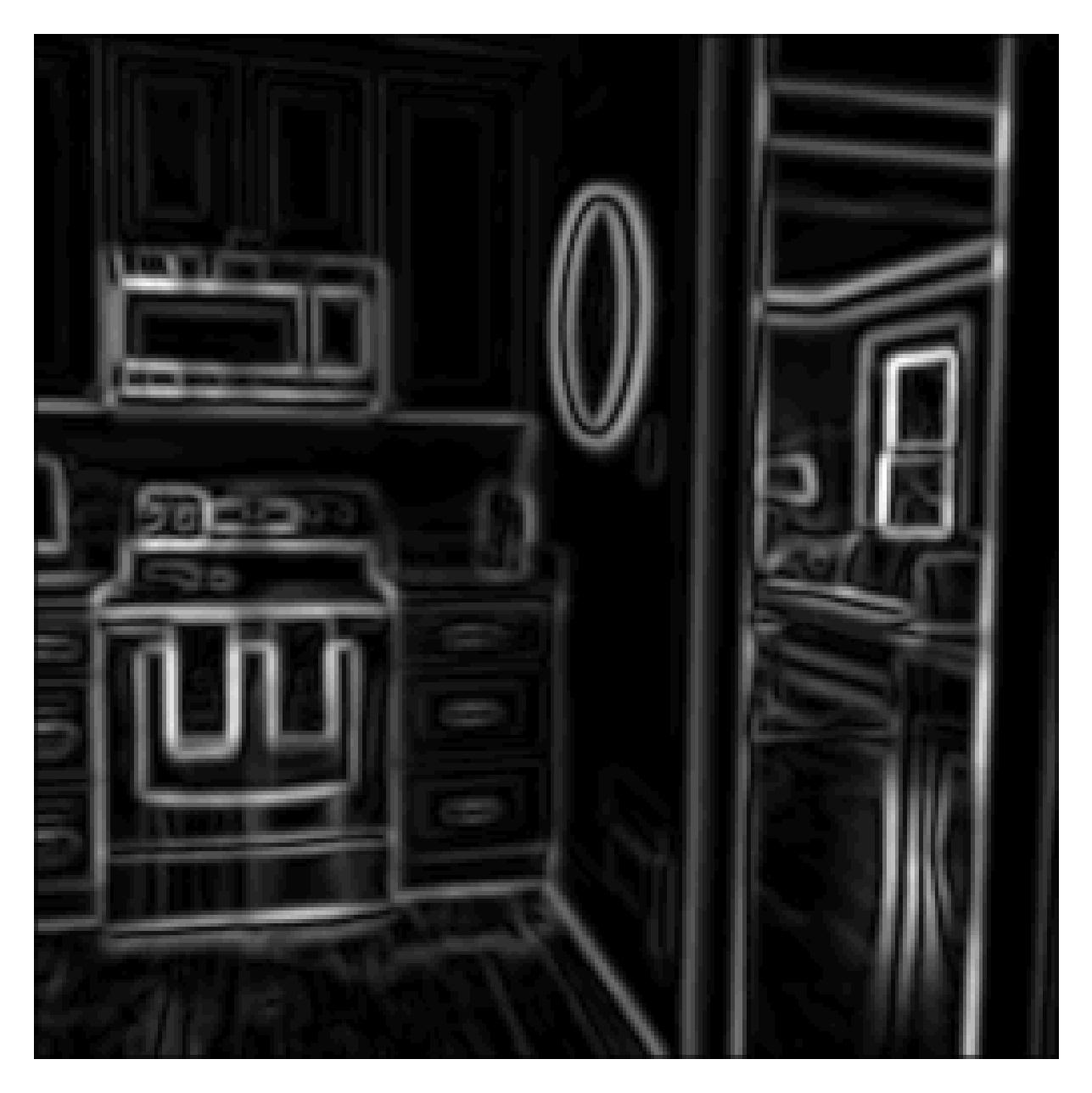} &
\multirow[b]{ 4}{*}{\adjustimage{height=1.5cm,valign=m}{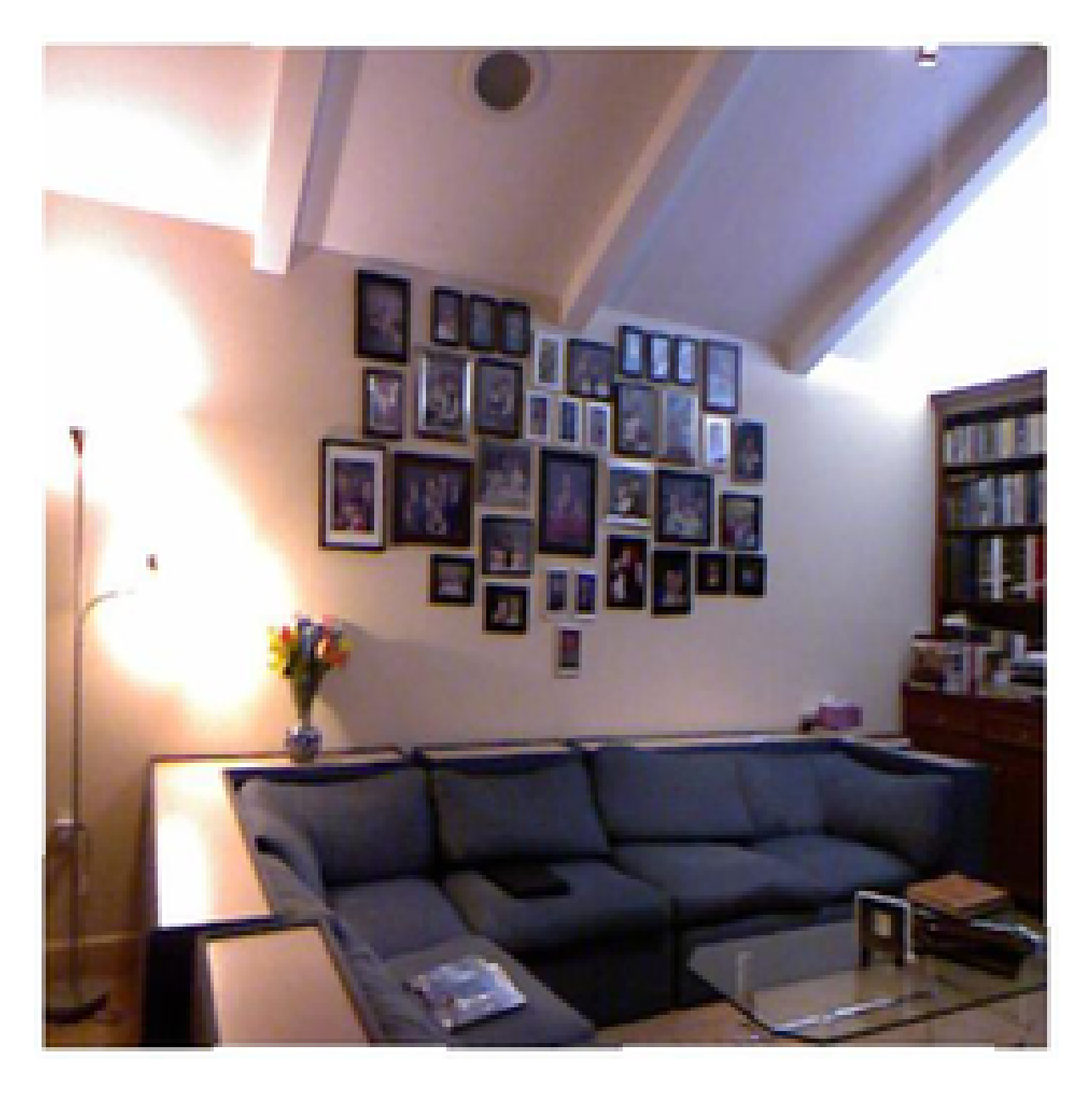}}&
\adjustimage{height=1.5cm,valign=m}{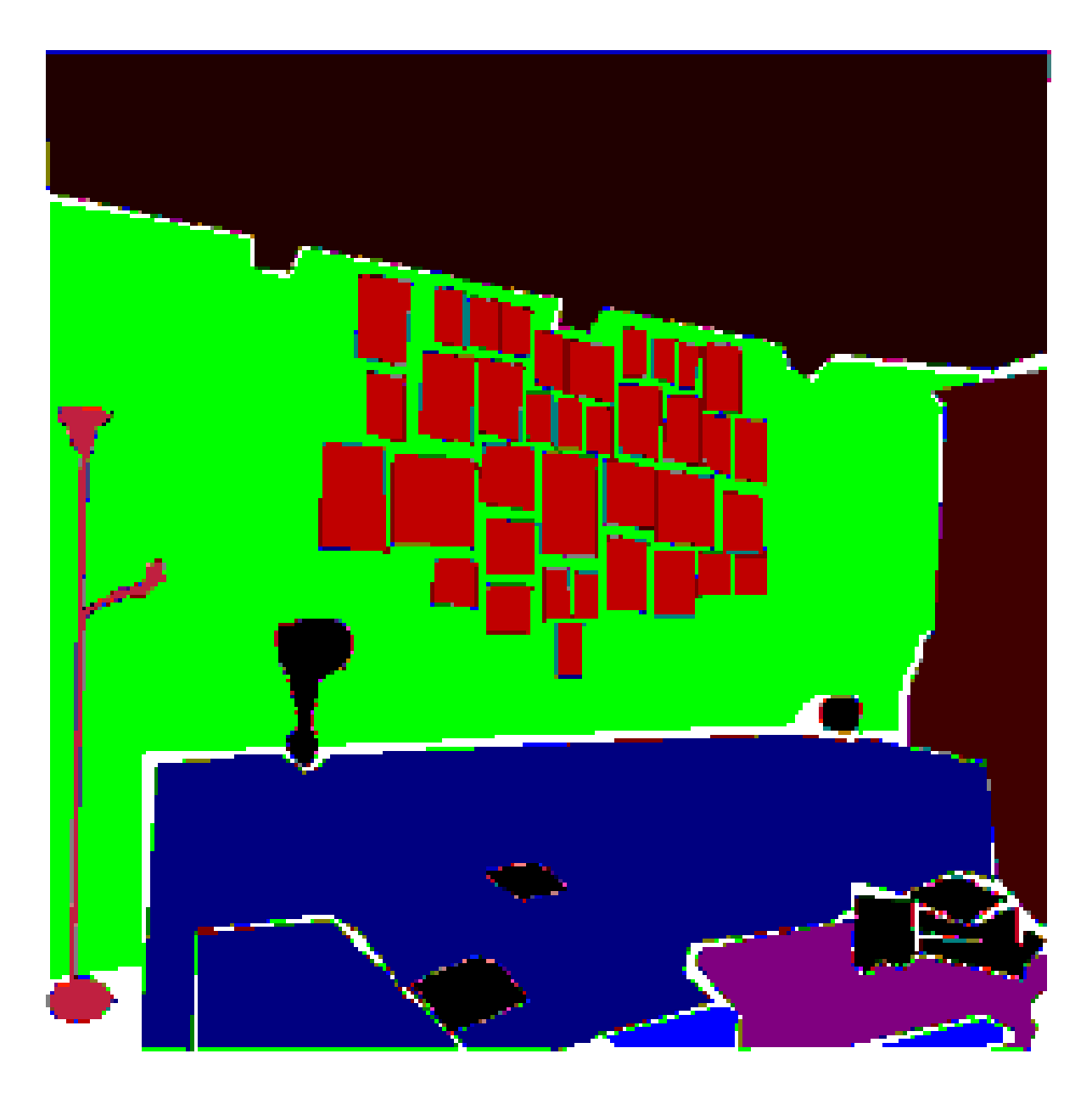} & 
\adjustimage{height=1.5cm,valign=m}{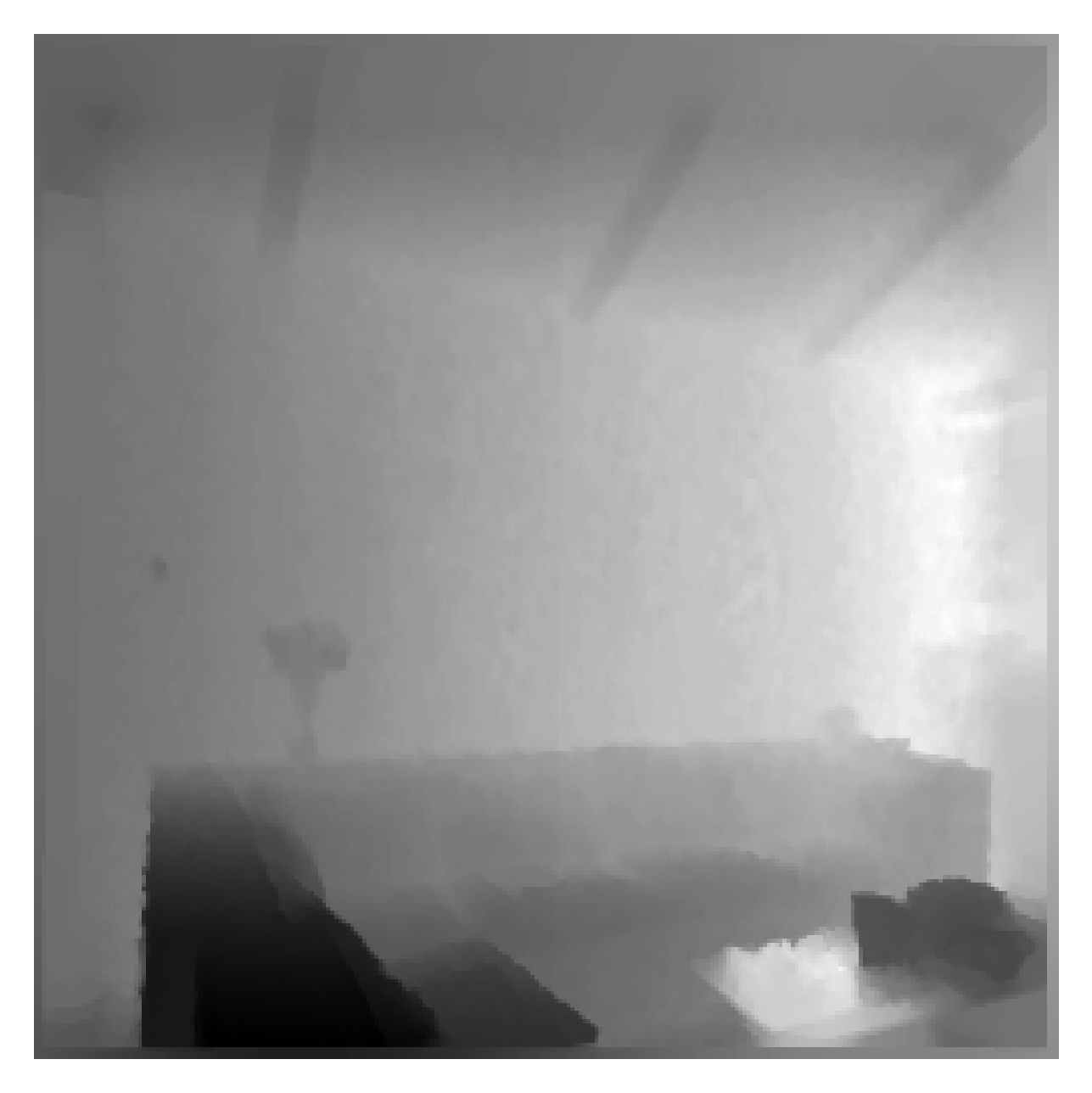} & \adjustimage{height=1.5cm,valign=m}{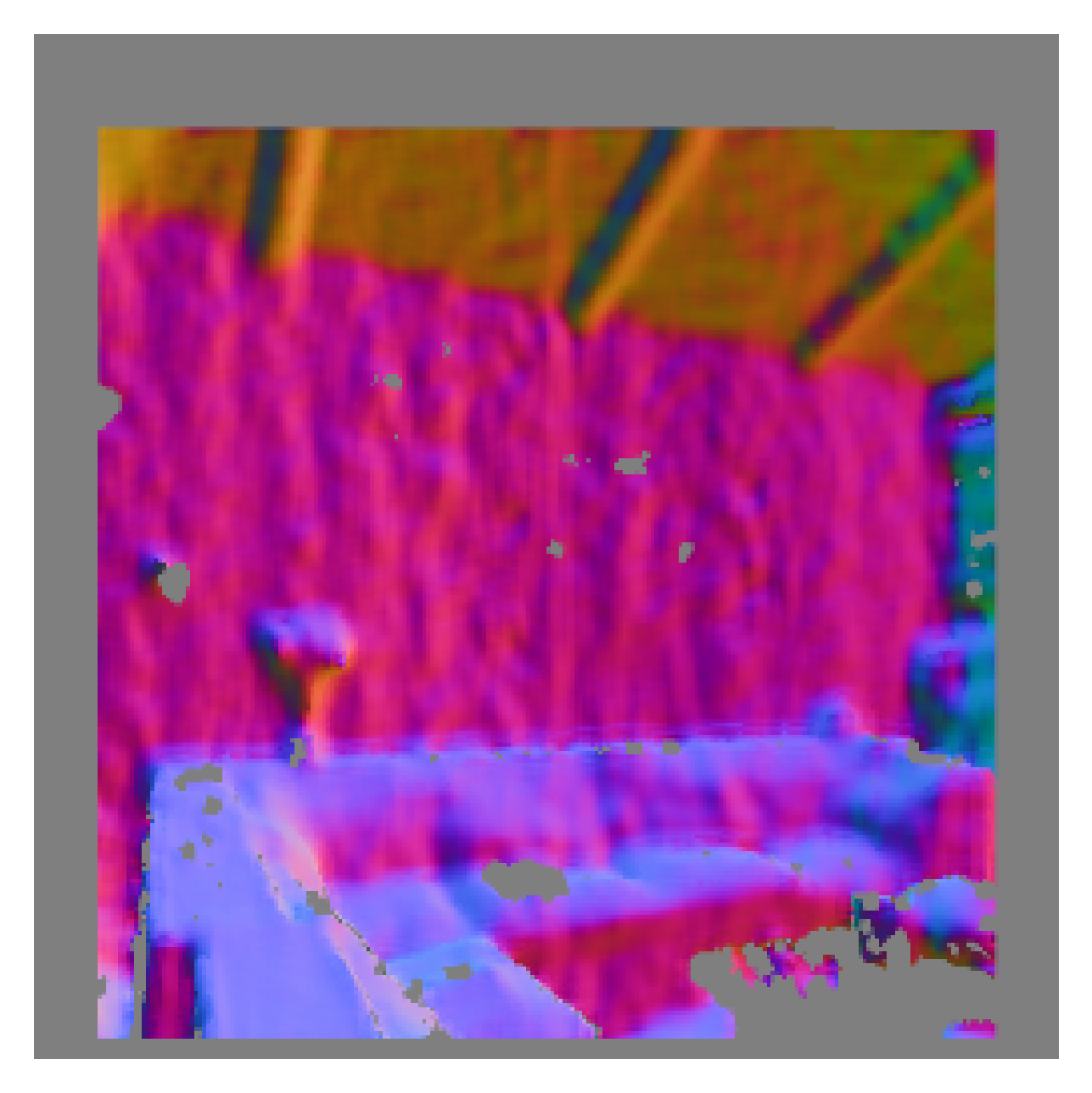}& \adjustimage{height=1.5cm,valign=m}{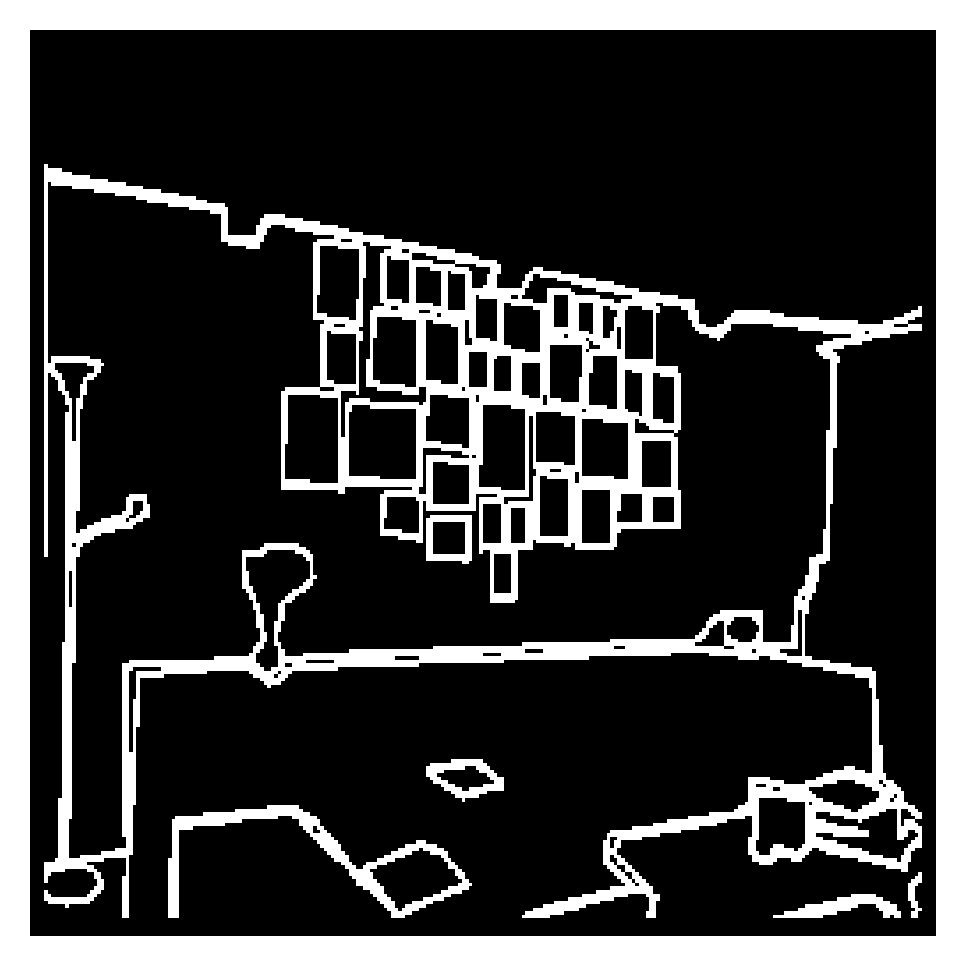}\\
\footnotesize{\rotatebox[origin=c]{90}{Single}} & &
\adjustimage{height=1.5cm,valign=m}{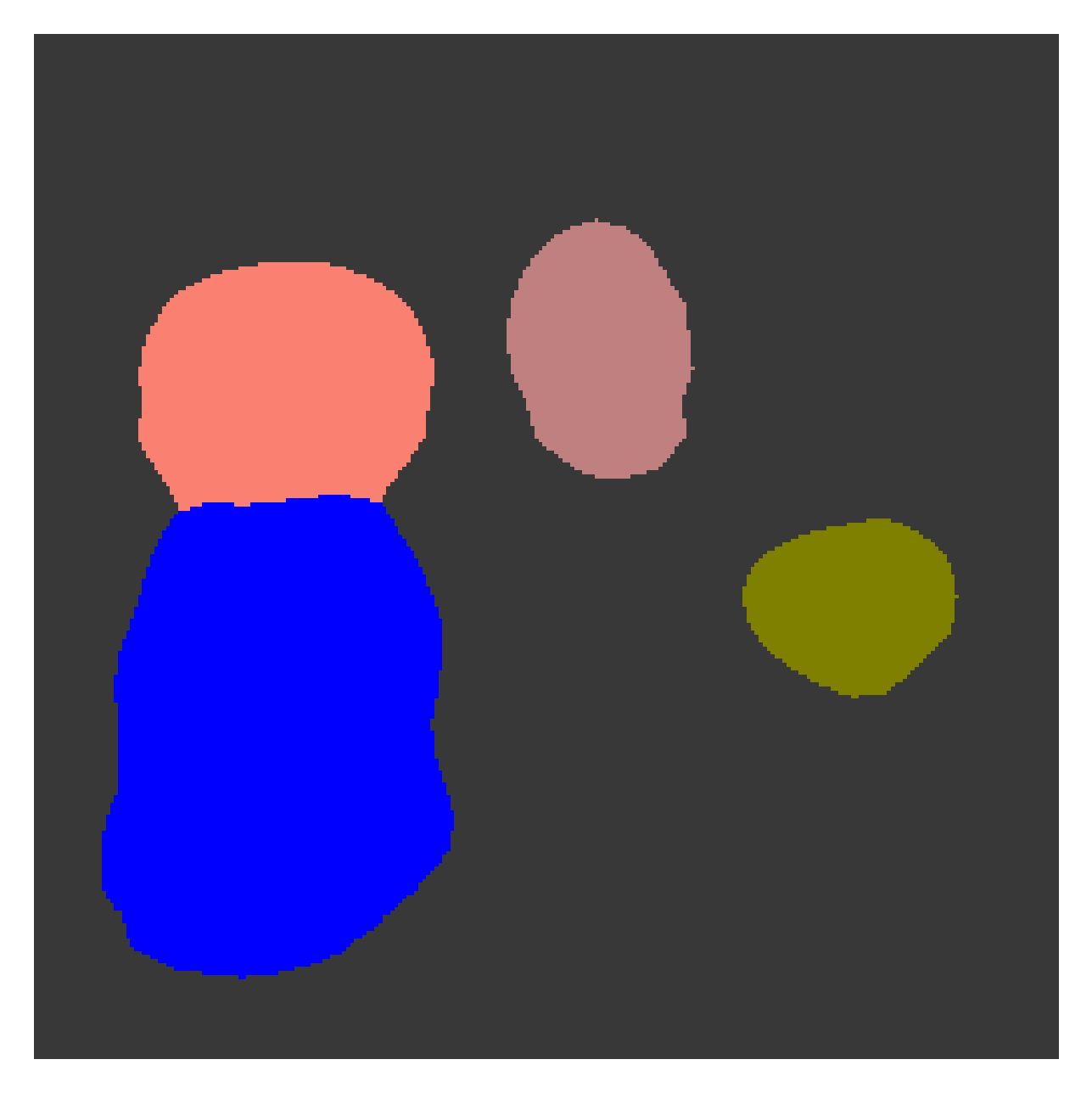} & \adjustimage{height=1.5cm,valign=m}{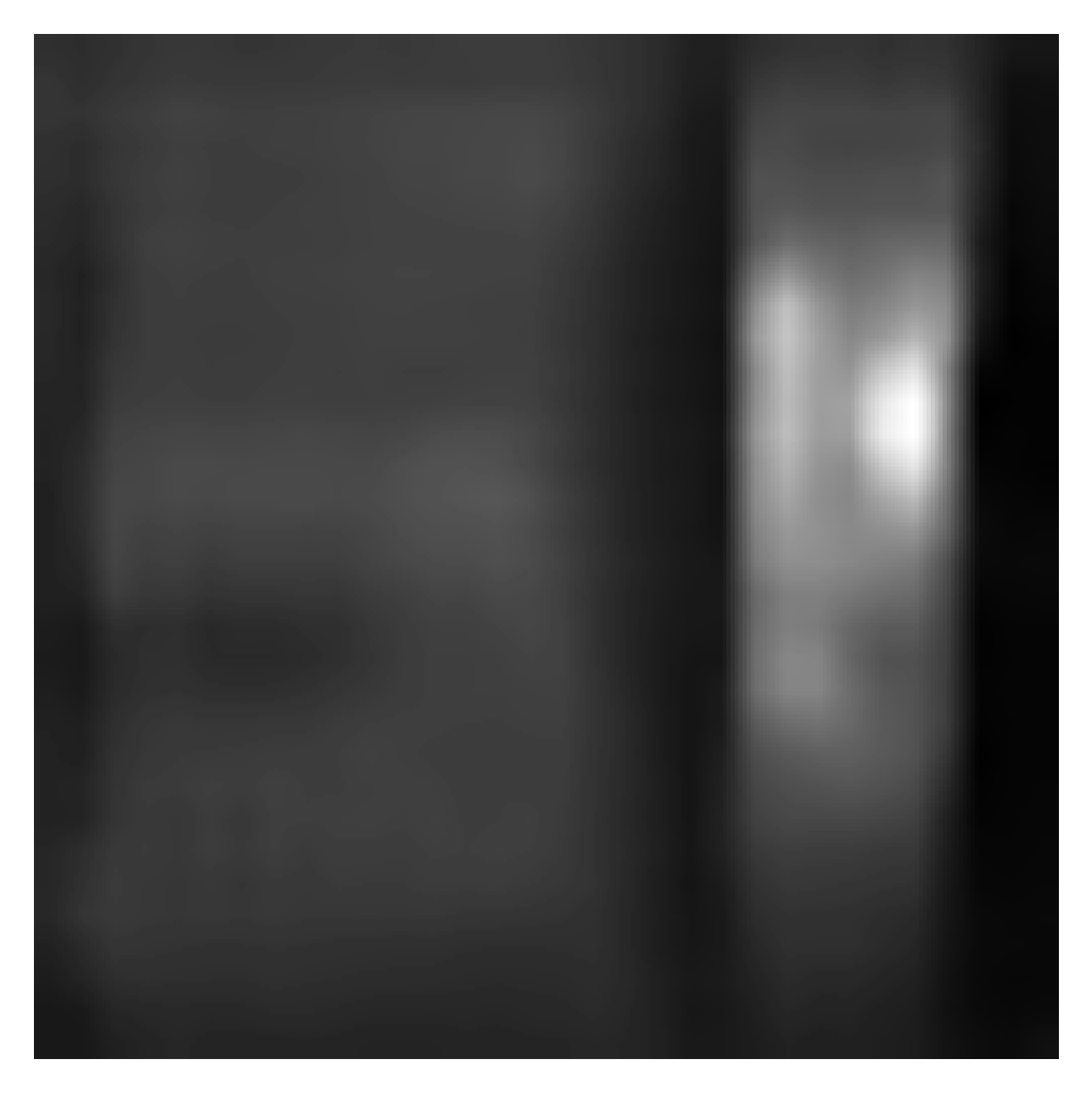} & \adjustimage{height=1.5cm,valign=m}{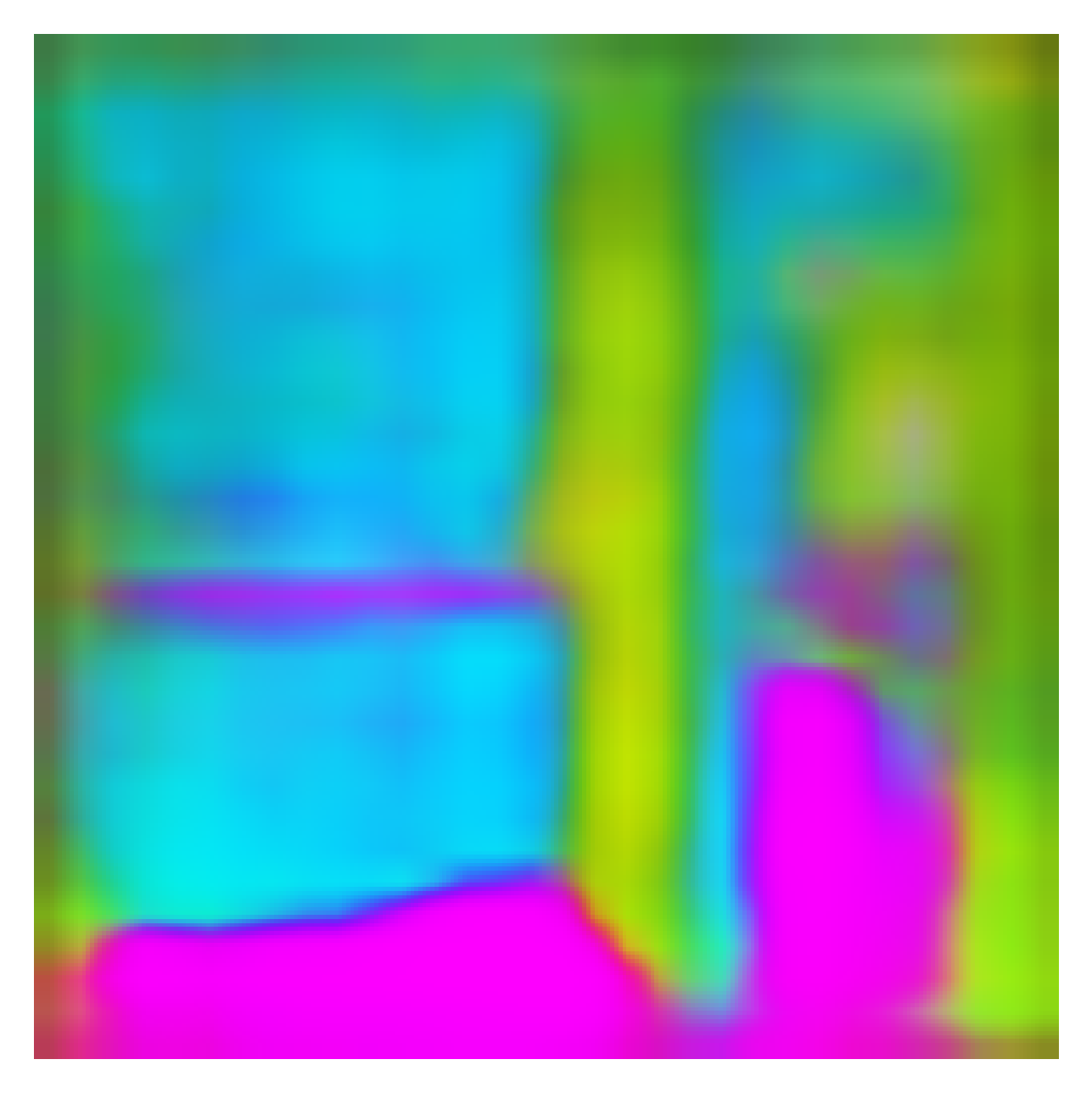} & \adjustimage{height=1.5cm,valign=m}{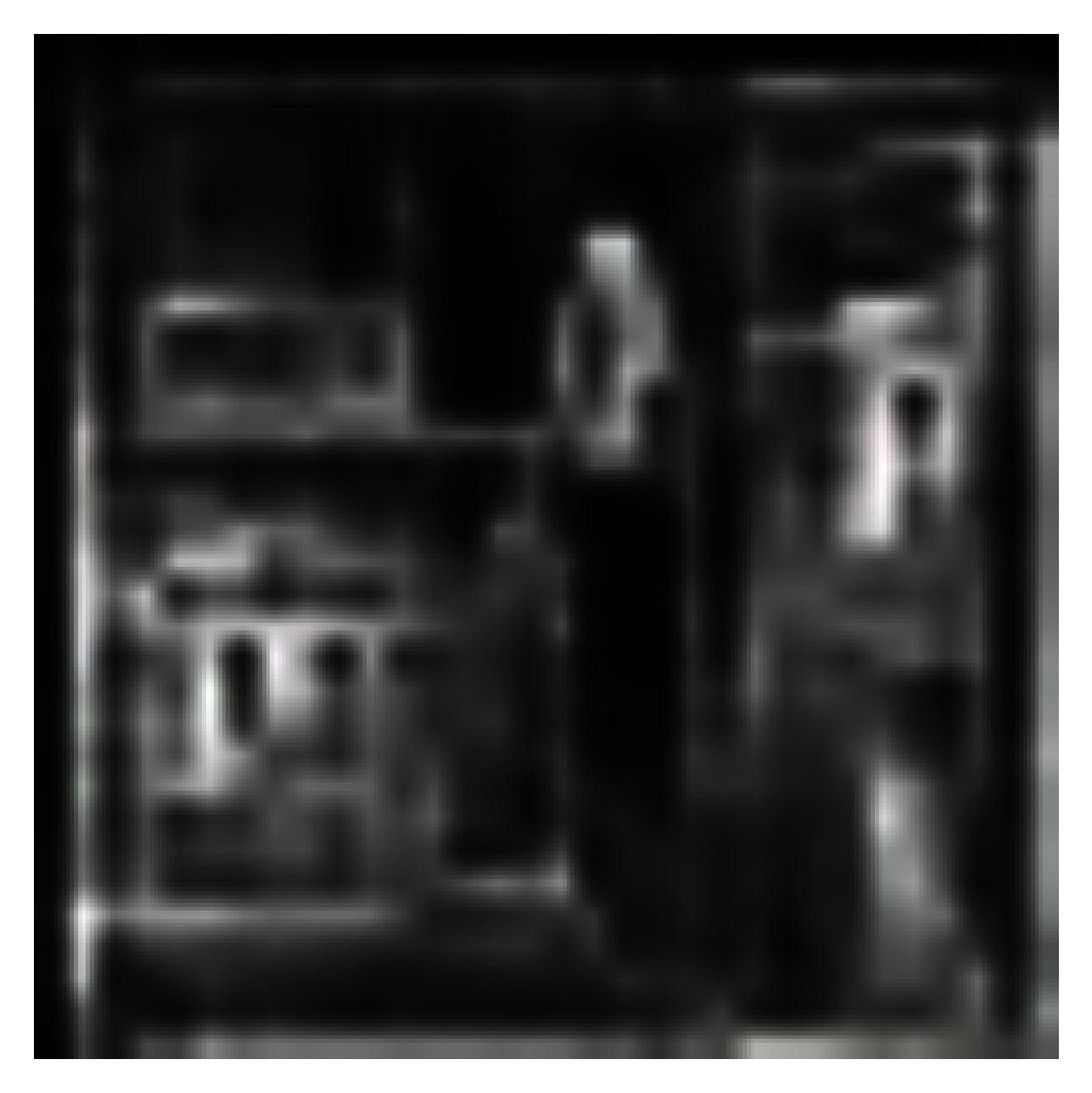} &
&
\adjustimage{height=1.5cm,valign=m}{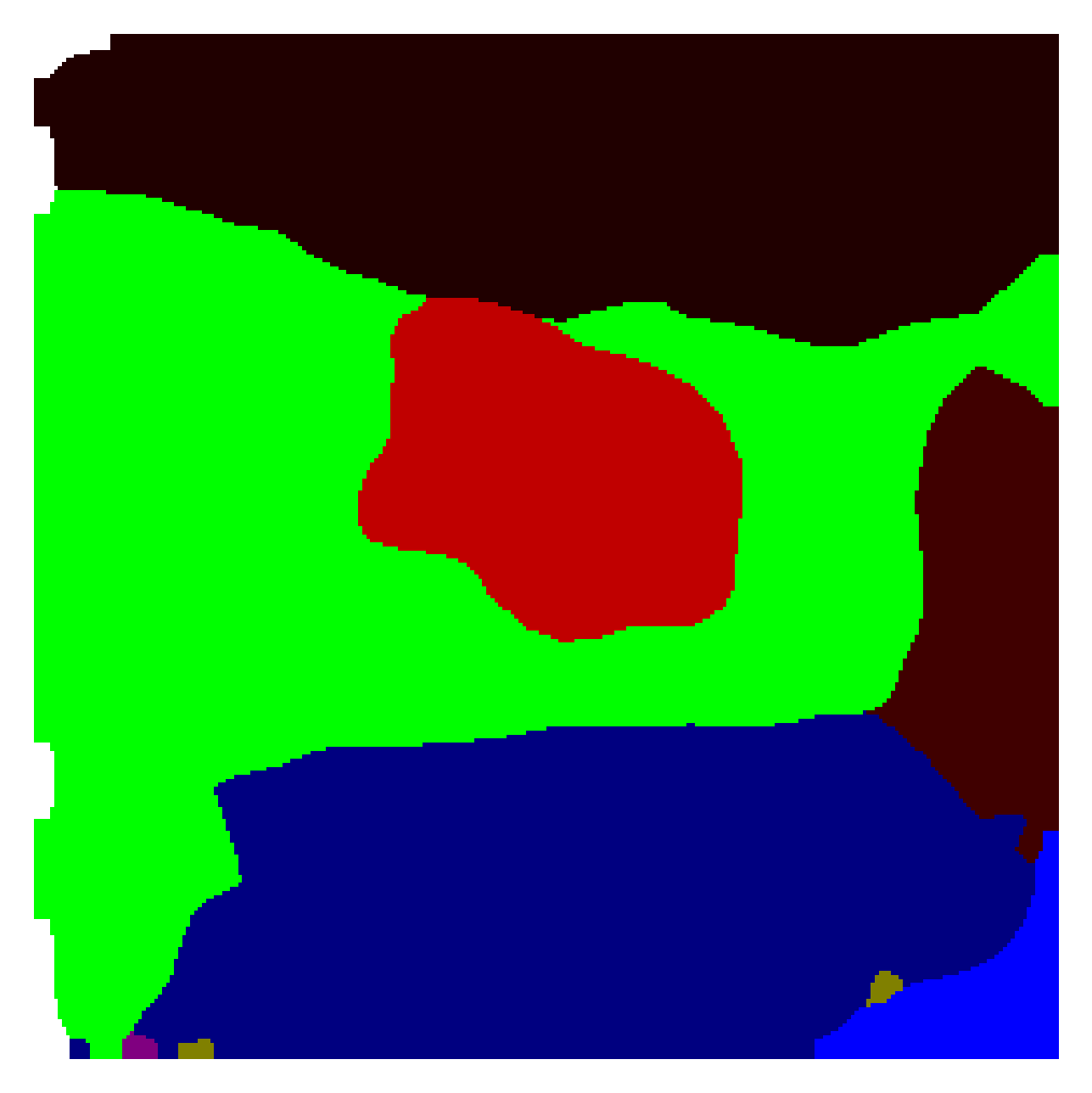} & 
\adjustimage{height=1.5cm,valign=m}{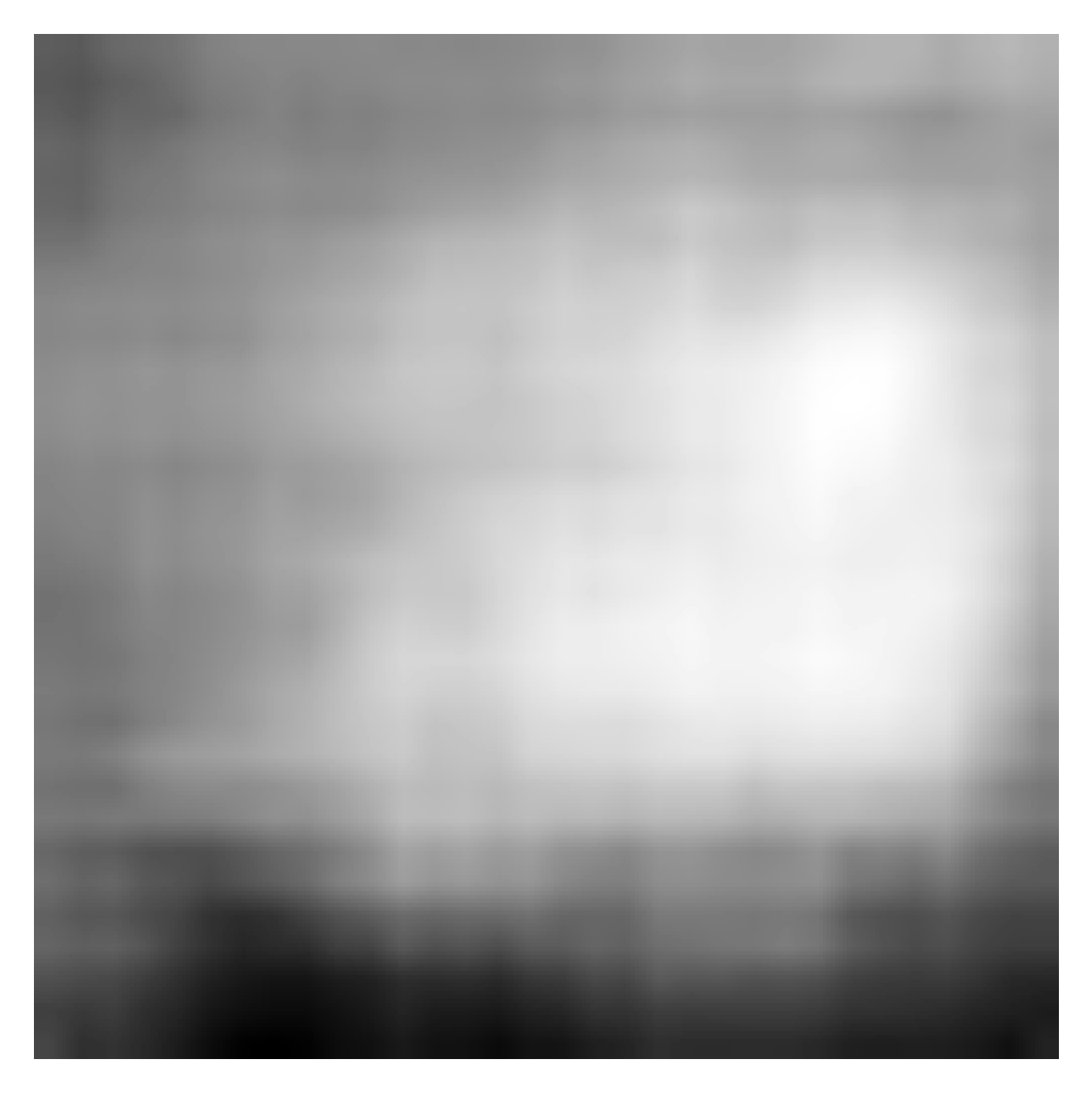} & \adjustimage{height=1.5cm,valign=m}{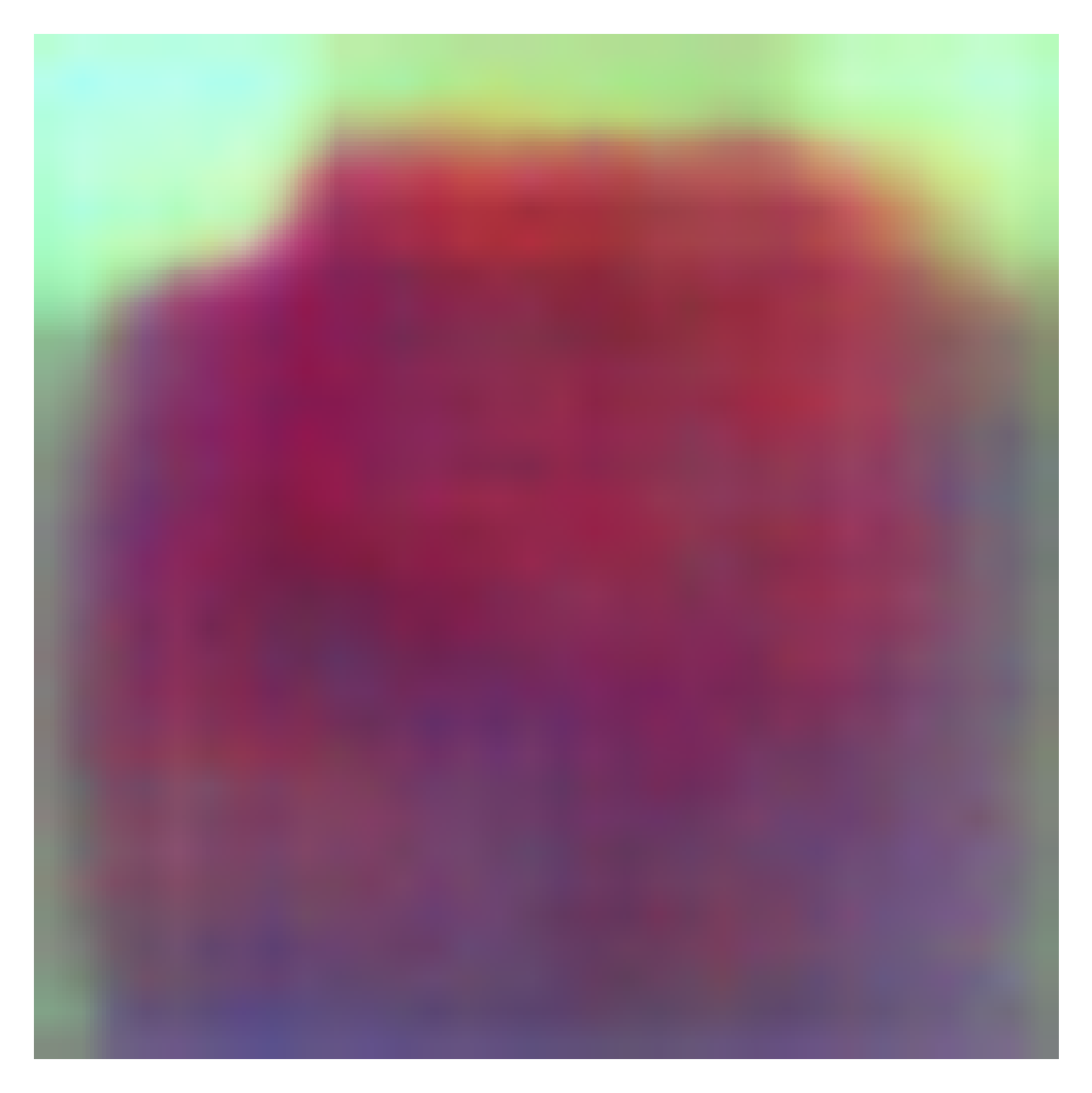}& \adjustimage{height=1.5cm,valign=m}{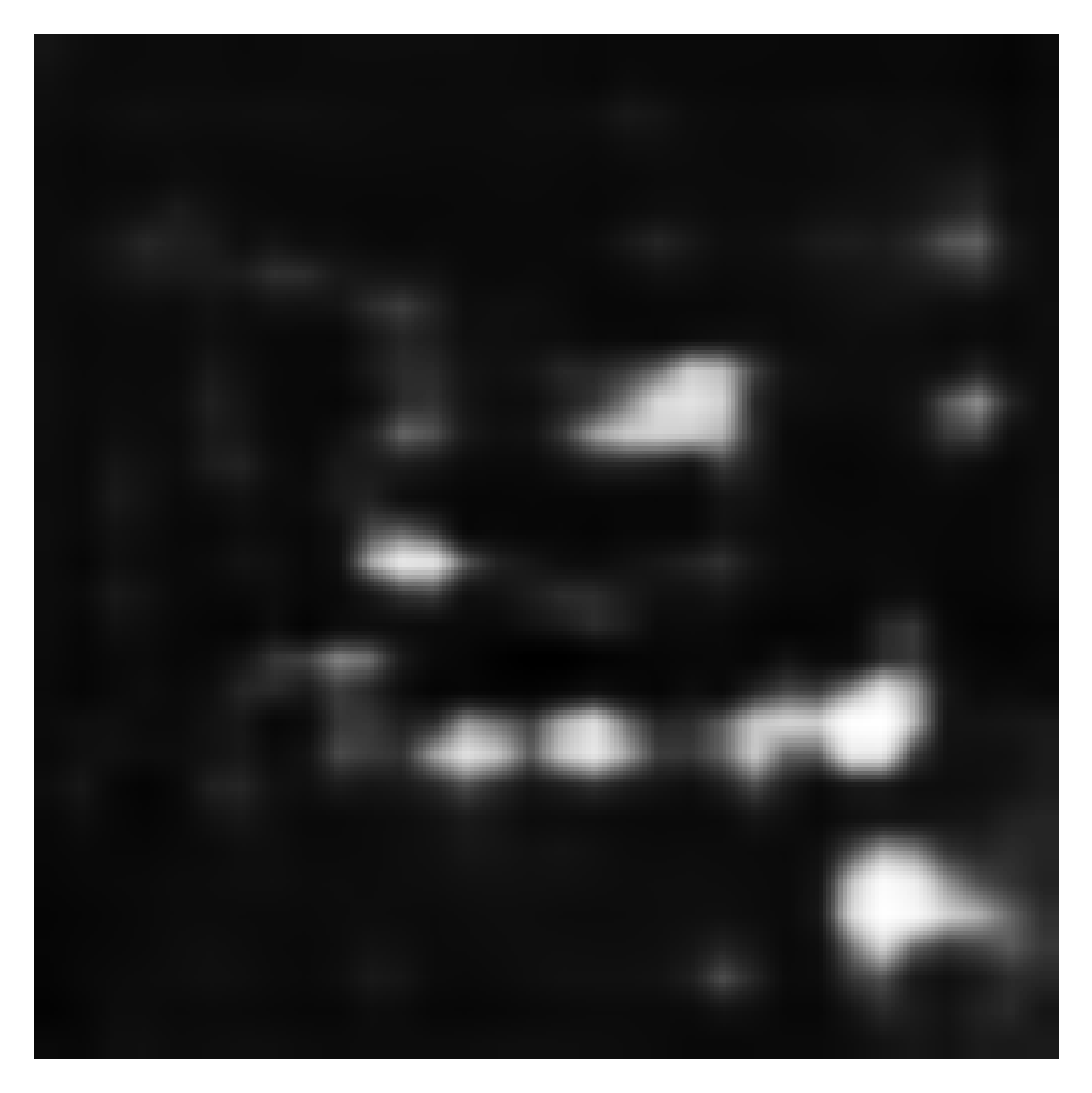}\\
\footnotesize{\rotatebox[origin=c]{90}{\ac{MTL}}} & &
\adjustimage{height=1.5cm,valign=m}{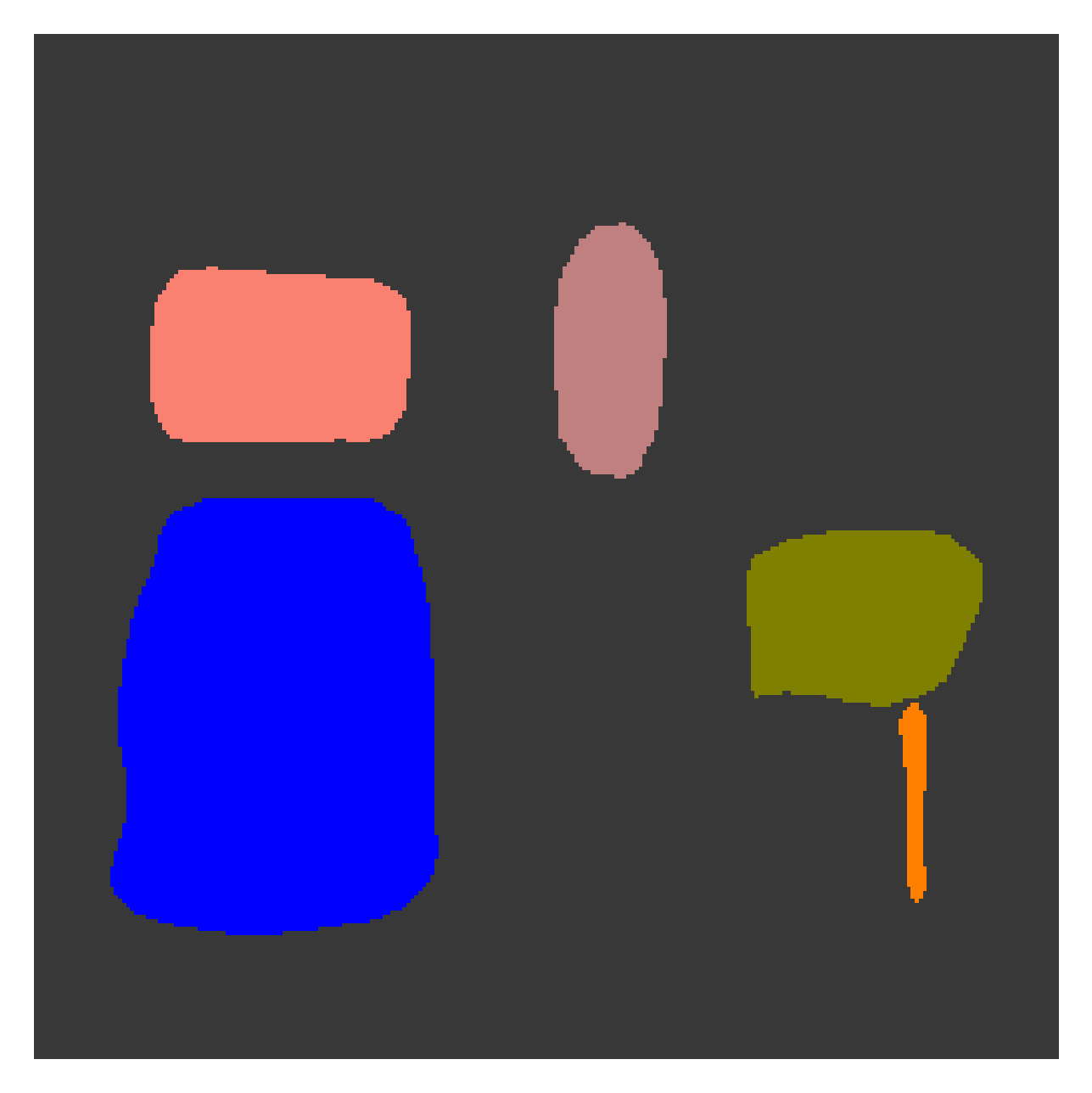} & \adjustimage{height=1.5cm,valign=m}{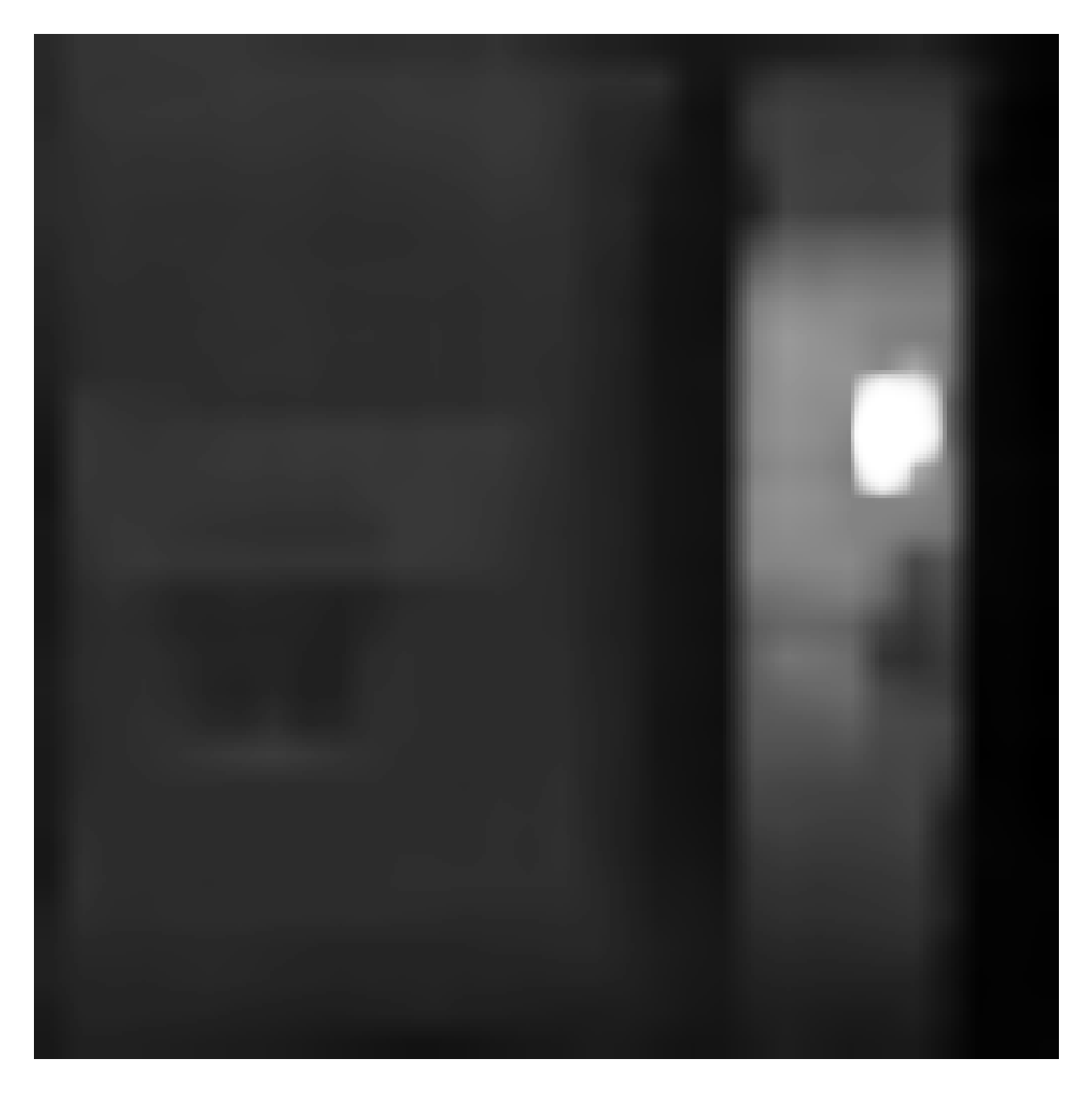} & \adjustimage{height=1.5cm,valign=m}{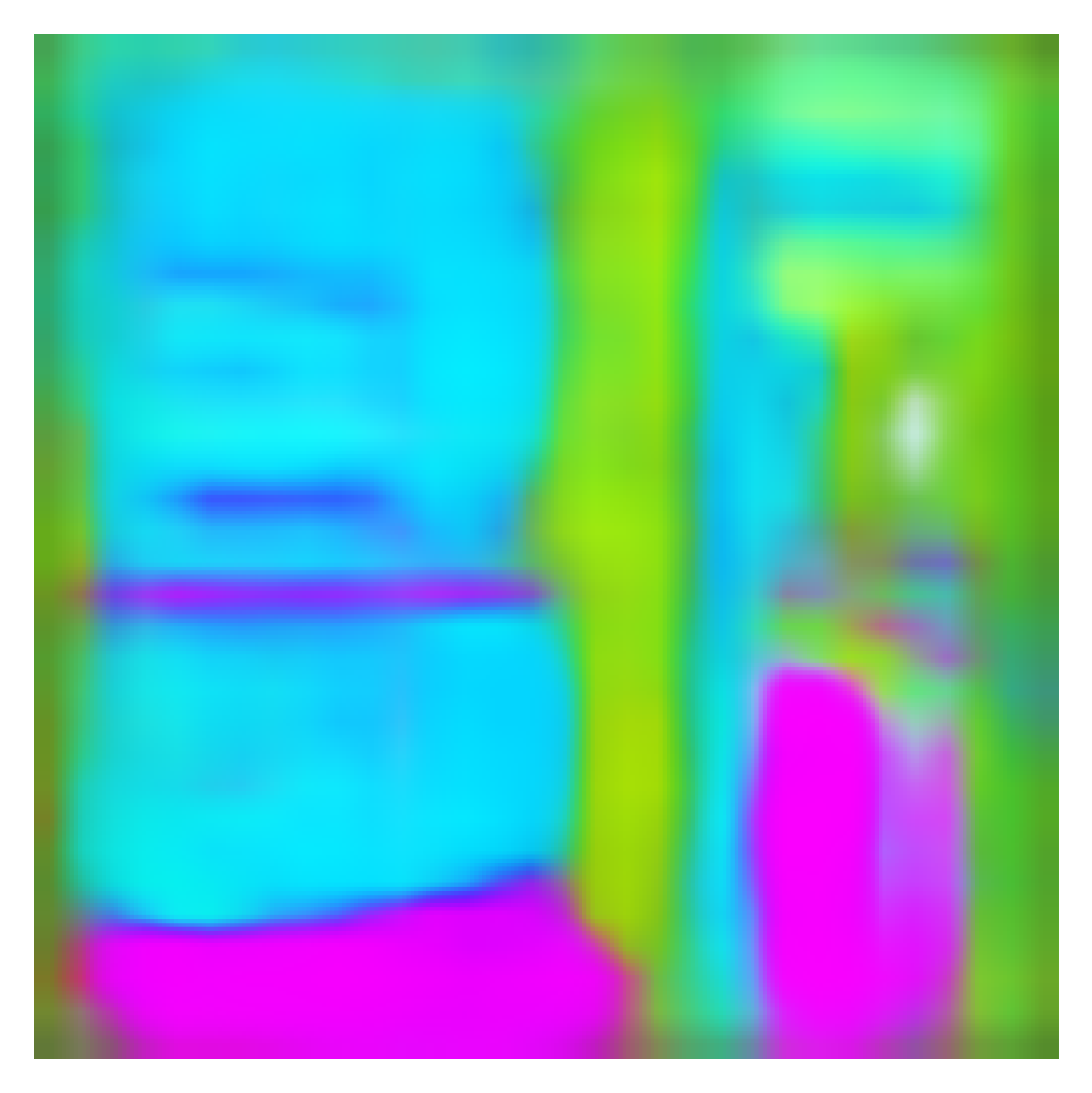} & \adjustimage{height=1.5cm,valign=m}{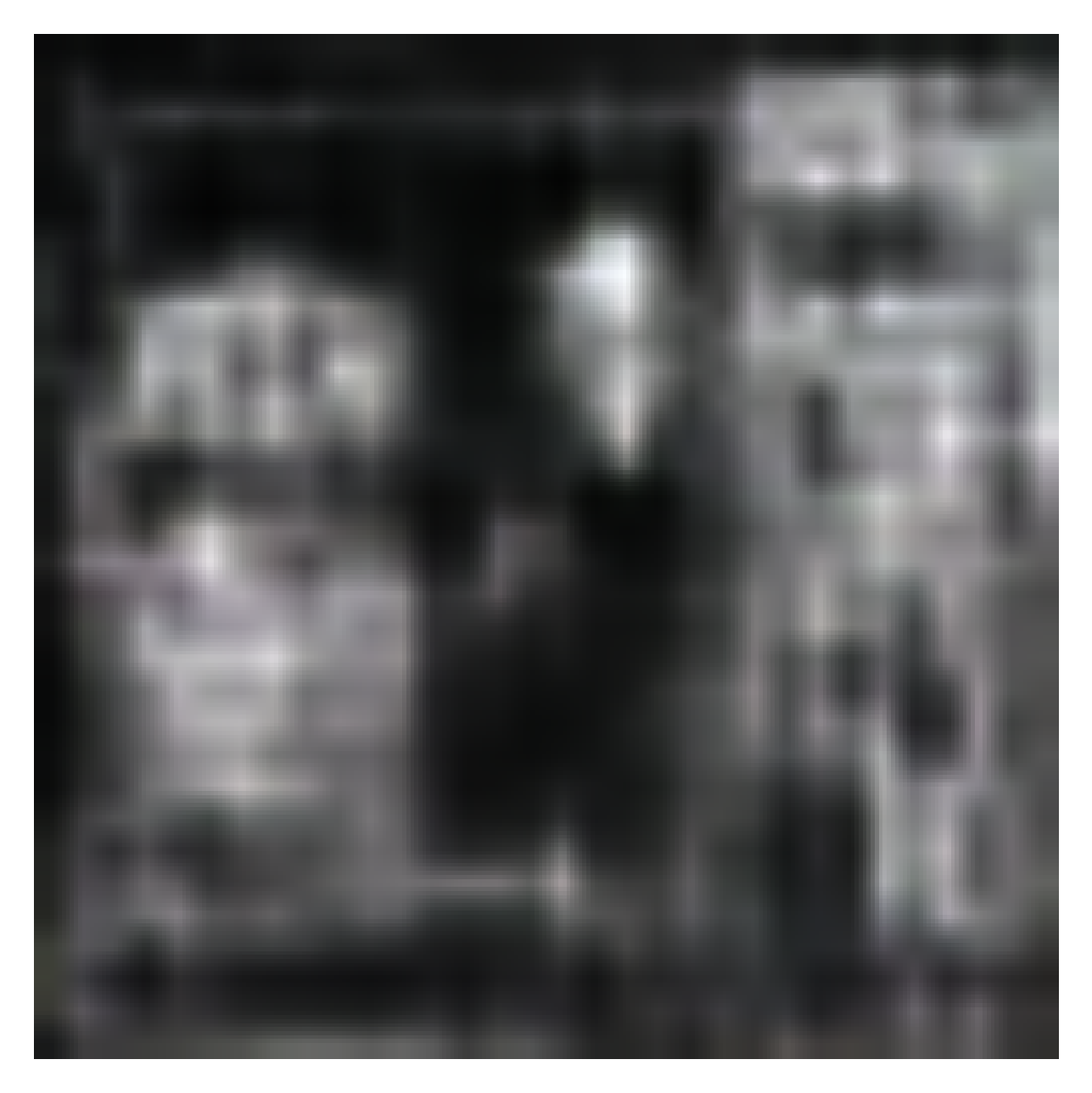} &
&
\adjustimage{height=1.5cm,valign=m}{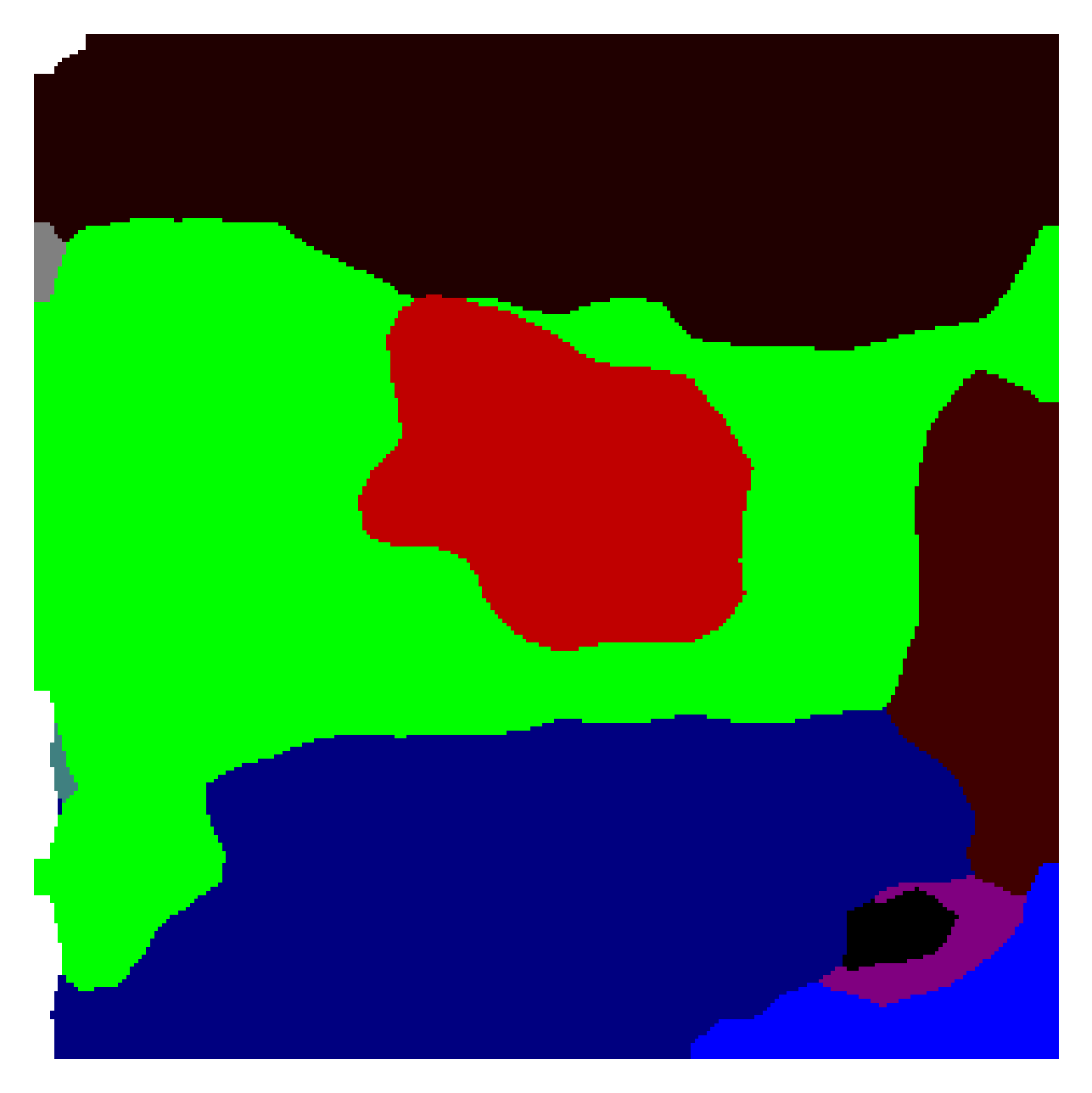} & 
\adjustimage{height=1.5cm,valign=m}{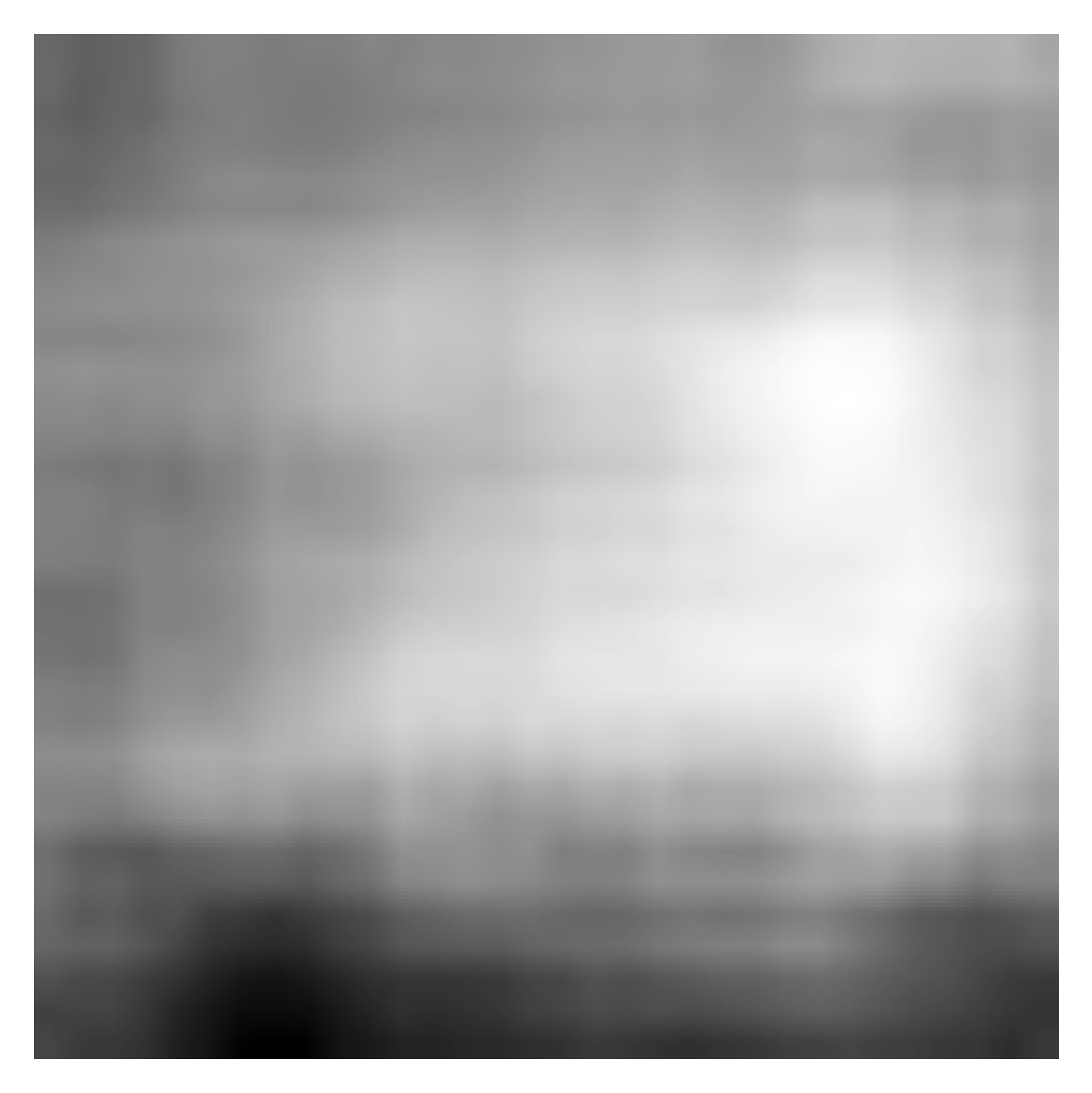} & \adjustimage{height=1.5cm,valign=m}{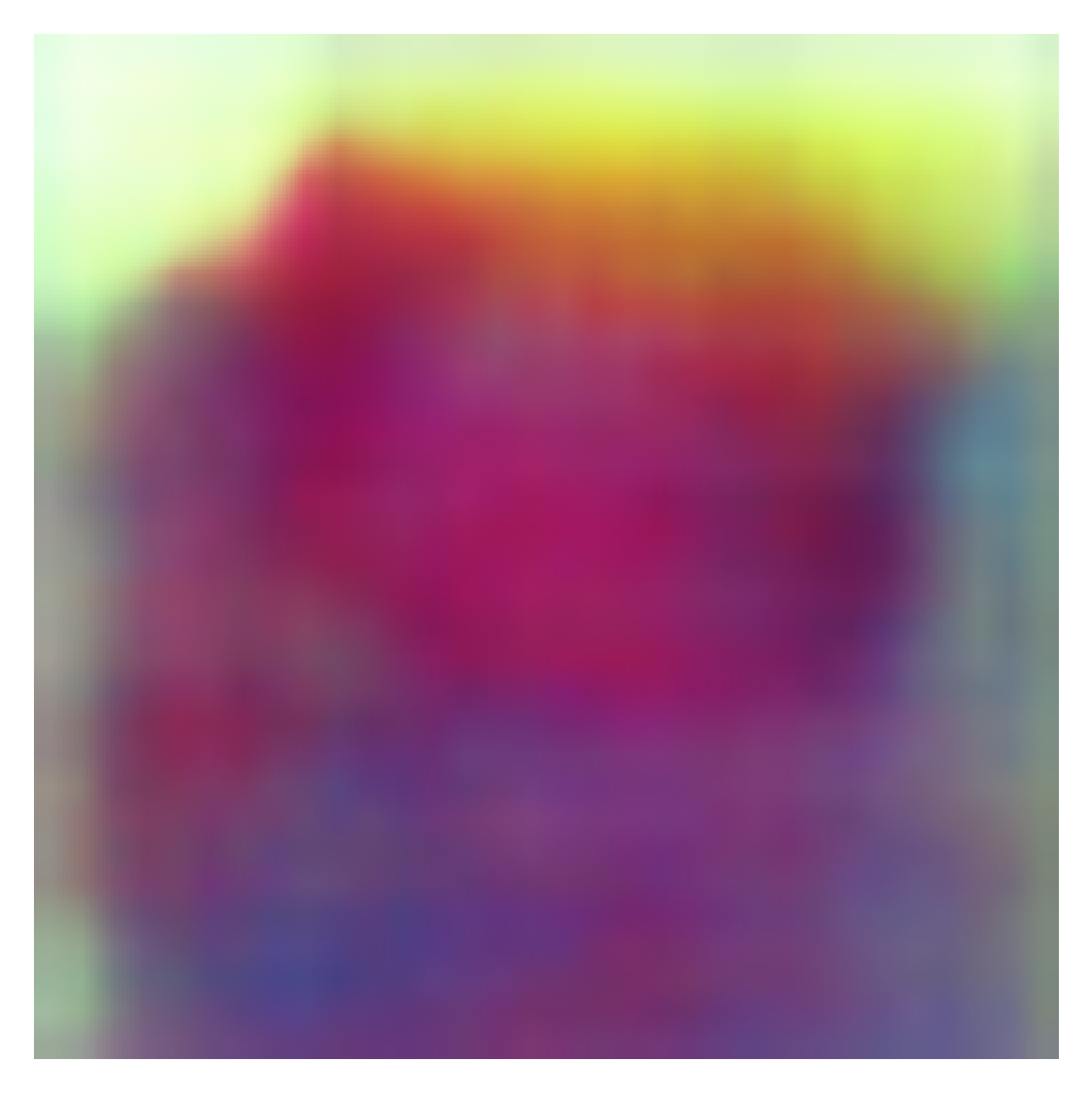}& \adjustimage{height=1.5cm,valign=m}{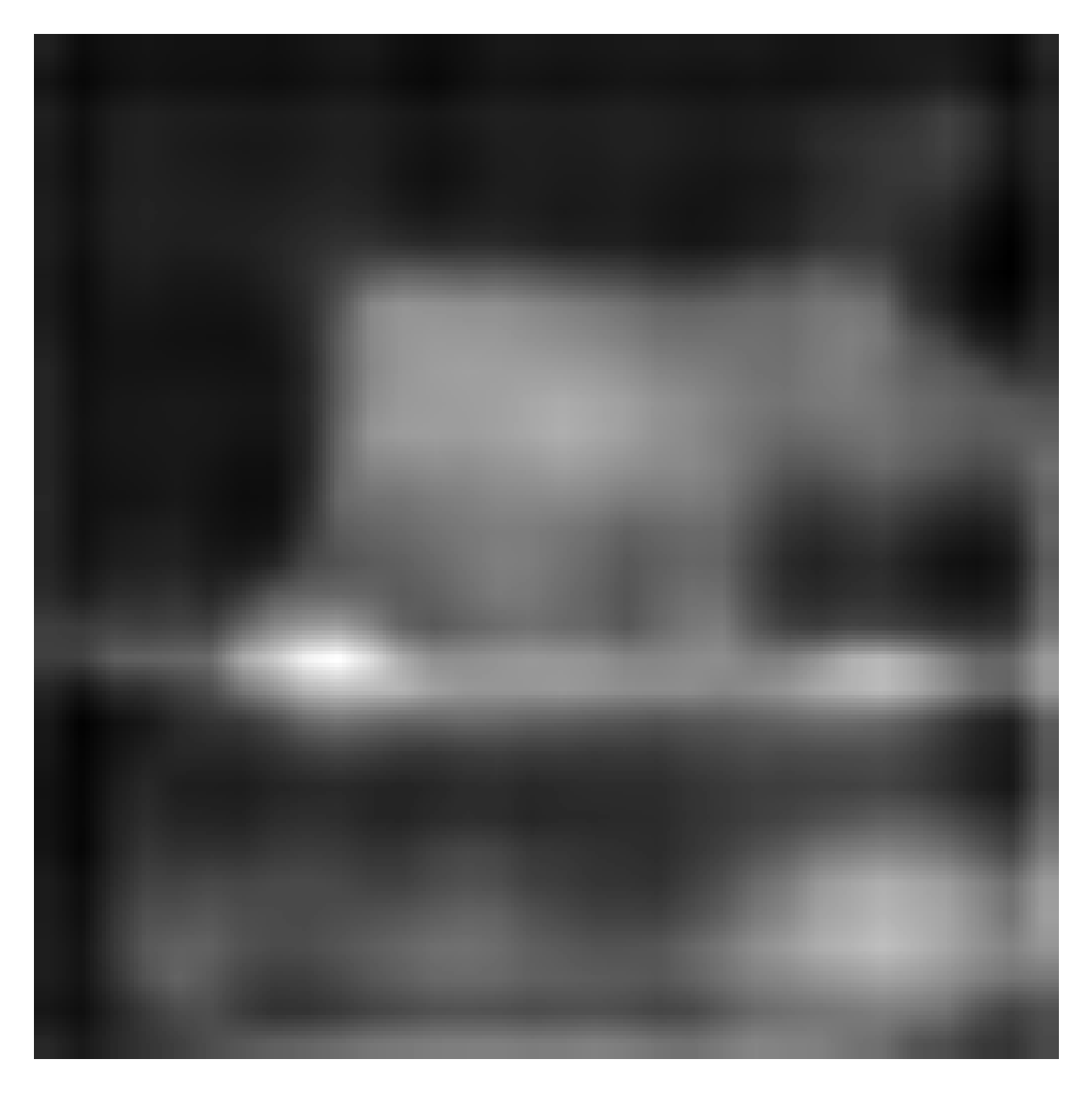}\\
\footnotesize{\rotatebox[origin=c]{90}{\ac{MTML}}} & &
\adjustimage{height=1.5cm,valign=m}{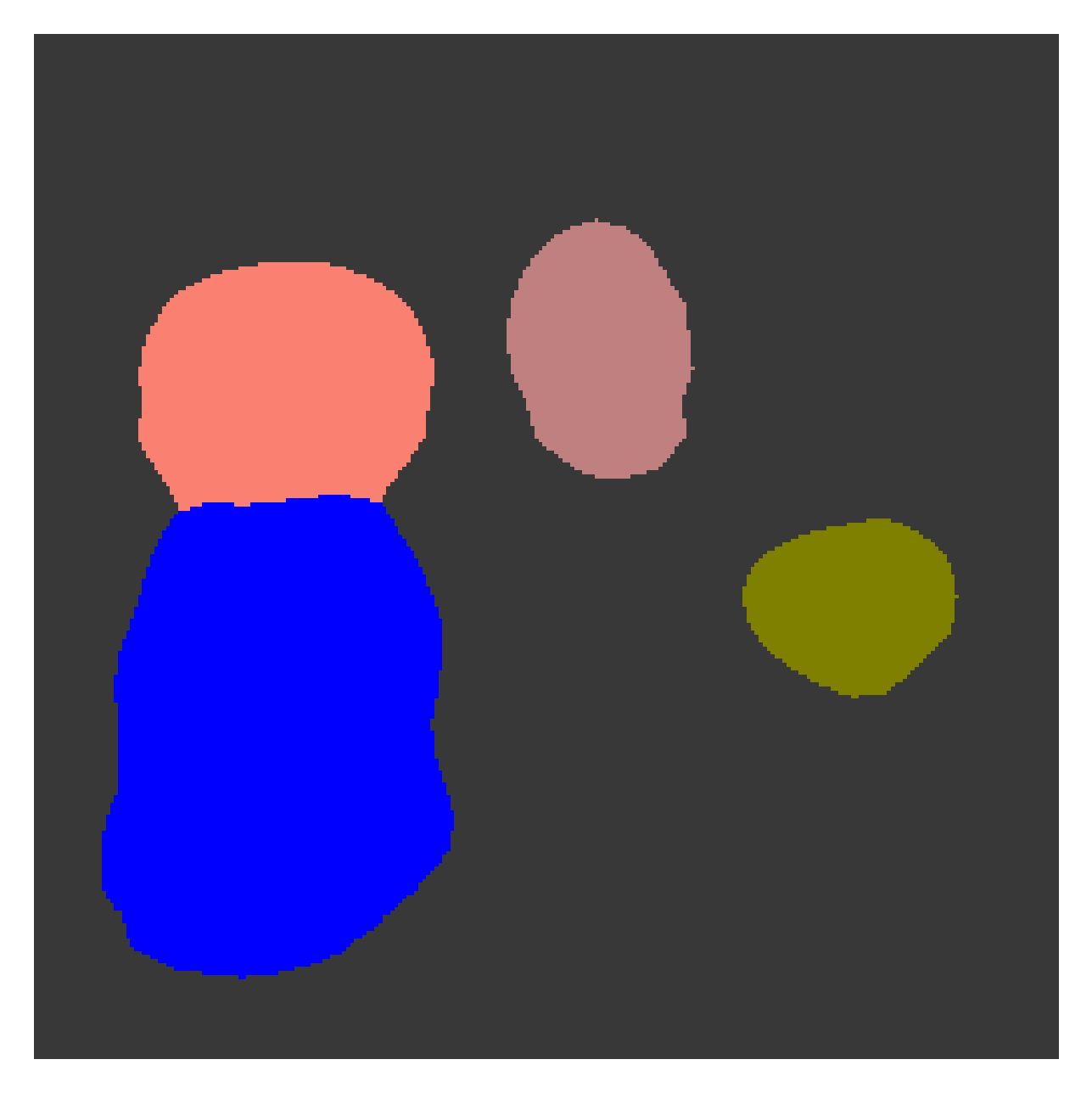} & \adjustimage{height=1.5cm,valign=m}{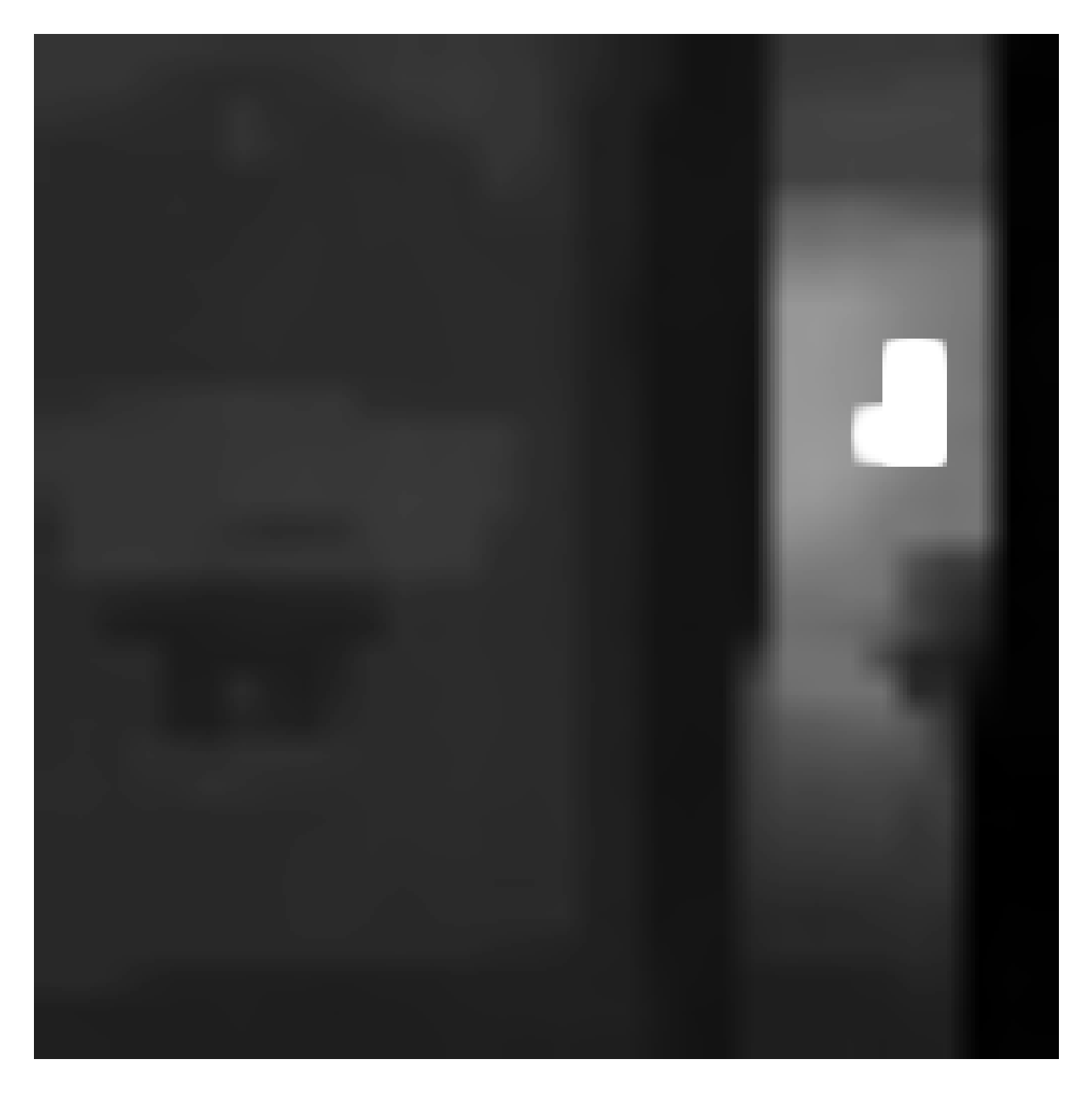} & \adjustimage{height=1.5cm,valign=m}{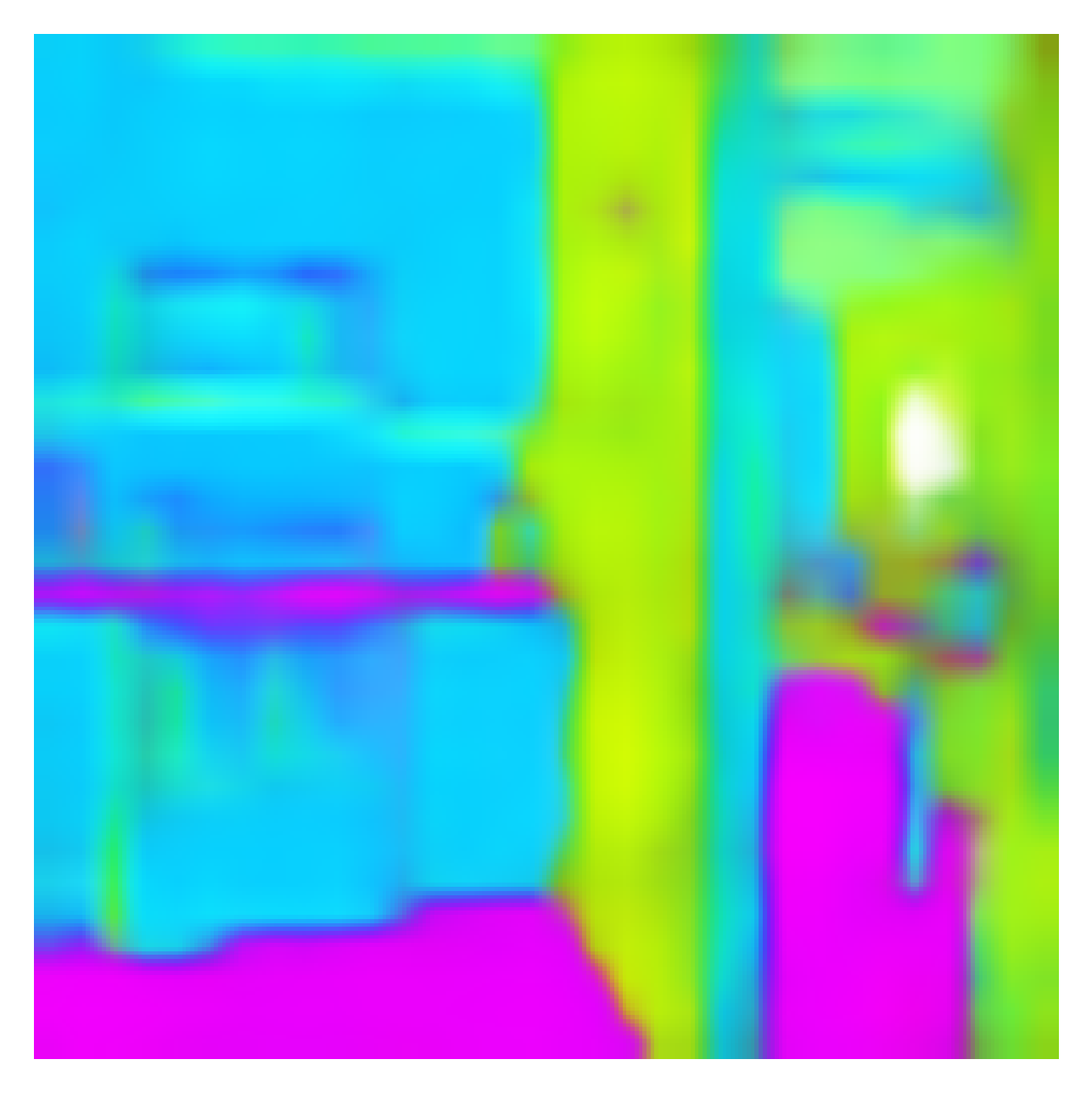} & \adjustimage{height=1.5cm,valign=m}{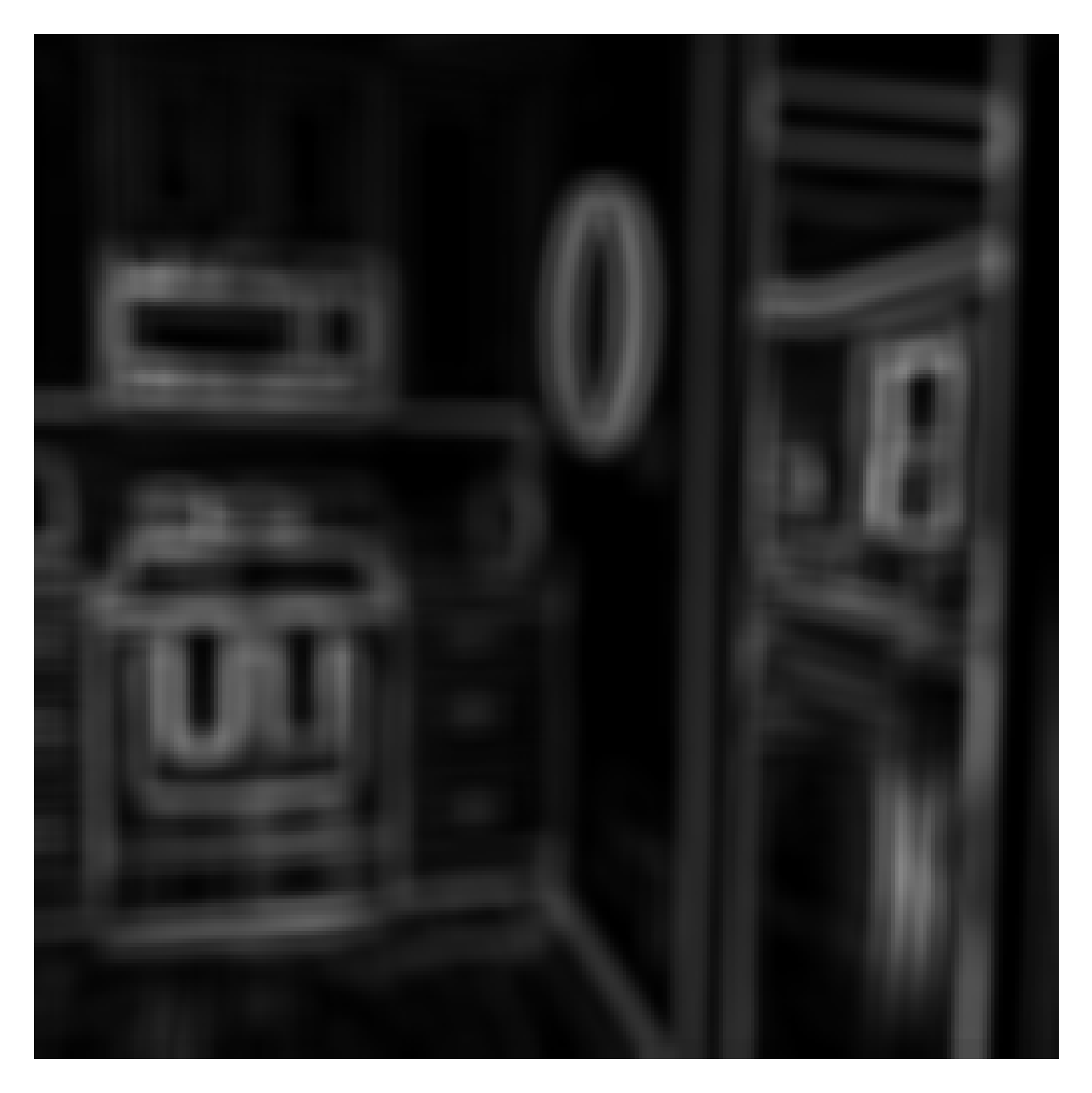} &
&
\adjustimage{height=1.5cm,valign=m}{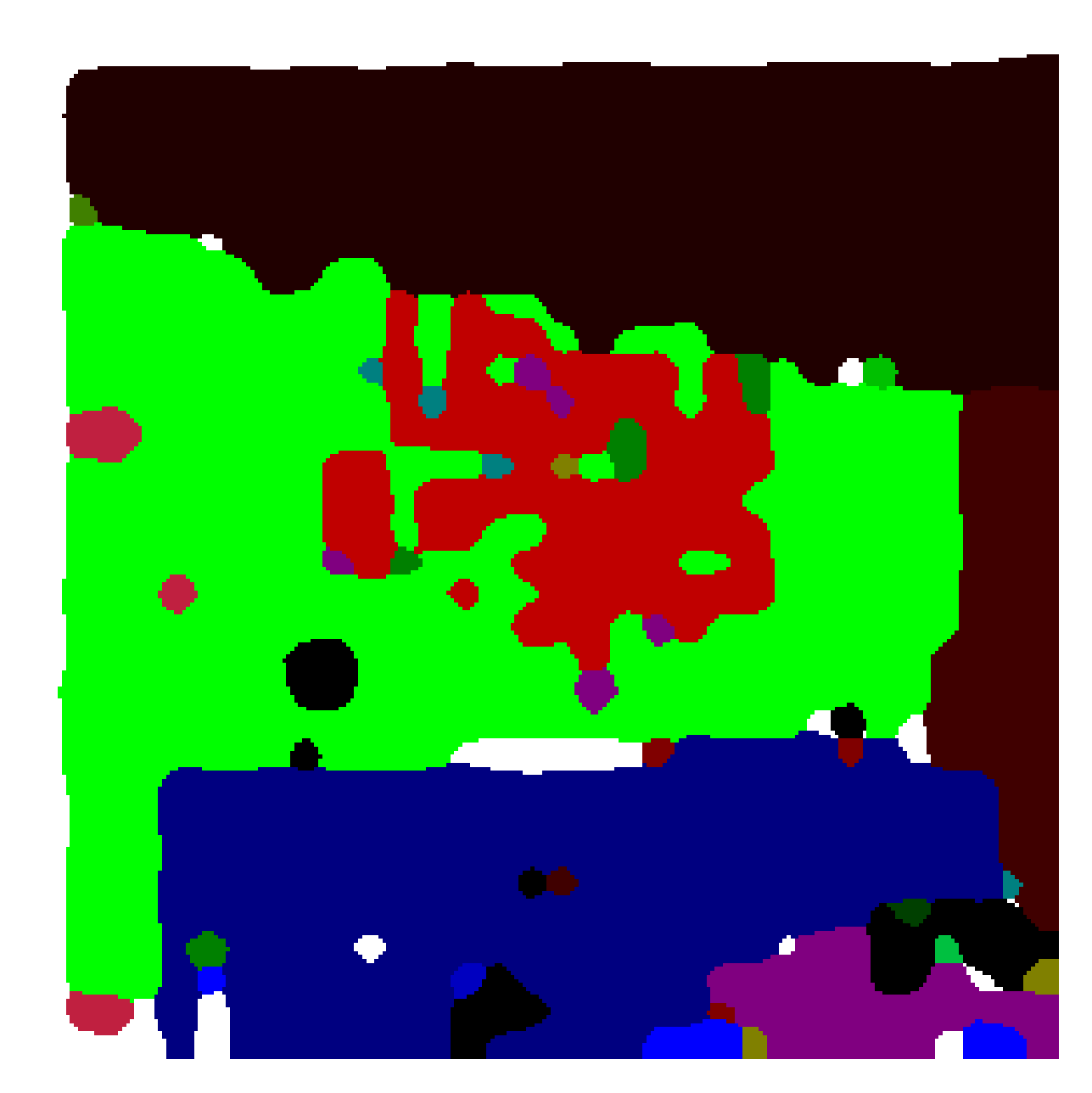} & 
\adjustimage{height=1.5cm,valign=m}{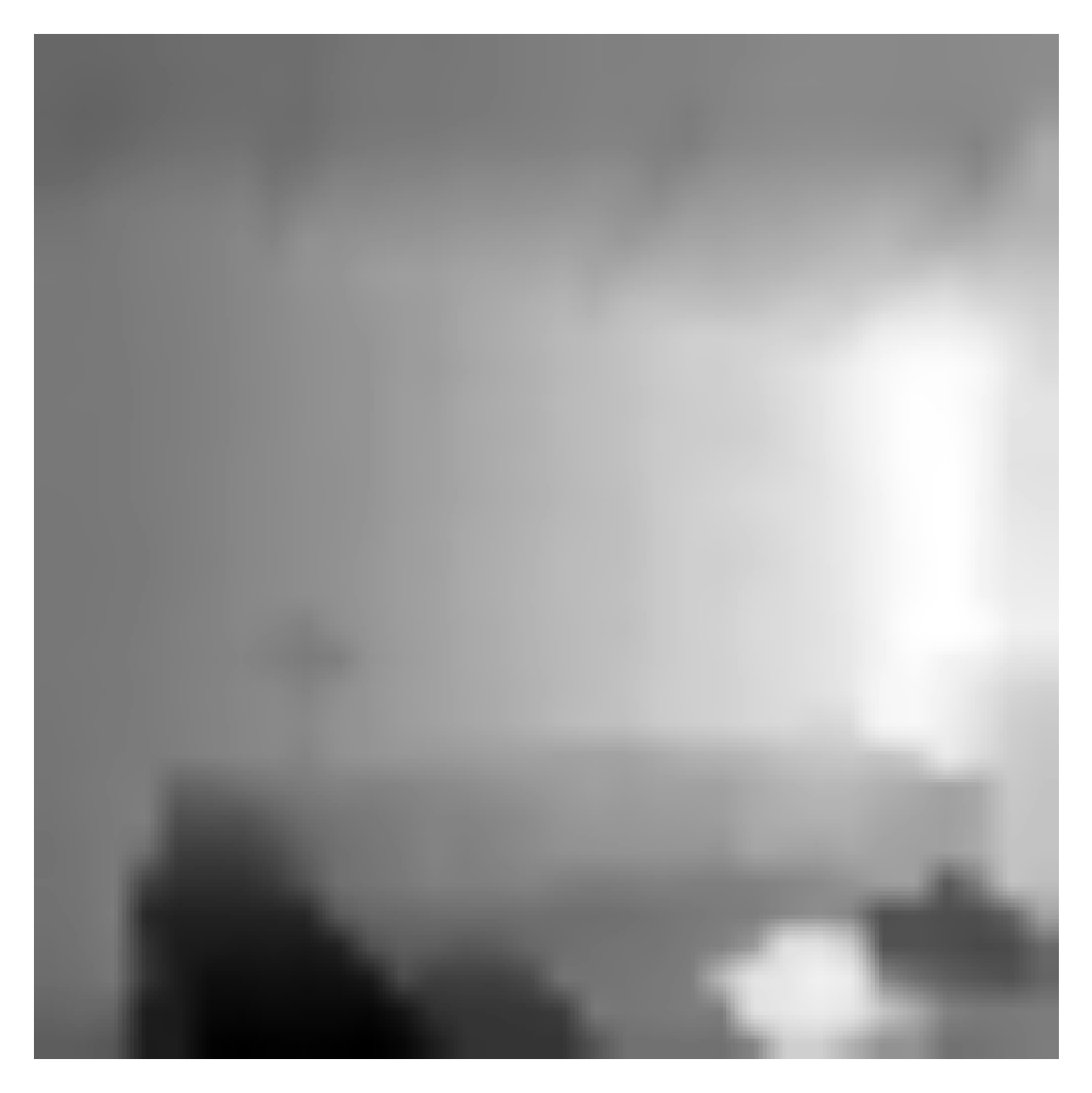} & \adjustimage{height=1.5cm,valign=m}{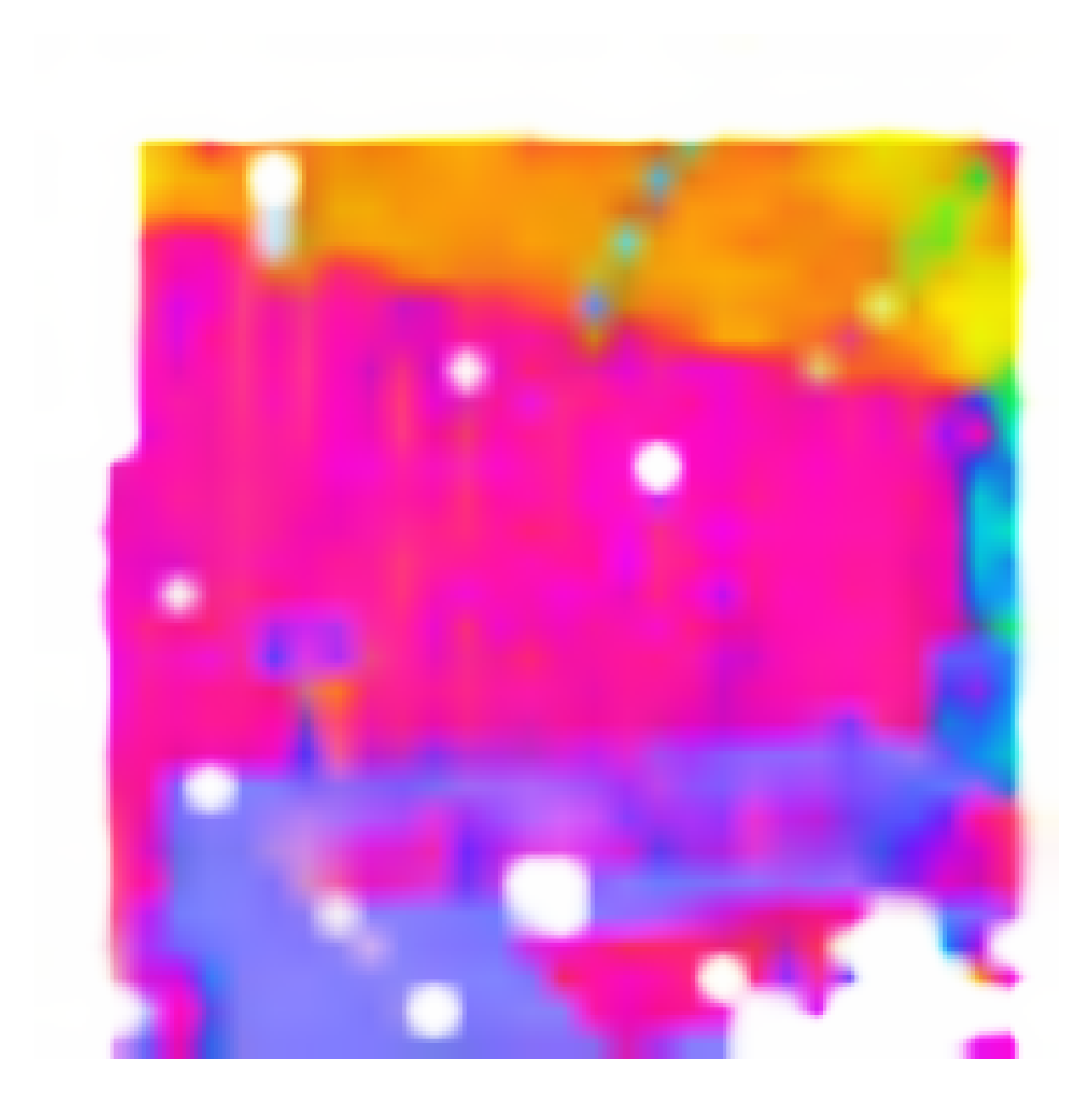}& \adjustimage{height=1.5cm,valign=m}{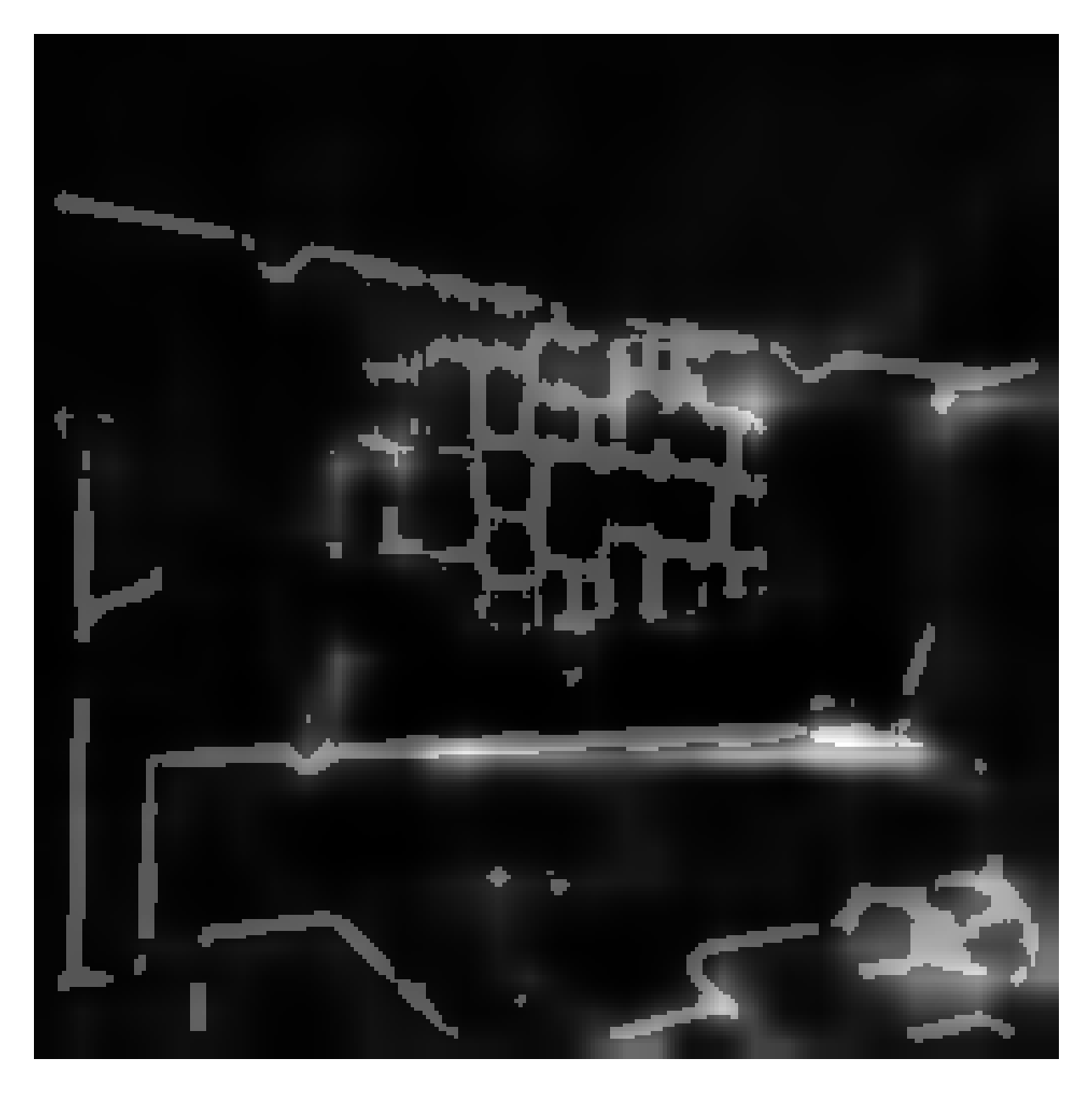}\\
\end{tabular}    
\end{center}
\vspace{-0.4cm}
\caption{The figure illustrates sample images of the input, its corresponding ground truths, single task (Exp. 1), \ac{MTL} (Exp. 2.3) and \ac{MTML} (Exp. 4.4) outputs for semantic segmentation (Seg.), depth estimation (Depth), surface normal estimation (SN), and edge detection (Edge) for both the NYU-v2 and taskonomy datasets.} 
\vspace{-1.5em}
\label{fig:img_mat}
\end{figure*}

\textbf{Data sets and tasks:} For the performance analysis of our proposed approach, two publicly available data sets are used: the NYU-v2 data set \cite{Silberman:ECCV12} and the tiny taskonomy data set \cite{zamir2018taskonomy}.
Both of these data sets contain a large variety of indoor scene images in standard 3-channel RGB image format.
Four tasks used in this work are: semantic segmentation ($T_1$), depth estimation ($T_2$), surface normal estimation ($T_3$), and edge detection ($T_4$). 
The train-validation-test data split for the NYU-v2 and taskonomy datasets is given in \cite{sun2020adashare} and \cite{zamir2018taskonomy}, respectively.

\textbf{Network architecture:} A very commonly used multi-task architecture is used for this work: a shared backbone network, followed by task-specific heads.
The common backbone in the network allows sharing of the low and mid-level features through their model parameters, while specific high-level features are learned by the task specific heads \cite{vandenhende2021multi}.
In this work, dilated ResNet-50 \cite{Yu2017} is employed as the backbone, which gives the shared representations of the input RGB image. 
These representations are then fed as inputs to the task heads or task specific network.
For all the four tasks, DeepLab V3 \cite{chen2017rethinking} network is used as the task heads, which make use of atrous convolutions\footnote{Atrous convolution comes from the french, \emph{convolution à trous} which could be translated as \emph{sparse kernel convolution}.} \cite{giusti2013fast, holschneider1990real}. 
The Atrous Spatial Pyramid Pooling (ASPP) module used in DeepLab v3 architecture helps to extract dense feature maps by discarding the down-sampling in the last layers and employing up-sampling in the filters of the corresponding layers.
This makes it suitable for pixel-level dense prediction tasks. 

\begin{table*}[ht]
\centering
\caption{Performance of the single task, \ac{MTL}, and \ac{MTML} approaches introduced in this work on the NYU-v2 and taskonomy datasets (across the three learning paradigms, p $<$ 0.05 for all the tasks), along with some previous state-of-the-art works in the literature. In this table, MAE is mean absolute error, CE denotes cross entropy metric, mIoU represents mean intersection over union metric and CS stands for cosine similarity metric. The top-2 results are highlighted in bold.
}
\vspace{-0.2cm}
\label{tab:sota}
\resizebox{\textwidth}{!}{%
\begin{tabular}{cc|cccc|cccccccc}
\hline
\textbf{} & \textbf{} & \multicolumn{4}{c|}{\textbf{Taskonomy}} & \multicolumn{8}{c}{\textbf{NYU}} \\ \hline
\multicolumn{1}{c|}{\textbf{Model}} & \textbf{Parameters} & \textbf{Segmen-} & \textbf{Depth} & \textbf{Surface} & \textbf{Edge} & \textbf{Segmen-} & \textbf{Depth} & \multicolumn{5}{c}{\textbf{Surface}} & \textbf{Edge} \\ 
\multicolumn{1}{c|}{\textbf{}} & \textbf{} & \textbf{tation} & \textbf{Estimation} & \textbf{Normal} & \textbf{Detection} & \textbf{tation} & \textbf{Estimation} & \multicolumn{5}{c}{\textbf{Normal}} & \textbf{Detection} \\
\multicolumn{1}{c|}{\textbf{}} & \textbf{} & \textbf{T1} & \textbf{T2} & \textbf{T3} & \textbf{T4} & \textbf{T1} & \textbf{T2} & \multicolumn{5}{c}{\textbf{T3}} & \textbf{T4} \\ \cline{9-13}
\multicolumn{1}{c|}{\textbf{}} & \textbf{} & \textbf{} & \textbf{} & \textbf{} & \textbf{} & \textbf{} & \textbf{} & \multicolumn{2}{c}{Error $\downarrow$} & \multicolumn{3}{c}{Theta $\uparrow$} & \textbf{} \\\hline

\multicolumn{1}{c|}{} & in Millions & CE $\downarrow$ & MAE $\downarrow$ & CS $\uparrow$ & MAE $\downarrow$ & mIoU $\uparrow$ & MAE $\downarrow$ & Mean & Median & $11.25^{\circ}$ & $22.5^{\circ}$ & $30^{\circ}$ & MAE $\downarrow$ \\ \hline
\multicolumn{1}{c|}{Multi-Task \cite{sun2020adashare}} & 41 & 0.587 & 0.024 & 0.696 & 0.203 & 24.1 & 0.58 & 16.6 & 13.4 & 42.5 & 73.2 & 84.6 & - \\
\multicolumn{1}{c|}{Adashare \cite{sun2020adashare}} & 41 & 0.566 & 0.025 & 0.702 & 0.2 & 30.2 & 0.55 & 16.6 & \textbf{12.9} & \textbf{45} & 71.7 & 83 & - \\
\multicolumn{1}{c|}{Cross Stitch \cite{misra2016cross}} & 124 & 0.56 & 0.022 & 0.679 & 0.217 & 24.5 & 0.58 & 17.2 & 14 & 41.4 & 70.5 & 82.9 & - \\
\multicolumn{1}{c|}{MTAN \cite{liu2019end}} & 114 & 0.637 & 0.023 & 0.687 & 0.206 & 26 & 0.57 & 16.6 & 13 & 43.7 & 73.3 & 84.4 & - \\
\multicolumn{1}{c|}{NDDR CNN \cite{gao2019nddr}} & 133 & \textbf{0.539} & 0.024 & 0.7 & 0.203 & 21.6 & 0.66 & 17.1 & 14.5 & 37.4 & \textbf{73.7} & 85.6 & - \\
\multicolumn{1}{c|}{Sluice \cite{ruder2019latent}} & 124 & 0.596 & 0.024 & 0.695 & 0.207 & 23.8 & 0.58 & 17.2 & 14.4 & 38.9 & 71.8 & 83.9 & - \\
\multicolumn{1}{c|}{Learn2branch \cite{guo2020learning}} & 51 & \textbf{0.462} & \textbf{0.018} & 0.709 & 0.136 & \multicolumn{7}{c}{No results on NYU dataset} & \multicolumn{1}{l}{} \\ \hline
\multicolumn{1}{c|}{\ac{MTL} (Ours)} & 88 & 0.628 & \textbf{0.013} & \textbf{0.930}& \textbf{0.050}& \textbf{42.25} & \textbf{0.12} & \textbf{15.04} & 16.06 & 42.24 & 72.52 & \textbf{87.72} & \textbf{0.16} \\
\multicolumn{1}{c|}{\ac{MTML} (Ours)} & 88 & 2.300 & 0.063 & \textbf{0.931} & \textbf{0.046} & \textbf{41.51} & \textbf{0.10} & \textbf{13.34} & \textbf{10.24} & \textbf{52.40} & \textbf{76.17} &\textbf{88.51} & \textbf{0.10} \\ \hline
\end{tabular}%
}
\end{table*}

\textbf{Training specifications:} To train the above architecture, an input RGB image of size 256 $\times$ 256 is normalized and fed to the backbone network in batches of 64 images.
A minimal learning rate of 0.00001 is considered, to administer sufficient training for all the tasks, since some tasks are harder than others. 
Task-specific early stopping (patience = 35) and overall early stopping (patience = 50) are employed to avoid overfitting. 
For single task experiments, multi-task experiments and the meta update ( \ie outer loop) in \ac{MTML}, AdamW \cite{adamw} is used as an optimizer since it decouples weight decay from gradient update by modifying Adam's \cite{adam} implementation of $L_2$ regularization. 
Due to large datasets like taskonomy and NYU-v2, \ac{MTML} only takes into account a single SGD \cite{sutskever2013importance} inner loop to reduce computational cost and memory usage.
The losses are cross-entropy loss for semantic segmentation, a combined depth loss \cite{hu2019revisiting} for depth estimation,  inverse cosine similarity loss for surface normal prediction, and Huber loss \cite{PAUL2022100218} for edge detection. 
To make an even comparison, the evaluation metrics used in this work are similar to those used in \cite{sun2020adashare, misra2016cross, liu2019end, gao2019nddr, ruder2019latent}, also mentioned in Table \ref{tab:sota}.
For semantic segmentation, cross-entropy (lower is better) and intersection over union (IoU, higher is better) are respectively used for the taskonomy and NYU-v2 datasets.
Mean absolute error (lower is better) is calculated for depth estimation on both data sets.
For the taskonomy data set, cosine similarity is used to evaluate surface normal prediction.
In contrast, for the NYU-v2 dataset, the mean and median of angular error (lower is better) between prediction and ground truth and percentage of pixels whose predicted values are within $11.25^{\circ}$, $22.5^{\circ}$, and $30^{\circ}$ \cite{sun2020adashare} (higher is better) of the ground truth are calculated. 
For edge detection mean absolute error (lower is better) is calculated between the prediction and ground truth for both data sets.
The loss and evaluation metric for semantic segmentation do not include background pixels, similar to the other works discussed in Table \ref{tab:sota}.
For fair comparison, the hyper-parameters were determined for the single task models to perform their best, and the multi-task models were then trained under similar setting. 

For combining the losses from all the tasks, there are many techniques as discussed by \cite{vandenhende2021multi}. In this work we have used uncertainty \cite{kendall2018multi} to balance the losses of the four dense prediction tasks. 
Because the focus of this work was on combining multi-task and meta learning, other task balancing techniques were not investigated.

In a \ac{MTML} arrangement, as discussed in Section \ref{formulation}, the network is trained using multi-task learning episodes. 
In episodes having less than four tasks, it is trained as usual, but for the task absent in the combination, the task loss is set to zero and the losses are combined as usual for backpropagation. The parameters of the corresponding task head are not updated by freezing the layers (same in case of task-wise early-stopping). 
To carry out a comparative performance analysis of all the experiments, they are evaluated using the same test set, and all the models are trained with the same hyper-parameters.
The models are trained on NVIDIA A100 Tensor Core GPUs, with 40 GB on-board HBM2 VRAM.
All experiments are repeated three times with different random seeds to ensure and evaluate the consistency of the model.
The results are shown in terms of mean and standard deviation.

\section{Experimental evaluations}\label{sec:diss_of_results}
\begin{figure*}[ht]
    \centering
    \includegraphics[width = \textwidth]{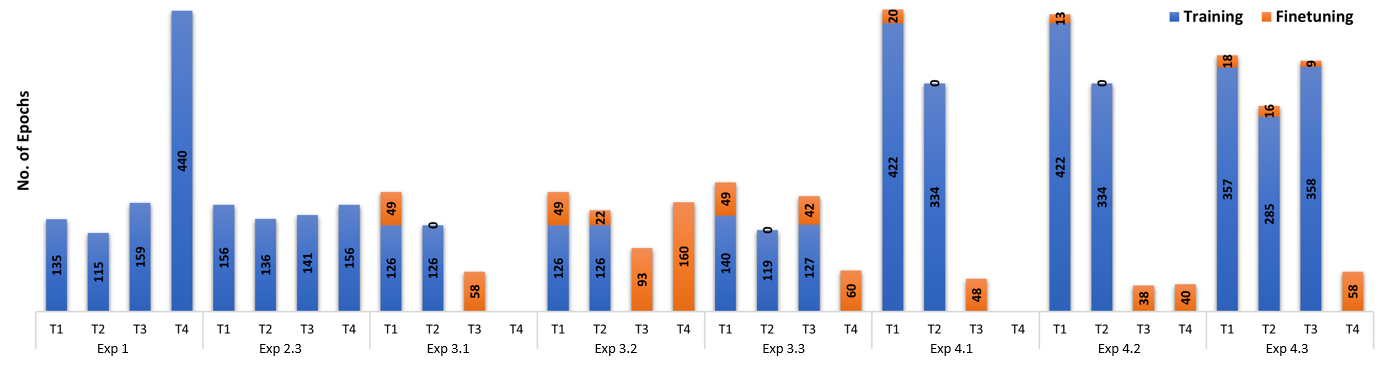}
    \caption{The charts demonstrate the number of epochs (gradient steps to train) for the single-task, multi-task, and \ac{MTML} for NYU-v2 dataset.The x-axis displays the tasks (T1 - T4) for the experiments mentioned in Table \ref{tab:NYU} The blue bar represents the number of training epochs, and the orange refers to the epochs required by an unseen task to fine-tune the already trained multi-task and \ac{MTML} model. The orange bar on top of the blue depicts the number of epochs the model is further trained during fine-tuning the unseen task. Similar pattern in the number of epochs is also accounted for the taskonomy dataset.}
    \label{fig:epochs}
    \vspace{-0.6cm}
\end{figure*}

To analyze the performance of the proposed methods for various task combinations, the experiments listed in Table \ref{tab:NYU} (NYU-v2) and \ref{tab:taskonomy} (taskonomy) were designed.
Experiments are classified into three types: single task learning, multi-task learning, and \ac{MTML}. 
A comprehensive analysis of the results of these experiment is provided in this section.
These experiments compare the effect of adding a new task to an already-trained single task, multi-task and meta multi-task network.
To test the various task combinations, two formats are considered: augmenting tasks (Exp. 2, 3, and 4), and leave-one-out (Exp. 5, and 7).
In these experiments, `+Tn' indicates the addition of new $n$th task to a network previously trained on other tasks.
Fig. \ref{fig:epochs} compares the average number of epochs needed to train the single task, \ac{MTL} and \ac{MTML} models.
\begin{table*}[t]
\centering
\caption{Test set evaluation results for a single task, multi-task, and \ac{MTML} experiments on the NYU-v2 dataset}
\label{tab:NYU}
\begin{tiny}
\resizebox{0.97\textwidth}{!}{%
\begin{tabular}{lllcccccccc}
\hline
\textbf{Exp. No.} &
  \multicolumn{2}{c}{\textbf{Tasks Involved}} &
  \multicolumn{8}{c}{\textbf{Tasks}} \\ \cmidrule(lr){2-3} \cmidrule(lr){4-11} 
\textbf{} &
  \multicolumn{1}{c}{\textbf{Trained}} &
  \multicolumn{1}{c}{\textbf{Tested}} &
  \textbf{T1} &
  \textbf{T2} &
  \multicolumn{5}{c}{\textbf{T3}} &
  \textbf{T4} \\ 
 &
   &
  \multicolumn{1}{c}{( +Tn is fine-tuning} &
  \textbf{Segmentation} &
  \textbf{Depth Estimation} &
  \multicolumn{5}{c}{\textbf{Surface Normal}} &
  \textbf{Edge Detection} \\
 &
   &
  \multicolumn{1}{c}{on $\mathrm{n^{th}}$ new task)} &
  mIoU $\uparrow$ &
  Mean abs.error $\downarrow$ &
  \multicolumn{2}{c}{Error $\downarrow$} &
  \multicolumn{3}{c}{Theta $\uparrow$} &
  Mean abs.error $\downarrow$\\ \cmidrule(lr){6-10} 
 &
   &
   &
   &
   &
  mean &
  median &
  $11.25^{\circ}$ &
  $22.5^{\circ}$&
  $30^{\circ}$&
  \\ \hline
\textbf{1} &
  \multicolumn{10}{c}{\textbf{Single task learning}} \\\hline
 &
   &
   &
  42.53±0.083 &
  0.11±0.000 &
  15.88±0.510 &
  13.97±0.524 &
  41.62±1.514 &
  73.20±1.878 &
  88.56±0.780 &
  0.15±0.010 \\ \hline
\textbf{2} &
  \multicolumn{10}{c}{\textbf{Multi-task learning}} \\ \hline
2.1 &
  T1, T2 &
  T1, T2 &
  42.38±0.123 &
  0.11±0.001 &
 -  &
 -  &
 -  &
 -  &
 -  &
 -  \\
2.2 &
  T1, T2, T3 &
  T1, T2, T3 &
  42.36±0.335 &
  0.11±0.001 &
  15.52 ± 0.574 &
  13.55±0.632 &
  43.47±1.908 &
  73.01±1.530 &
  88.01±0.942 & -
   \\
2.3 &
  T1, T2, T3, T4 &
  T1, T2, T3, T4 &
  42.25±0.141 &
  0.12±0.002 &
  15.04 ± 0.769 &
  16.06±3.060 &
  42.24±0.725 &
  72.52±0.918 &
  87.72±1.716 &
  0.16±0.033 \\ \hline
\textbf{3} &
  \multicolumn{10}{c}{\textbf{Multi-task learning, addition of new task}} \\ \hline
3.1 &
  T1, T2 &
  T1, T2 (+ T3) &
  42.41±0.299 &
  0.12±0.002 &
  15.18±0.209 &
  13.37±0.185 &
  44.18±0.406 &
  73.22±0.725 &
  88.31±0.580 &
  - \\
3.2 &
  T1, T2 &
  T1, T2 (+ T3, T4) &
  42.48±0.258 &
  0.11±0.001 &
  14.84±0.230 &
  12.90±0.318 &
  45.42±0.932 &
  74.37±0.427 &
  88.46±0.271 &
  0.20±0.029 \\
3.3 &
  T1, T2, T3 &
  T1, T2, T3 (+T4) &
  42.50±0.063 &
  0.11±0.001 &
  14.84±0.552 &
  12.75±0.646 &
  45.86±1.802 &
  74.56±1.485 &
  88.34±0.412 &
  0.24±0.010 \\ \hline

\textbf{4} &
  \multicolumn{10}{c}{\textbf{\ac{MTML}, meta testing phase involves addition of new task}} \\ \hline
4.1 &
  T1, T2 &
  T1, T2 (+ T3) &
  37.09±0.355 &
  0.11±0.002 &
  13.84±0.254 &
  11.89±0.370 &
  48.02±1.238 &
  78.54±0.456 &
  90.51±0.307 &
 -  \\
4.2 &
  T1, T2 &
  T1, T2 (+ T3, T4) &
  37.06±0.163 &
  0.11±0.001 &
  14.28±0.635 &
  12.47±0.715 &
  46.37±2.238 &
  77.04±1.526 &
  90.19±0.251 &
  0.17±0.048 \\
4.3 &
  T1, T2, T3 &
  T1, T2, T3 (+T4) &
  39.59±1.580 &
  0.11±0.004 &
  13.82±0.385 &
  11.36±0.593 &
  49.64±1.594 &
  76.26±0.279 &
  89.02±0.143 &
  0.15±0.074 \\ 
  4.4 & T1, T2, T3, T4 & T1, T2, T3, T4 & 41.41±0.111 & 0.10±0.001 & 13.34±0.025 & 10.24±0.010 & 52.40±0.005 & 76.17±0.130 & 88.51±0.095 & 0.10±0.000 \\ \hline

\textbf{5} &
  \multicolumn{10}{c}{\textbf{Multi-task learning, leave one task out format}} \\ \hline
5.1 &
  T2, T3, T4 &
  T2, T3, T4 &
  - & 
  0.14±0.027 &
  15.27±0.444 &
  13.48±0.553 &
  43.26±1.824 &
  73.96±1.336 &
  89.05±0.343 &
  0.18±0.014 \\
5.2 &
  T1, T3, T4 &
  T1, T3, T4 &
  42.52±0.102 &
  -  &
  15.30±0.439 &
  13.41±0.600 &
  43.57±2.031 &
  73.98±0.749 &
  88.57±0.410 &
  0.12±0.004 \\
5.3 &
  T1, T2, T4 &
  T1, T2, T4 &
  42.04±0.142 &
  0.12±0.003 &
  - &
  - &
  - &
  - &
  - &
  0.12±0.006 \\
5.4 &
  T1, T2, T3 &
  T1, T2, T3 &
  42.36±0.335 &
  0.11±0.001 &
  15.52±0.574 &
  13.55±0.632 &
  43.47±1.908 &
  73.01±1.530 &
  88.01±0.942 &
  - \\ \hline
\textbf{6} &
  \multicolumn{10}{c}{\textbf{Multi-task learning, leave one task out format, addition of the left out task}} \\ \hline
6.1 &
  T2, T3, T4 &
  T2, T3, T4 (+ T1) &
  42.36±0.108 &
  0.12±0.000 &
  14.78±0.078 &
  12.86±0.096 &
  45.47±0.191 &
  74.82±0.523 &
  88.25±0.306 &
  0.16±0.008 \\
6.2 &
  T1, T3, T4 &
  T1, T3, T4 (+T2) &
  43.05±0.042 &
  0.12±0.000 &
  15.14±0.340 &
  13.21±0.389 &
  44.06±1.280 &
  74.65±0.982 &
  89.12±0.029 &
  0.12±0.002 \\
6.3 &
  T1, T2, T4 &
  T1, T2, T4 (+T3) &
  42.49±0.239 &
  0.11±0.000 &
  14.60±0.105 &
  12.79±0.185 &
  45.43±0.706 &
  76.01±0.368 &
  89.73±0.306 &
  0.12±0.004 \\
6.4 &
  T1, T2, T3 &
  T1, T2, T3 (+T4) &
  42.50±0.063 &
  0.11±0.001 &
  14.84±0.552 &
  12.75±0.646 &
  45.86±1.802 &
  74.56±1.485 &
  88.34±0.412 &
  0.24±0.010 \\ \hline
\textbf{7} &
  \multicolumn{10}{c}{\textbf{\ac{MTML}, leave one task out format, addition of left out task in meta testing}} \\ \hline
7.1 &
  T2, T3, T4 &
  T2, T3, T4 (+ T1) &
  45.05±0.536 &
  0.10±0.002 &
  13.44±0.008 &
  10.76±0.078 &
  51.23±0.185 &
  76.65±0.174 &
  89.18±0.091 &
  0.10±0.001 \\
7.2 &
  T1, T3, T4 &
  T1, T3, T4 (+T2) &
  38.30±0.298 &
  0.11±0.002 &
  13.63±0.021 &
  11.03±0.083 &
  50.55±0.212 &
  76.33±0.196 &
  88.97±0.140 &
  0.10±0.002 \\
7.3 &
  T1, T2, T4 &
  T1, T2, T4 (+T3) &
  38.90±0.457 &
  0.10±0.000 &
  13.59±0.080 &
  11.55±0.147 &
  49.23±0.456 &
  78.31±0.419 &
  90.24±0.227 &
  0.10±0.001 \\
7.4 &
  T1, T2, T3 &
  T1, T2, T3 (+T4) &
  39.59±1.580 &
  0.10±0.004 &
  13.37±0.244 &
  10.70±0.348 &
  51.40±0.899 &
  76.95±0.705 &
  89.38±0.383 &
  0.11±0.019 \\ \hline
\end{tabular}%
}
\end{tiny}
\vspace{-0.3cm}
\end{table*}
\begin{table*}[ht]
\centering
\caption{Test set evaluation results for a single task, multi-task, and \ac{MTML} experiments on the taskonomy dataset}
\label{tab:taskonomy}
\begin{tiny}
\resizebox{0.75\textwidth}{!}{%
\begin{tabular}{lllcccc}
\hline
\textbf{Exp. No.} &
  \multicolumn{2}{c}{\textbf{Tasks Involved}} &
  \multicolumn{4}{c}{\textbf{Tasks}} \\ \cmidrule(lr){2-3} \cmidrule(lr){4-7}
\textbf{} &
  \multicolumn{1}{c}{\textbf{Trained}} &
  \multicolumn{1}{c}{\textbf{Tested}} &
  \textbf{T1} &
  \textbf{T2} &
  \textbf{T3} &
  \textbf{T4} \\ 
 &
   &
  \multicolumn{1}{c}{( +Tn is fine-tuning} &
  \textbf{Segmentation} &
  \textbf{Depth Estimation} &
  \textbf{Surface Normal} &
  \textbf{Edge Detection} \\
 &
   &
  \multicolumn{1}{c}{on $\mathrm{n^{th}}$ new task)} &
  Cross-entropy $\downarrow$ &
  Mean abs.error $\downarrow$&
  Cosine similarity $\uparrow$&
  Mean abs.error $\downarrow$\\ \hline
\textbf{1} &
  \multicolumn{6}{c}{\textbf{Single task learning}} \\ \hline
 &
   &
   &
  0.491±0.025 &
  0.013±0.001 &
  0.931±0.001 &
  0.049±0.001 \\
 \hline
\textbf{2} &
  \multicolumn{6}{c}{\textbf{Multi-task learning}} \\ \hline
2.1 &
  T1, T2 &
  T1, T2 &
  0.650±0.076 &
  0.013±0.000 &
   &
   \\
2.2 &
  T1, T2, T3 &
  T1, T2, T3 &
  0.736±0.080 &
  0.013±0.000 &
  0.930±0.002 &
   \\
2.3 &
  T1, T2, T3, T4 &
  T1, T2, T3, T4 &
  0.628±0.025 &
  0.013±0.000 &
  0.930±0.001 &
  0.050±0.001 \\ \hline
\textbf{3} &
  \multicolumn{6}{c}{\textbf{Multi-task learning, addition of new task}} \\ \hline
3.1 &
  T1, T2 &
  T1, T2 (+ T3) &
  1.286±0.184 &
  0.012±0.000 &
  0.931±0.002 &
   \\
3.2 &
  T1, T2 &
  T1, T2 (+ T3, T4) &
  1.217±0.246 &
  0.013±0.000 &
  0.931±0.003 &
  0.050±0.000 \\
3.3 &
  T1, T2, T3 &
  T1, T2, T3 (+T4) &
  1.330±0.080 &
  0.014±0.000 &
  0.931±0.001 &
  0.049±0.000 \\ \hline

\textbf{4} &
  \multicolumn{6}{c}{\textbf{\ac{MTML}, meta testing phase involves addition of new task}} \\ \hline
4.1 &
  T1, T2 &
  T1, T2 (+ T3) &
  2.000±0.136 &
  0.014±0.001 &
  0.937±0.006 &
   \\
4.2 &
  T1, T2 &
  T1, T2 (+ T3, T4) &
  1.826±0.143 &
  0.014±0.000 &
  0.939±0.002 &
  0.050±0.001 \\
4.3 &
  T1, T2, T3 &
  T1, T2, T3 (+T4) &
  2.175±0.126 &
  0.014±0.000 &
  0.929±0.000 &
  0.049±0.000 \\
4.4 &
  T1, T2, T3, T4 &
  T1, T2, T3, T4 &
  2.300±0.044 &
  0.063±0.066 &
  0.931±0.001 &
  0.046±0.000 \\ \hline

\textbf{5} &
  \multicolumn{6}{c}{\textbf{Multi-task learning, leave one task out format}} \\ \hline
5.1 &
  T2, T3, T4 &
  T2, T3, T4 &
   - &
  0.013±0.001 &
  0.930±0.001 &
  0.051±0.001 \\
5.2 &
  T1, T3, T4 &
  T1, T3, T4 &
  0.639±0.038 &
   - &
  0.930±0.001 &
  0.052±0.003 \\
5.3 &
  T1, T2, T4 &
  T1, T2, T4 &
  0.658±0.089 &
  0.013±0.000 &
  - &
  0.048±0.001 \\
5.4 &
  T1, T2, T3 &
  T1, T2, T3 &
  0.736±0.080 &
  0.013±0.000 &
  0.930±0.002 & -
   \\ \hline
\textbf{6} &
  \multicolumn{6}{c}{\textbf{Multi-task learning, leave one task out format, addition of the left out task}} \\ \hline
6.1 &
  T2, T3, T4 &
  T2, T3, T4 (+ T1) &
  {\color[HTML]{444444} 0.707±0.100} &
  {\color[HTML]{444444} 0.013±0.001} &
  {\color[HTML]{444444} 0.932±0.001} &
  {\color[HTML]{444444} 0.050±0.001} \\
6.2 &
  T1, T3, T4 &
  T1, T3, T4 (+T2) &
  {\color[HTML]{444444} 0.818±0.108} &
  {\color[HTML]{444444} 0.013±0.000} &
  {\color[HTML]{444444} 0.930±0.001} &
  {\color[HTML]{444444} 0.048±0.000} \\
6.3 &
  T1, T2, T4 &
  T1, T2, T4 (+T3) &
  {\color[HTML]{444444} 1.281±0.041} &
  {\color[HTML]{444444} 0.013±0.000} &
  {\color[HTML]{444444} 0.935±0.000} &
  {\color[HTML]{444444} 0.050±0.000} \\
6.4 &
  T1, T2, T3 &
  T1, T2, T3 (+T4) &
  {\color[HTML]{444444} 1.330±0.080} &
  {\color[HTML]{444444} 0.014±0.000} &
  {\color[HTML]{444444} 0.931±0.001} &
  {\color[HTML]{444444} 0.049±0.000} \\ \hline
\textbf{7} &
  \multicolumn{6}{c}{\textbf{\ac{MTML}, leave one task out format, addition of left out task in meta testing}} \\ \hline
7.1 &
  T2, T3, T4 &
  T2, T3, T4 (+ T1) &
  0.626±0.021 &
  0.013±0.001 &
  0.930±0.002 &
  0.050±0.001 \\
7.2 &
  T1, T3, T4 &
  T1, T3, T4 (+T2) &
  1.303±0.056 &
  0.014±0.000 &
  0.932±0.000 &
  0.050±0.000 \\
7.3 &
  T1, T2, T4 &
  T1, T2, T4 (+T3) &
  2.250±0.040 &
  0.014±0.000 &
  0.937±0.001 &
  0.051±0.001 \\
7.4 &
  T1, T2, T3 &
  T1, T2, T3 (+T4) &
  2.175±0.126 &
  0.014±0.000 &
  0.929±0.000 &
  0.049±0.000 \\ \hline
\end{tabular}
}
\end{tiny}
\vspace{-0.4cm}
\end{table*}

\textbf{Comparable to the state-of-the-art}: Table \ref{tab:sota} displays performance for the four tasks as available in the literature, along with the multi-task, and \ac{MTML} performance obtained in this work. 
The performance of our single task models is displayed in Exp.1 of Table \ref{tab:NYU} and \ref{tab:taskonomy}.
For the NYU-v2 data set, our proposed \ac{MTML} approach outperforms all the baseline works for three out of four tasks. For the semantic segmentation task, our single-task learning model performs best.
While in the taskonomy data set, \ac{MTML} shows best performance for surface normal prediction, and edge detection.
Both our single-task models and our multi-task models perform equally well when it comes to depth estimation.
For the semantic segmentation task, it is clear that our models' performance is not up to par with the performance reported in the literature. 
Similar deterioration in the performance of semantic segmentation task was observed in many of the experiments and is discussed in the Section \ref{seg_perform}.
Fig. \ref{fig:img_mat} displays the qualitative output of all the tasks on NYU-v2 and taskonomy datasets for a single task, \ac{MTL} and \ac{MTML} experiments, \ie `ours' in Table \ref{tab:sota} (corresponding to Exp. 1, 2.3 and 4.4 in Tables \ref{tab:NYU} \& \ref{tab:taskonomy}).
In terms of qualitative performance, \ac{MTML} clearly excels in all tasks.
Exp 4.4 uses \ac{MTML} to train and test all four tasks, so no new task is added.

\textbf{Why multi-task over single task?}: Exp 1, 2, and 5 in Table \ref{tab:NYU} and \ref{tab:taskonomy} show that vanilla \ac{MTL} works just as well, and in some cases even better, than its single-task counterpart.
It is important to note that our single-task models are now being used as the baseline for all comparisons.
Fig. \ref{fig:epochs} demonstrates that comparable results for \ac{MTL} can be achieved in far fewer training epochs than when they are trained individually (plots for Exp. 1 vs 2.3).
These results also help determine which task combination work best and which suffers from negative information transfer between the tasks. 
For example, in Table \ref{tab:taskonomy} Exp 2, shows three types of task combinations wherein the semantic segmentation task performs the worst in the ensemble as compared to their single task (Exp.1).
While the other tasks in the combination are unaffected.
Similarly in Table \ref{tab:NYU} Exp.5, the task combination [$\mathrm{T_1, T_2, T_3}$] \ie Exp. 5.4 is performing comparatively better than the others.

\textbf{Better and faster adaptation of new task}:
Exp 3,4,6, \& 7, display the performance of \ac{MTL} and \ac{MTML} when a new task is added during the test phase.
To begin with, first consider the \ac{MTL} performance on NYU-v2 dataset \ie Table \ref{tab:NYU}.
In \ac{MTL} setting, when surface normal estimation is added as a new task (Exp. 3.1, 3.2, 7.3), it performs better than or similar to single task learning but in fewer training epochs for both the datasets.   
While, the error increases w.r.t. single task when edge detection is added as unseen task \ie Exp. 3.2, 3.3, 7.4.
For some of the multi-task models  like, Exp. 6.1, 6.2, 6.3 the addition of an unseen task to the pre-trained models  which are Exp. 7.1, 7.2, 7.3 not only gives comparably good performance on the unseen task, but it also enhances the metrics for the already trained tasks (compare Exp. 6 and 7). 
An overall comparative analysis shows that, \ac{MTL} is better when the tasks are trained together \ie Exp 2.3 (T1,T2,T3,T4 trained together) instead of adding a new task during the testing \ie Exp 3.3 (T1,T2,T3, +T4).
Almost similar traits as above are also valid for the taskonomy dataset.

In the \ac{MTML} setting \ie Exp 4 \& 7, the meta trained models are introduced with new unseen tasks.
Exp 7 for the NYU-v2 dataset Table \ref{tab:NYU} reveals that \ac{MTML} outperforms the single task baseline performance, for various task combinations.
It is also discovered that, while semantic segmentation performance degrades slightly when meta trained, it performs best when added as a new task during testing, \ie Exp 7.1.
When comparing \ac{MTML} to single-task learning and their equivalent \ac{MTL} experiments, very similar, and in some cases even better, evaluation metrics can be observed.
Therefore, if models are trained using \ac{MTML}, it is simple to add new, previously unseen tasks (both homogeneous and heterogeneous).
While comparing the performance of \ac{MTML} and \ac{MTL} when an unseen task is added (more precisely fine-tuned) during testing, it is observed that \ac{MTML} excels, note Exp. 3 vs Exp. 4 and Exp. 6 vs Exp. 7. 
Table \ref{tab:taskonomy}, Exp. 7 from the taskonomy dataset shows that the new task achieves performance on par with that of the single-task and multi-task variants with significantly fewer training iterations.
For both the datasets better performance is attained in significantly lower number of fine-tuning epochs than in \ac{MTL} and corresponding single-task learning (see Fig. \ref{fig:epochs}), behavior that can be anticipated when employing MAML\cite{pmlr-v70-finn17a}.
Although, the trade off in the proposed \ac{MTML} is that it takes a large number of multi-task meta training epochs.
It is also evident that when tasks are trained jointly, the number of epochs decreases significantly in contrast to single task training. 

\textbf{Discussions}: 
After analyzing the experiments for both the datasets, it is evident that the proposed \ac{MTML} paradigm allows for easy addition and faster adaptation of a new task with equally good performance as compared to single task learning. 
For some instances in the result tables \ac{MTL} performs identical to \ac{MTML}, but it can be argued that \ac{MTML} achieves similar performance in fewer epochs than \ac{MTL}.
Even though MTML's performance falls short on the semantic segmentation task, better results are achieved when the task is added as a unseen one.
For the depth and surface normal estimation task, overall the qualitative results of \ac{MTML} demonstrate greater impact, even though quantitatively the performance is marginally better than \ac{MTL}. 
\ac{MTML} excels because meta-knowledge of task combinations is accumulated in the meta parameters that are shared with the unseen task.
The bi-level optimization in the multi-task meta training stage facilitates learning generalized parameters for effectively incorporating an unknown task in a multi-task setting.
In \ac{MTL}, however, the optimal parameters for collectively learning the source tasks are shared.
Furthermore, the highlight of the few-shot variant of meta learning is that, it performs very well on an unseen task; however, the target tasks (unseen tasks) are substantially similar (but not identical) to the source tasks (learning episodes), \ie usually it is a new label class.
The \ac{MTML} paradigm, however, that is put forth in this work has provided a method for effectively adding a new (dissimilar) task  while also improving the performance of all (both source and target) tasks.
On the other hand \ac{MTL} performs very well when the multiple tasks are trained together, as compared to addition (or fine-tuning) an unseen task.

\section{Unsatisfactory performance of the semantic segmentation task} \label{seg_perform}
\begin{figure}[ht]
\begin{center}

\begin{tabular}{c c c c}

Image & GT & Pred. & Act. map \\

\adjustimage{height=2cm,valign=m}{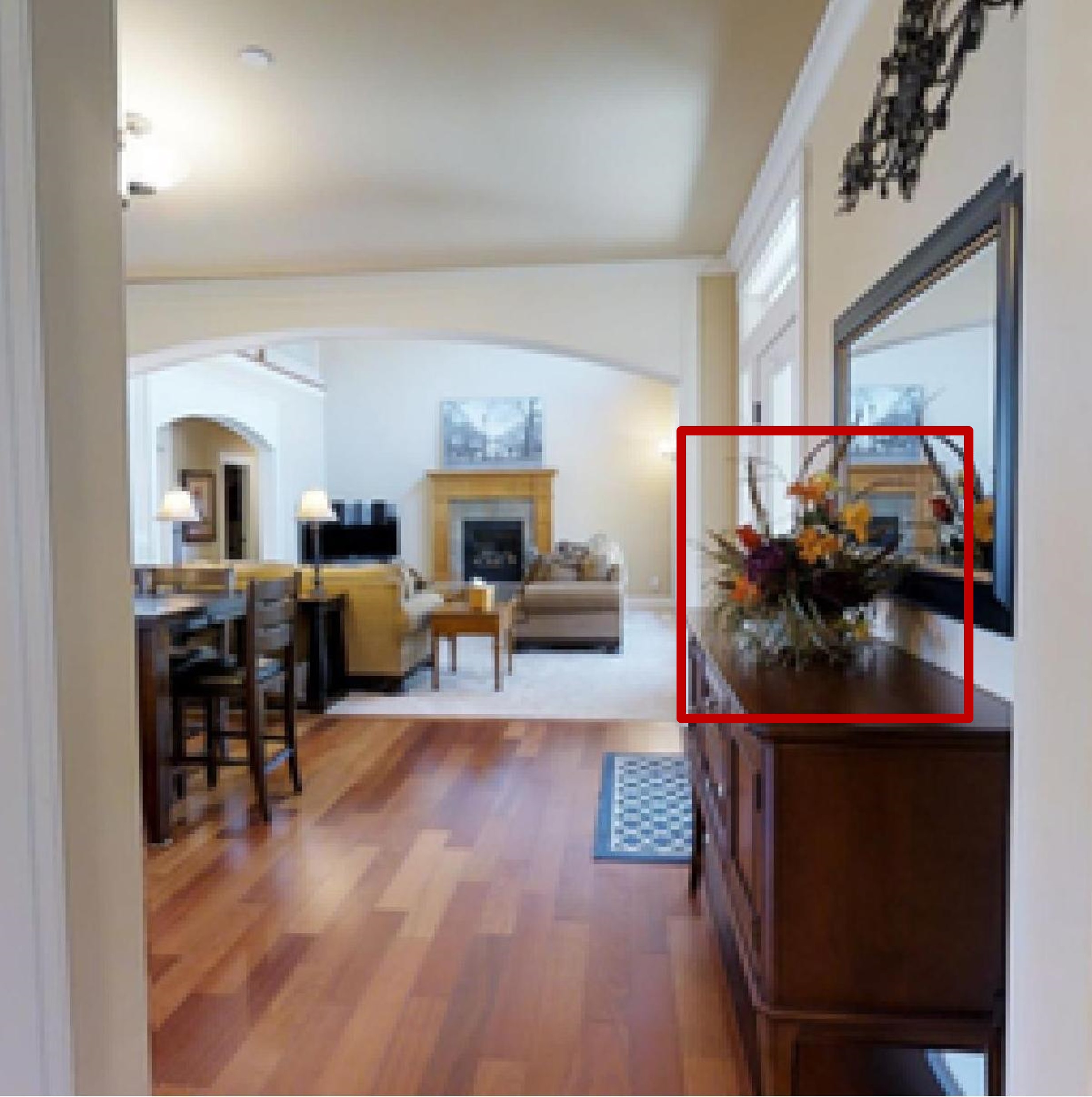} &
\adjustimage{height=2cm,valign=m}{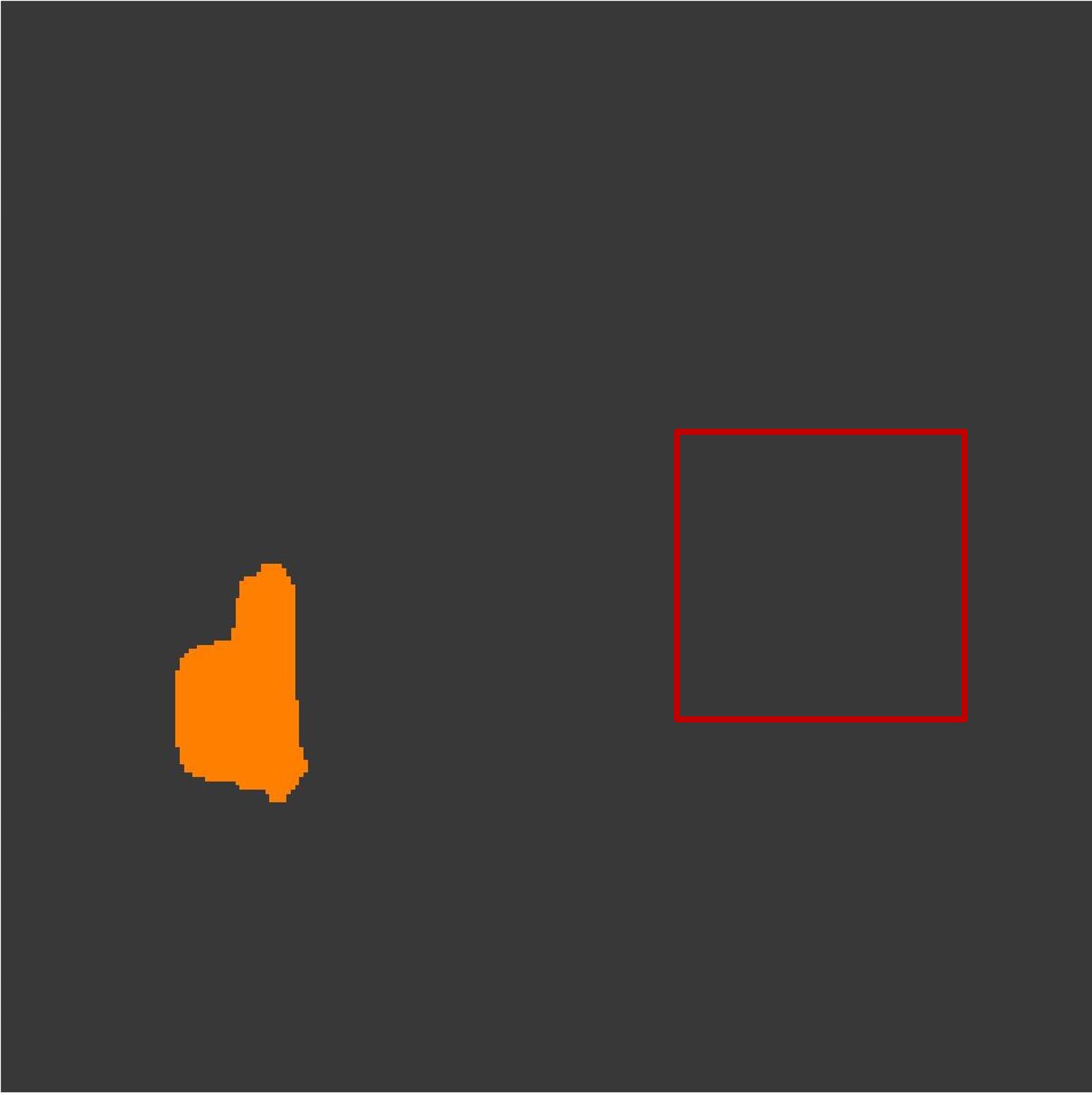} & 
\adjustimage{height=2cm,valign=m}{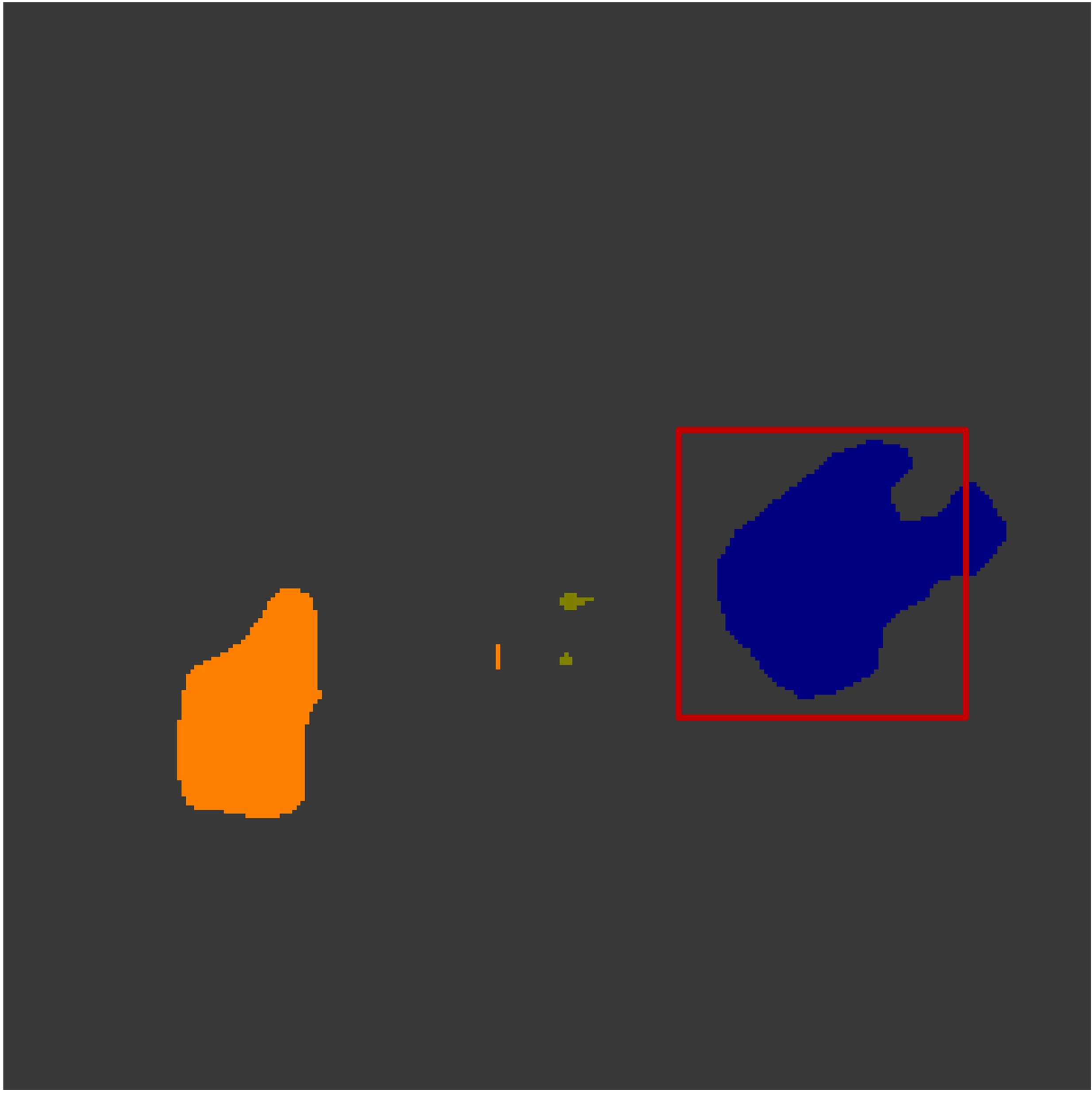} & \adjustimage{height=1.9cm,valign=m}{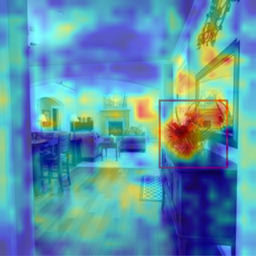} \\

\adjustimage{height=2.1cm,valign=m}{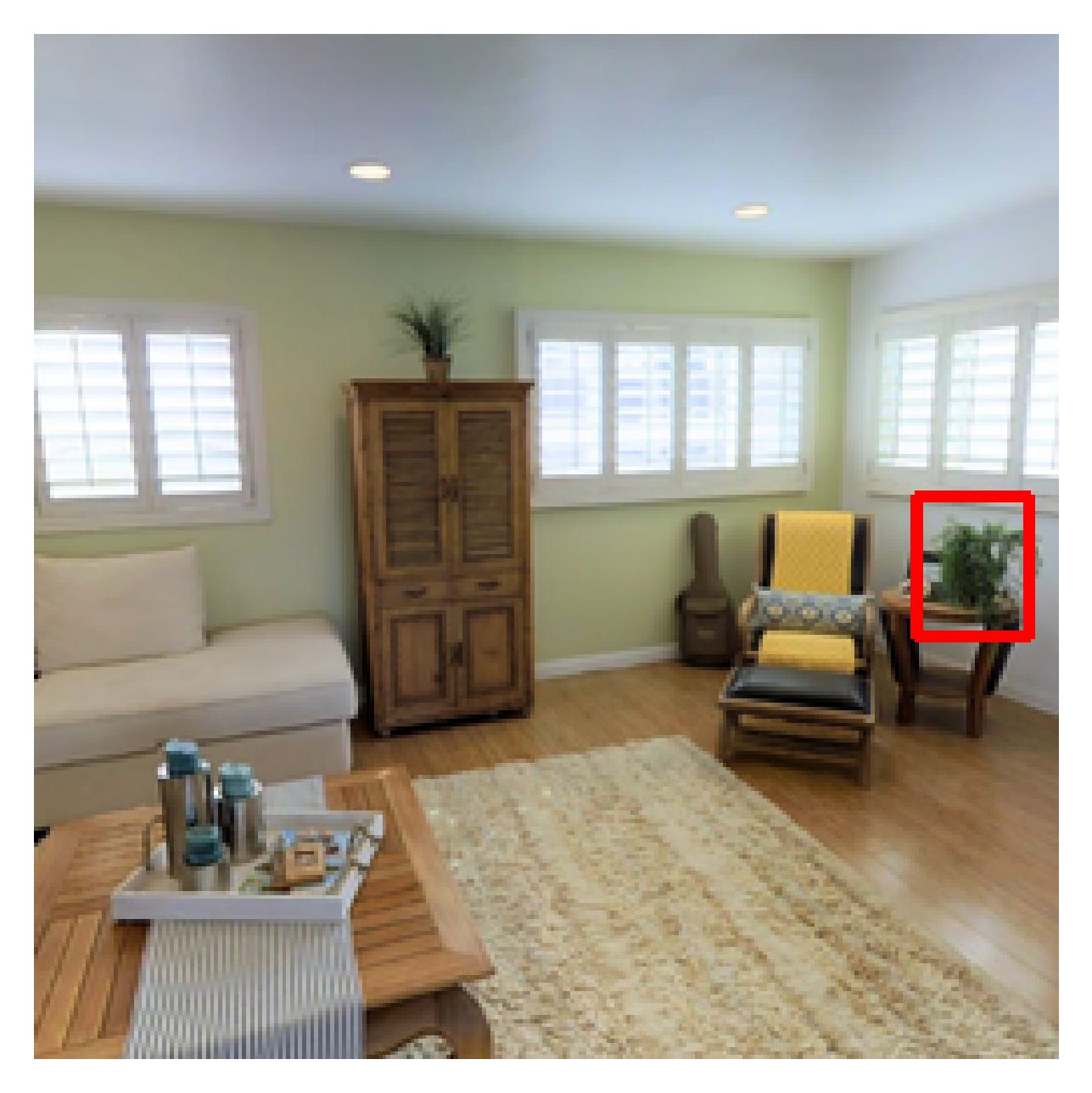} & 
\adjustimage{height=2.1cm,valign=m}{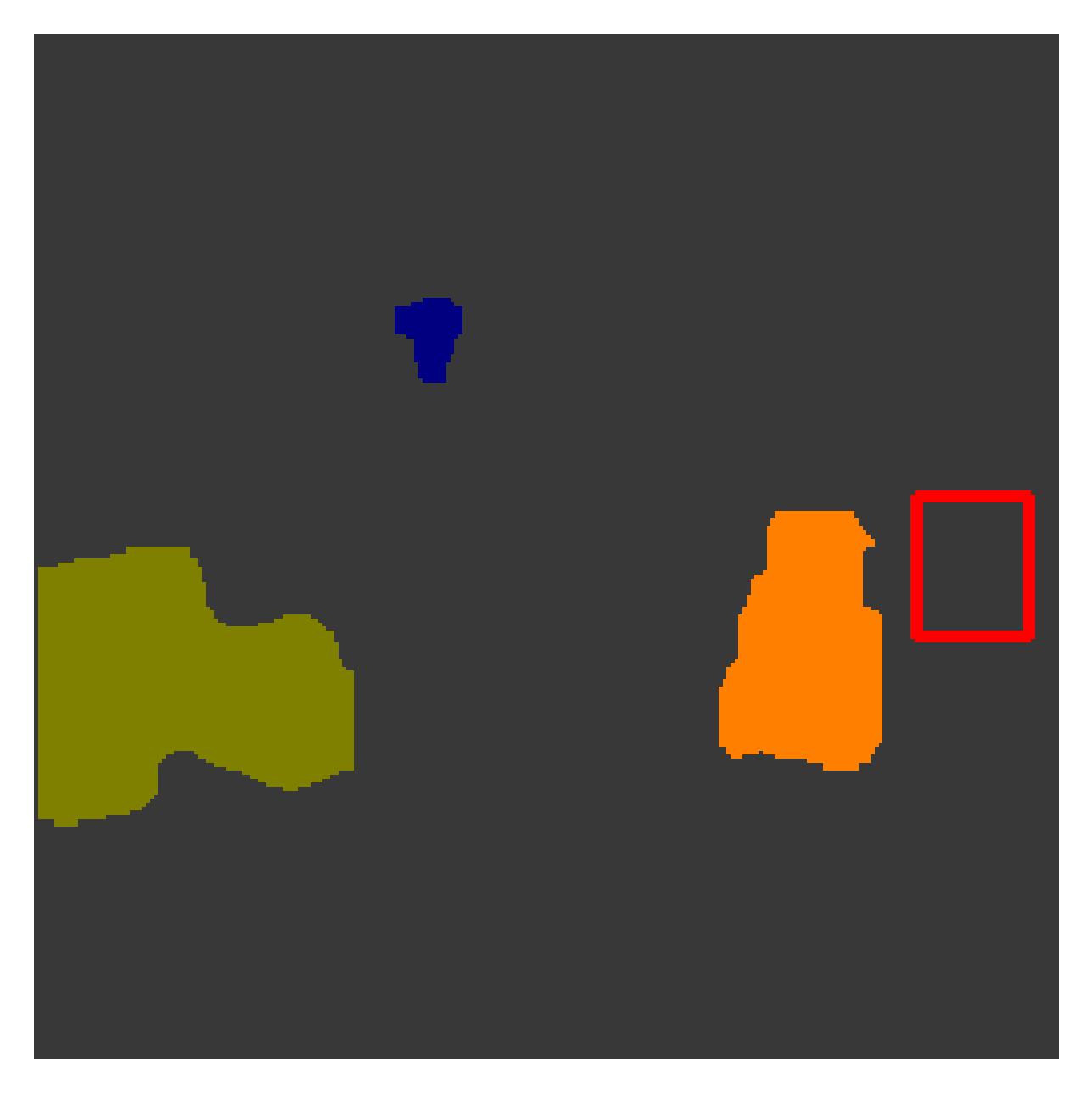} & \adjustimage{height=2.1cm,valign=m}{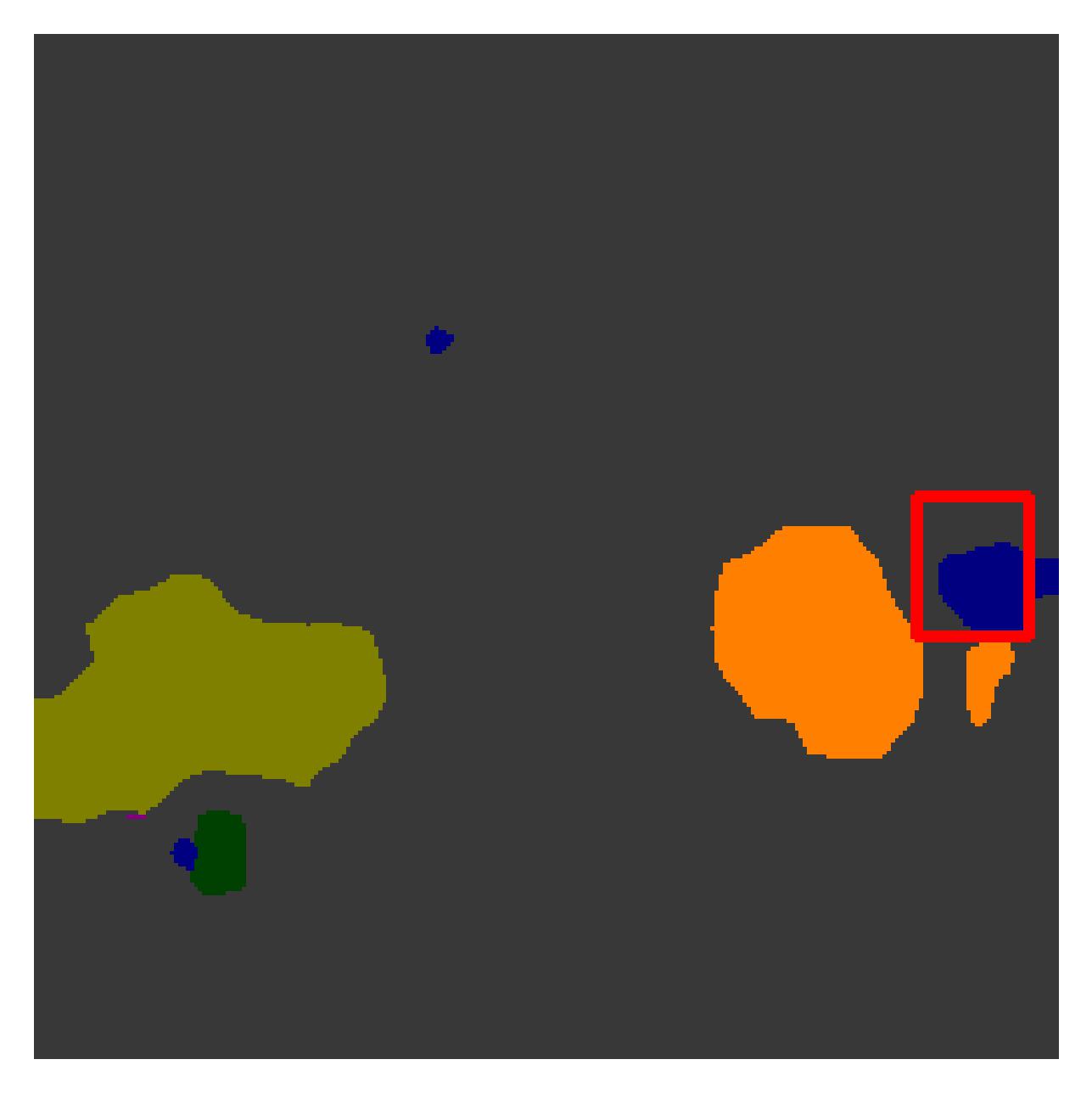}& \adjustimage{height=1.9cm,valign=m}{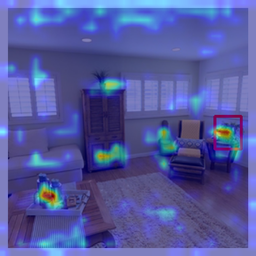}\\

\adjustimage{height=2cm,valign=m}{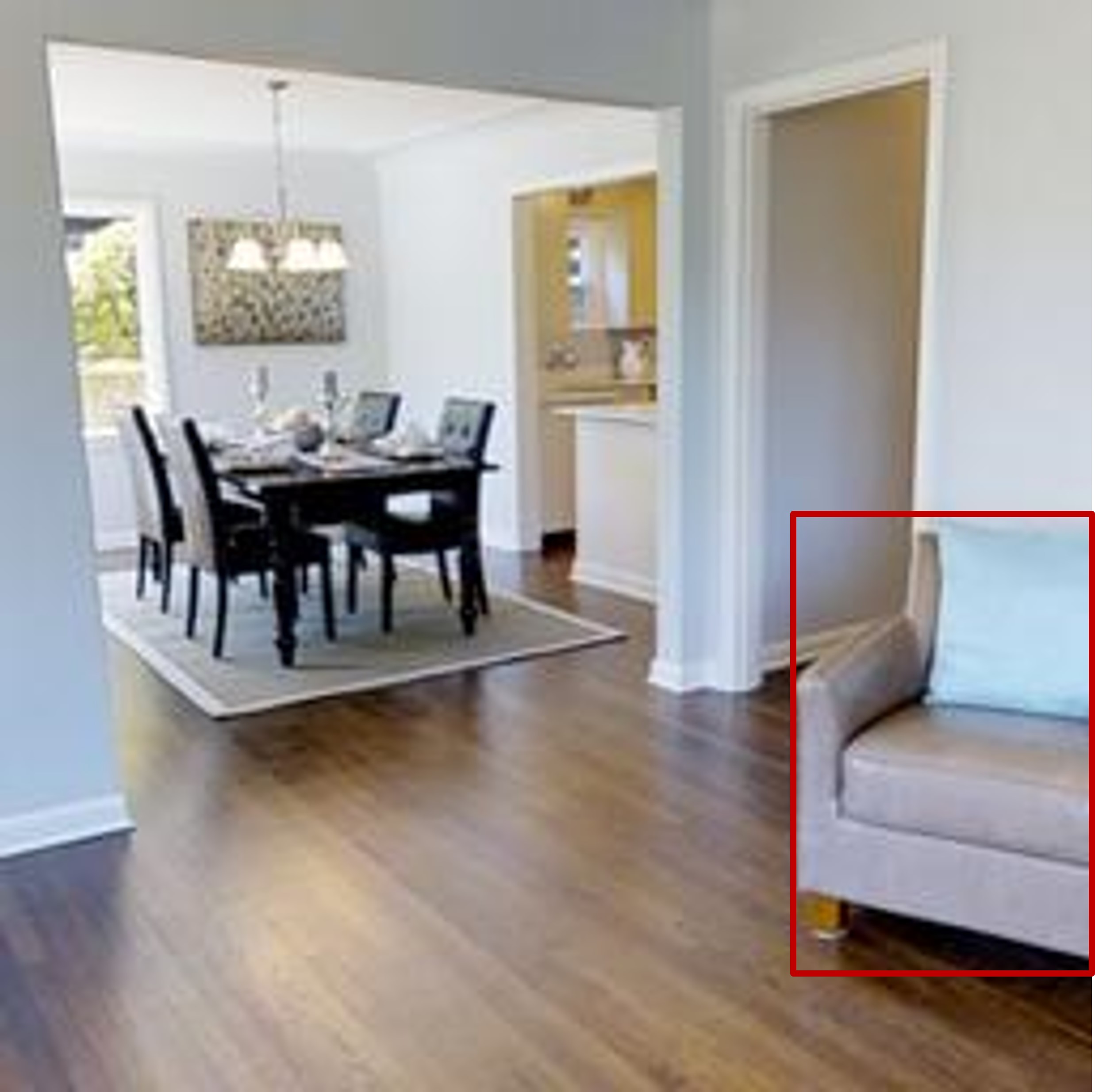} & 
\adjustimage{height=2cm,valign=m}{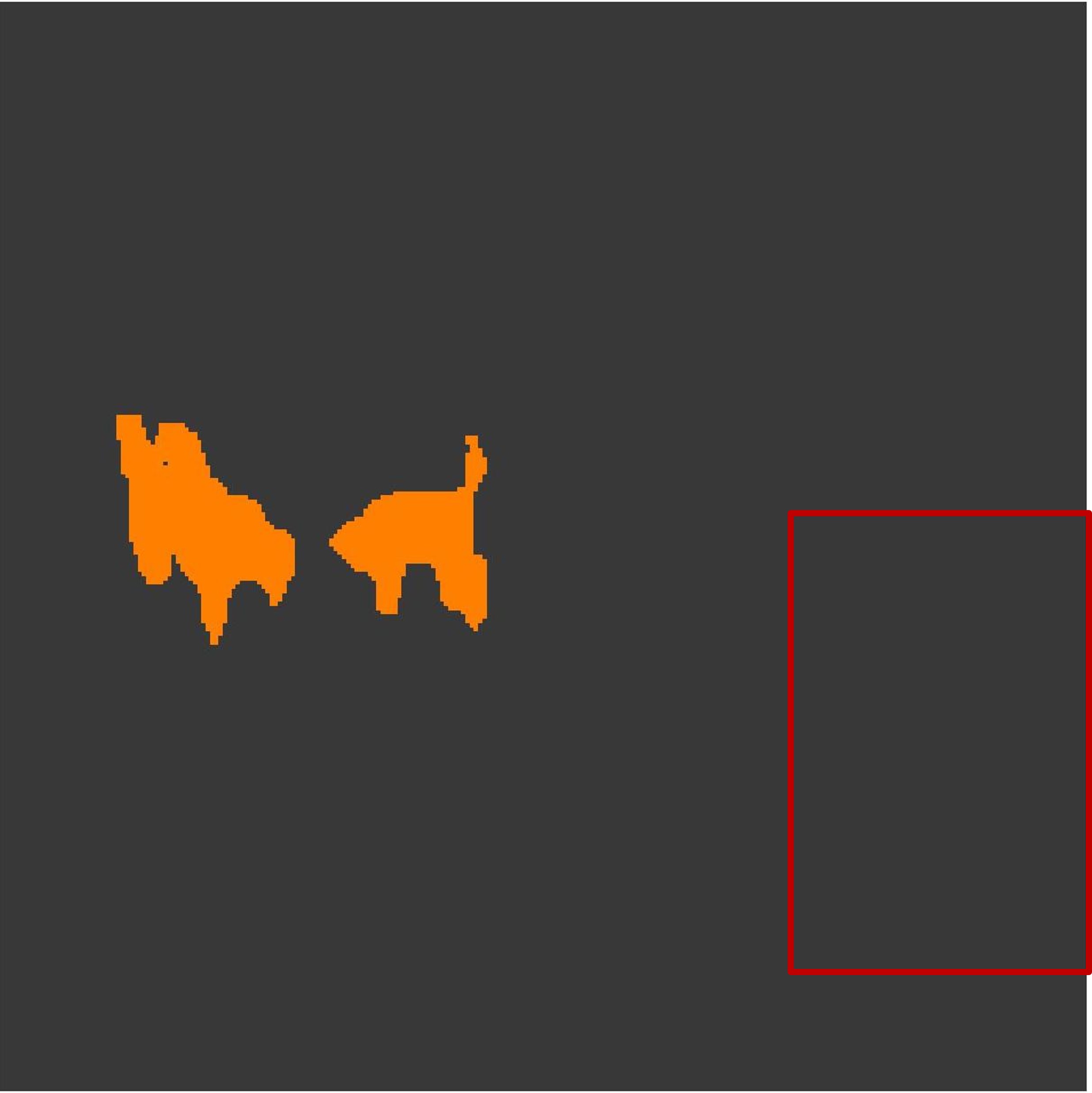} & \adjustimage{height=2cm,valign=m}{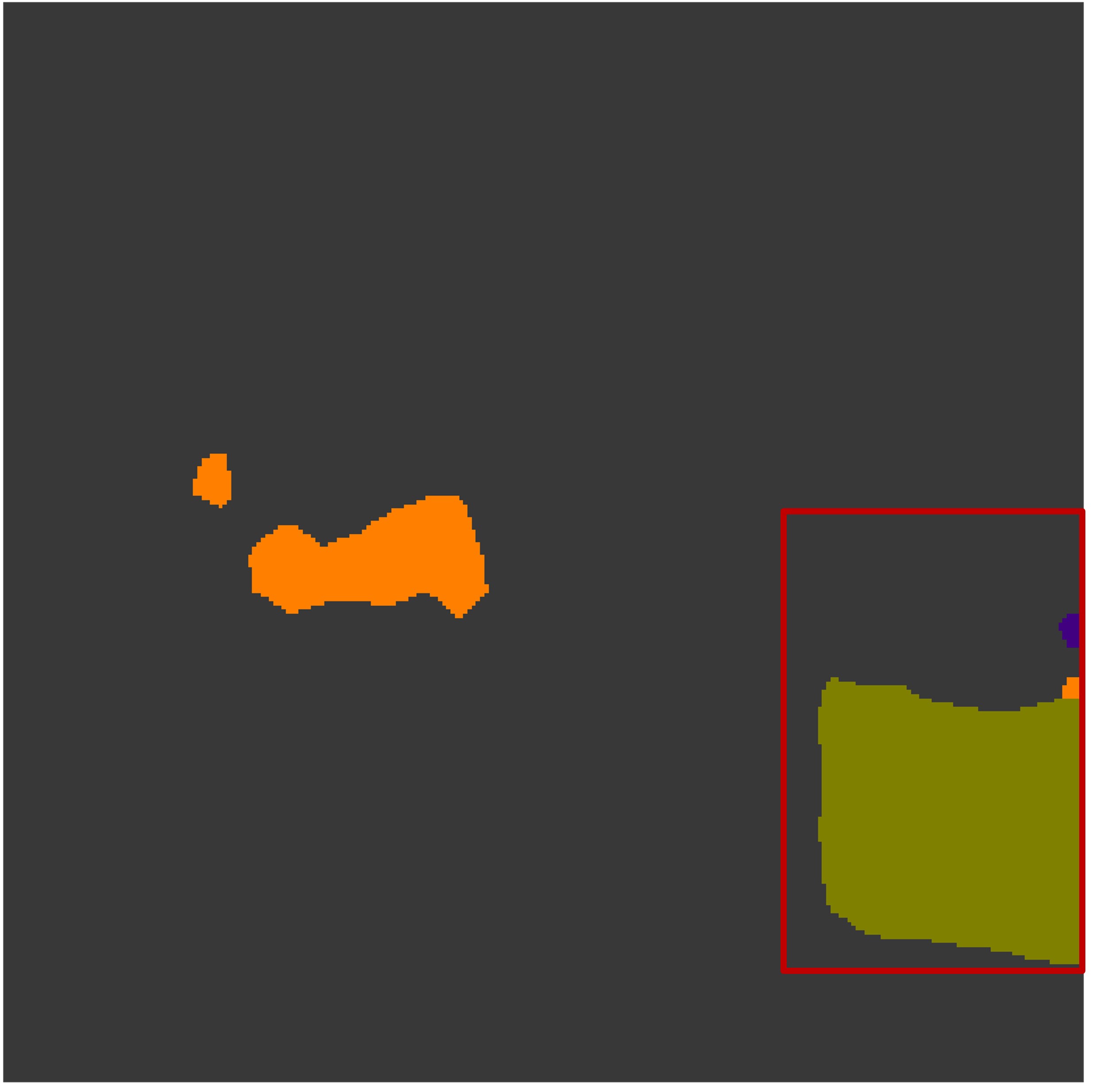}& \adjustimage{height=1.9cm,valign=m}{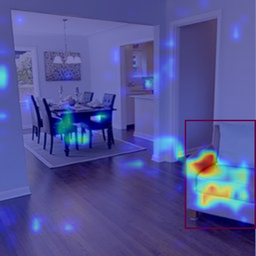}\\


\end{tabular}
\vspace{-0.2cm}
\caption{This figure shows the input RGB image, the corresponding ground truth (GT) segmentation mask and the predicted segmentation mask (Pred.) by the \ac{MTML} model and the class activation maps (Act. map) of the missing class. The red box highlights the object not present in the ground truth (pseudo labels), but the proposed \ac{MTML} model learns to detect and segment them. For example, in row 1 and 2, the plant (class-\textit{potted plant}) is not identified in the pseudo labels. Rows 1 and 2 show the class activation map of class-\textit{potted plant}, and row 3 shows class-\textit{couch}.}
\label{fig:seg_fig}
\vspace{-0.5cm}
\end{center}

\end{figure}

For the taskonomy data set, from the performance analysis of table \ref{tab:taskonomy}, we observe that for the semantic segmentation task the outcomes are very unsatisfying quantitatively.
In spite of this, the models are learning to segment the images, and the qualitative performance is encouraging.
We investigated the cause for this and came up with a few possible explanations:

    \textit{First}, the segmentation masks given for each image are not annotated by humans, in fact, they are pseudo labels, \ie  the masks are distilled from FCIS \cite{8099955}. 
    While testing the \ac{MTML} experiments 4, and 7, it was discovered that these models were found to be more effective at learning than the ground truth labels (masks) because they were able to recognize objects in the image that were not included in the ground truth annotation but should have been.
    A few such examples are shown in Fig. \ref{fig:seg_fig}.
    This is one of the reasons why the semantic segmentation results are degraded.
    Fig. \ref{fig:seg_fig} shows the class activation maps of the classes absent in the ground truth, which is learned by \ac{MTML}.
    The better explainability of class-specific discriminative regions is facilitated by the class activation maps \cite{jacobgilpytorchcam}.
    Observing the qualitative findings reveals that the semantic segmentation task is doing satisfactorily; however, it does not produce very good quantitative metrics when compared to the incomplete ground truth masks.
    
    \textit{Second}, the segmentation classes in the taskonomy dataset are highly unbalanced, several images only contain `background' class or just one class, behaving like a binary class problem, although the overall dataset is a multi-class. This is because only 17 of the 80 segmentation classes used to train the FCIS \cite{8099955} were in the taskonomy dataset, and rest are marked as background. 

    \textit{Third}, reason for the under-performance can probably be `negative transfer' \cite{MTL_rich}.
    Although all the tasks are pixel-level, negative transfer is possible because of the nature of the tasks: segmentation is the only pixel-level classification task while all others are regression tasks. 
    This can be solved using task-specific hyper-parameters, like learning rate, weight decay, etc. 
    Since, in this work to obtain an appropriate comparison between learning paradigms all tasks use identical hyper-parameters. 
    We believe this is one of the main reason of slight degradation in the performance of the semantic segmentation task for the \ac{MTML} experiments (Exp. 4, 7.2, 7.3, 7.4 of Table \ref{tab:NYU}), also for the NYU-v2 dataset.


\setlength{\tabcolsep}{2pt}
\renewcommand{\arraystretch}{1.5}

\section{Conclusion and Future scope}
This work proposes to combine multi-task and meta learning by introducing multi-task learning episodes to meta train the network and allows to further test the network by introducing a new (unseen) task. 
Theoretically, if the properties of meta and multi-task learning are combined, such an ensemble should deliver good performance for the new task in fewer steps than training the single task from scratch.
Comprehensive empirical analysis of \ac{MTML} performance supports the hypothesis that \ac{MTML} indeed performs best compared to vanilla \ac{MTL} and single-task learning. 
In addition to that, it allows for swift adaptation to an unseen task.
However, we do observe that the semantic segmentation task under-performs, because the presences of incorrect pseudo labels, or negative task transfer. 
Overall, \ac{MTML} is a approach that can efficiently train several tasks together and is capable of faster adaptation to new tasks if these are not too far off from the already learned tasks.  
Furthermore, MTML is a task and model agnostic paradigm that may be utilized for any heterogeneous or homogeneous tasks (due to multi-task learning) and any multi-task architecture (due to optimization based meta learning).

Future research could benefit from using better loss-balancing techniques and task-specific hyper-parameters to further boost performance across all the tasks and, in particular, to prevent negative transfer.
Because the proposed method is model and task agnostic, other complex multi-task architectures can be investigated for more complex tasks.
However, such intricate fusion models, while extremely useful, come at a high computational cost.
While we outline one strategy for fusing multi-task and meta-learning, we acknowledge that there are likely other strategies for doing so to reap the benefits of both learning paradigms.

\section*{Acknowledgment}
The authors express their gratitude for the computational resources made available by the National Supercomputer Centre at Link\"{o}ping University, specifically the Berzelius supercomputing system, support by the Knut and Alice Wallenberg Foundation. 
We would also like to extend our sincere appreciation to Konstantina Nikolaidou (Lule\aa~University of Technology) and Sameer Prabhu for their insightful comments and constructive feedback on earlier drafts of this paper.
\bibliography{egbib}
\bibliographystyle{IEEEtran}


\end{document}